\documentclass{article} 
\usepackage{iclr2026_conference,times}

\usepackage{amsmath,amsfonts,bm}

\def\eqref#1{equation~\ref{#1}}

\def\1{\bm{1}}

\DeclareMathAlphabet{\mathsfit}{\encodingdefault}{\sfdefault}{m}{sl}
\SetMathAlphabet{\mathsfit}{bold}{\encodingdefault}{\sfdefault}{bx}{n}

\usepackage{hyperref}
\usepackage{url}
\usepackage{multirow}
\usepackage{booktabs}
\usepackage{graphicx} 
\usepackage{algorithm}
\usepackage{algpseudocode}
\usepackage{amsmath}
\usepackage{multirow}
\usepackage{array}
\usepackage{colortbl}
\usepackage{xcolor}
\usepackage{pifont}
\usepackage{wrapfig}

\usepackage{float} 
\newcommand{\downred}[1]{\textcolor{red}{#1 $\downarrow$}}
\newcommand{\upgreen}[1]{\textcolor{green!60!black}{#1 $\uparrow$}}

\usepackage{geometry}
\usepackage{pgfplots, pgfplotstable}
\pgfplotsset{compat=1.18}
\usetikzlibrary{pgfplots.groupplots} 
\usepgfplotslibrary{statistics} 
\usetikzlibrary{fit,arrows.meta,positioning,shapes.misc}
\usepgfplotslibrary{fillbetween}
\usepackage[capitalize]{cleveref}
\crefname{section}{Section}{Sections}
\crefname{table}{Table}{Tables}
\crefname{table}{Table}{Tables}
\crefname{figure}{Figure}{Figures}
\Crefname{figure}{Figure}{Figures}
\pgfmathdeclarefunction{invgauss}{2}{%
  \pgfmathparse{sqrt(-2*ln(#1))*cos(deg(2*pi*#2))}%
}

\tikzset{
  tightplot/.style={
    title style={at={(0.5,0.98)}, anchor=north, font=\bfseries\footnotesize},
    ylabel style={at={(0.02,0.98)}, anchor=north west, rotate=0, font=\scriptsize},
    xlabel style={at={(0.98,0.02)}, anchor=south east, font=\scriptsize},
    tick label style={font=\scriptsize},
    label style={font=\scriptsize},
    ymajorgrids,
    xmajorgrids,
    grid style={line width=.1pt, draw=black!15},
    tick align=outside,
    axis on top,
    enlargelimits=false,
    clip=true
  }
}
\usepgfplotslibrary{statistics} 
\usepackage{subcaption}        

\usepackage{multirow}
\usepackage{xcolor}
\usepackage{graphicx,booktabs, caption}
\usepackage{colortbl}
\usepackage{xcolor}
\definecolor{lightgreen}{RGB}{220, 255, 220}
\definecolor{medgreen}{RGB}{170, 235, 170}
\definecolor{darkgreen}{RGB}{120, 200, 120}
\usetikzlibrary{arrows.meta,positioning,fit,calc,shapes.misc,shapes.geometric}
\usepackage{amsmath}

\title{CAT:\\ Post-Training Quantization Error
Reduction via Cluster-based Affine Transformation}

\author{Ali Zoljodi\thanks{Corresponding author} ~\& Masoud Daneshtalab \\
Department of Intelligent Future Technologies (IFT)\\
Mälardalen University\\
Västerås, 72123, Sweden \\
\texttt{\{ali.zoljodi,masoud.daneshtalab\}@mdu.se} \\
\And
Radu Timofte \\
Computer Vision Lab, CAIDAS \& IFI\\
University of Wurzburg \\
John Skilton Str. 4a, 97074 Würzburg, Germany \\
\texttt{\{radu.timofte\}@uni-wuerzburg.de} \\
}

\begin{document}

\maketitle

\begin{abstract}
Post-Training Quantization (PTQ) reduces the memory footprint and computational overhead of deep neural networks by converting full-precision (FP) values into quantized and compressed data types.
While PTQ is more cost-efficient than Quantization-Aware Training (QAT), it is highly susceptible to accuracy degradation under a low-bit quantization (LQ) regime (e.g., 2-bit).
Affine transformation is a classical technique used to reduce the discrepancy between the information processed by a quantized model and that processed by its full-precision counterpart; however, we find that using plain affine transformation, which applies a uniform affine parameter set for all outputs, worsens the results in low-bit PTQ.
To address this, we propose Cluster-based Affine Transformation (CAT), an error-reduction framework that employs cluster-specific parameters to align LQ outputs with FP counterparts.
CAT refines LQ outputs with only a negligible number of additional parameters, without requiring fine-tuning of the model or quantization parameters. We further introduce a novel PTQ framework integrated with CAT. Experiments on ImageNet-1K show that this framework consistently outperforms prior PTQ methods across diverse architectures and LQ settings, achieving up to 53.18\% Top-1 accuracy on W2A2 ResNet-18. Moreover, CAT enhances existing PTQ baselines by more than 3\% when used as a plug-in.
We plan to release our implementation alongside the publication of this paper.

\end{abstract}
\begin{figure}[ht]
    \centering
    \includegraphics[width=1\textwidth]{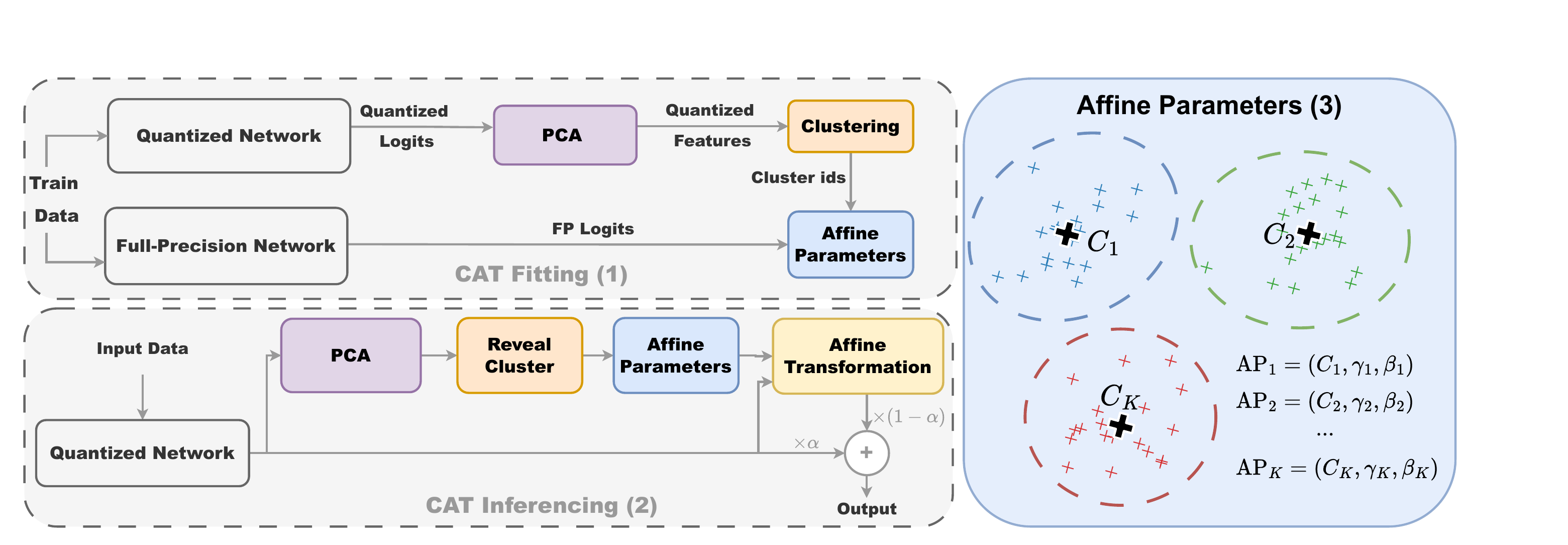}
    \caption{\small Cluster-based  Affine Transformation (CAT); (1): CAT Fitting is the process of fitting the clustering model and estimating $\gamma$ and $\beta$ parameters for each cluster.(2) CAT Inferencing, the process of error reduction using CAT. (3) Affine Parameters (AP), a set of parameters we store for each cluster of CAT. }
    \label{fig:CAT}
\end{figure}

\section{Introduction}
Deep neural networks (DNNs) achieve remarkable performance on different computer vision tasks~\cite{ravi2025sam,Xiao_2025_CVPR,Zhu_2025_CVPR,Li_2025_CVPR,wang2025retrievalaugmented,zanella2024boosting}.
but demand prohibitive memory and computation due to millions of floating-point parameters. Quantization, which reduces the precision of weights and activations to low bit-widths, offers an efficient compression strategy and is now widely supported by modern hardware. 
Post-Training Quantization (PTQ) is a practical model compression which determines quantization parameters without retraining or fine-tuning, requiring only a small calibration set.
Despite this promise, PTQ suffers from a severe accuracy degradation when pushed to LQ regimes. 
In this work, we investigate the affine transformation ~\cite{weisstein2004affine,ma2024affinequant} ability to reduce PTQ output errors to refine its accuracy degradation.
\begin{wraptable}{r}{0.5\linewidth}
\centering
    \label{tab:affine_transformation}
    \begin{tabular}{c c l | c}
    \toprule
    $w$ & $a$ & \textbf{Method} & \textbf{Acc (\%)} \\
    \midrule
    \multirow{3}{*}{2} & \multirow{3}{*}{2} 
      & No affine transformation & 52.84 \\
      & & Plain affine transformation & 52.32 \\
      & & CAT (Ours) & \textbf{53.18} \\
    \midrule
    \multirow{3}{*}{4} & \multirow{3}{*}{2} 
      & No affine transformation & 58.58 \\
      & & Plain affine transformation & 58.22 \\
      & &  CAT (Ours)& \textbf{58.80} \\
    \midrule
    \multirow{3}{*}{2} & \multirow{3}{*}{4} 
      & No affine transformation & 65.12 \\
      & & Plain affine transformation & 65.17 \\
      & &  CAT (Ours) & \textbf{65.25} \\
    \midrule
    \multirow{3}{*}{4} & \multirow{3}{*}{4} 
      & No affine transformation & 69.17 \\
      & & Plain affine transformation & 69.14 \\
      & &  CAT (Ours) & \textbf{69.27} \\
    \bottomrule
    \end{tabular}
\caption{Top-1 accuracy of ResNet-18 under quantization: comparison between no affine transformation, plain affine transformation, and CAT (ours).}
\label{tab:cat_results}
\end{wraptable}
We first investigate a uniform set of affine transformation parameters, referred to as a plain affine transformation. As shown in \cref{tab:cat_results}, this approach fails to recover predictions and often further degrades the Top-1 accuracy of LQ ResNet-18.
To alleviate the issue, we leverage clustering PTQ outputs and find a specific affine transformation parameter set for each cluster. 
Our proposed method, Cluster-based Affine Transformation (CAT), yields superior error reduction results for PTQ and especially the low-bit quantization (LQ) (e.g., 2-bit) regime.
This strategy substantially reduces the output gap between FP and LQ, improving accuracy under low-bit settings with only a negligible number of additional affine parameters. 
Building on the advantages of CAT, we propose a novel PTQ framework.
Compared to the baseline, our proposed PTQ framework raises the Top-1 accuracy of 2-bit quantized (W2A2) ResNet-18 to 53.18\%. For compact DNNs such as MNasX2, CAT achieves a $+1\%$ accuracy improvement in the 2-bit quantization of both weight (W) and activation (A)  (W2A2).
Because of its ability to directly reduce output errors, CAT can be used as a plug-in for a wide range of PTQ methods. Our results show that, with this capability, CAT improves Top-1 accuracy by more than 3\% for some PTQ methods.
Our contributions can be summarized as follows:
\begin{itemize}
    \item We propose Cluster-Affine Transformation (CAT), a novel output-level error reduction method that leverages the natural clusterability of logits to improve alignment.
    \item We introduce a novel state-of-the-art post-training quantization framework, which achieves consistent accuracy improvements across diverse architectures and LQ settings, surpassing prior PTQ methods with negligible overhead.
    \item We achieve higher Top-1 accuracy compare to PTQ baselines on ImageNet-1k for different DNNs such as ResNet-18/50, MobileNetV2, and RegNetX.
\end{itemize}
\section{Related Work}
Quantization~\cite{Li_2023_ICCV,NEURIPS2022_86e7ebb1,qin2025mixed,Cai_2020_CVPR,harma2025effective,li2025svdquant,li2024loftq,Zhou_2025_CVPR,saxena2025resq} is a widely used model compression technique for deep neural networks (DNNs) that reduces model size and accelerates inference by representing weights and activations with lower bit precision.
In practice, two paradigms exist: quantization-aware training (QAT), which incorporates quantization during model re-training (achieving high accuracy but at the cost of additional training on full datasets), and post-training quantization (PTQ), which converts a pre-trained model to low-bit format using only a small unlabeled calibration set without full re-training.
While QAT preserves accuracy better, PTQ is efficient in development.
\paragraph{Post-training Quantization (PTQ)}
~\cite{NEURIPS2019_c0a62e13,Liu_2023_CVPR,nahshan2021loss,NEURIPS2019_c0a62e13,pmlr-v119-nagel20a,pmlr-v119-wang20c,wei2022qdrop, 10.1007/978-3-031-19775-8_12,lin2021fq,10.1145/3503161.3547826, 10536014,shi2025on,ding2025cbq,10839431,10938638,Wu_2025_CVPR,Chen_2025_CVPR,Shen_2025_CVPR} is the process of determining quantization scale factors and zero-point without retraining or fine-tuning a model’s weights.
The key challenge in PTQ is estimating the minimum and maximum values of weights and activations, and identifying outliers to clip. Early works on low-bit post-training quantization employed analytic methods to derive optimal clipping thresholds. 
Banner et al.~\cite{NEURIPS2019_c0a62e13} proposed limiting activation ranges by statistically deriving activation distributions of tensors and determining the per-channel bit-width.
Recent PTQ research has introduced methods to minimize accuracy loss by optimizing layer-wise or block-wise reconstructions.
AdaRound~\cite{pmlr-v119-nagel20a} is a layer-wise reconstruction method that minimizes the local loss by adapting scale factors.
To cover cross-layer interaction, BRECQ~\cite{li2021brecq} extends layer-wise reconstruction to a group of layers (blocks) with second-order error approximations.
PD-Quant~\cite{Liu_2023_CVPR} proposes reconstructing layers and blocks via global loss minimization by comparing the network outputs before and after quantization.
To improve the stability of PTQ, QDrop~\cite{wei2022qdrop} randomly drops activation quantization during calibration to improve generalization and robustness.
Some other works adopt different PTQ formulations.
Mr.BiQ~\cite{9879931} introduces a non-uniform, multi-level binary quantizer, where both scaling factors and binary codes are treated as learnable parameters and optimized jointly to minimize block-wise reconstruction error.
Prepositive Feature Quantization (PFQ)~\cite{10239177} reorganizes the PTQ framework by moving feature quantization before, rather than after, each layer.
In practice, PFQ requires different calibration schedules to effectively align quantized and FP representations in LQ settings.

\paragraph{Quantization-Aware Training (QAT)}~\cite{JMLR:v18:16-456, tailor2021degreequant, 9463531, DBLP:journals/corr/abs-2407-11062,8999090,wei2025roste} integrates quantization into the training process to simulate low-bit computations during forward and backward passes~\cite{he2024efficientdm}.
Early QAT works~\cite{Esser2020LEARNED, Zhang_2018_ECCV} rely on the straight-through estimator (STE)~\cite{bengio2013estimating} to approximate gradients of non-differentiable quantization functions, allowing end-to-end optimization.
Recent works combined QAT with knowledge distillation~\cite{kim2019qkd} and mixed-precision strategies~\cite{Wang_2019_CVPR} to further reduce accuracy degradation, enabling robust deployment of convolutional and Transformer models under aggressive quantization constraints.
\section{Method}
(The quantization notation is described in \cref{section:qn}.) Our post-training procedure optimizes all quantization parameters exclusively via the relative entropy (KL divergence) between the full-precision (FP) and quantized output distributions. Let $z_{FP}(x)$ and $z_{LQ}(x)$ denote the FP and quantized logits for input $x$. With temperature $T$ and $p$ as the probability of the model output, define
\[
p_{FP}(x) := \operatorname{softmax}(z_{FP}(x)/T), \qquad
p_{LQ}(x;\Theta) := \operatorname{softmax}(z_{LQ}(x;\Theta)/T).
\]
We determine the scale-factor for each tensor by minimizing the KL divergence on a small calibration set $\mathcal{X}$:
\begin{equation}
\label{eq:kl-quant}
\mathcal{L}_{\mathrm{KL\_out}}
= \frac{1}{|\mathcal{X}|}\sum_{x\in\mathcal{X}}
\mathrm{KL}\!\big(p_{FP}(x)\,\big\|\,p_{LQ}(x)\big)
= \frac{1}{|\mathcal{X}|}\sum_{x\in\mathcal{X}}
\sum_{c} p_{FP}^{(c)}(x)\,\log\frac{p_{FP}^{(c)}(x)}{p_{LQ}^{(c)}(x)}.
\end{equation}
To avoid overfitting and preserve hardware-friendly ranges, we add a lightweight parameter regularizer:
\begin{equation}
\label{eq:param-reg}
\mathcal{L}_{\mathrm{reg}}
= \|\delta - \delta_0\|_2^2 + \|z - z_0\|_2^2,
\end{equation}
where $(\delta_0,z_0)$ are the initial (calibration) estimates. The objective is to minimize $\mathcal{L}_{\mathrm{CAT}}$ where,
\begin{equation}
\label{eq:stage1}
\mathcal{L}_{\mathrm{CAT}}
= \mathcal{L}_{\mathrm{KL\_out}} + \lambda_{\mathrm{p}}\,\mathcal{L}_{\mathrm{reg}}.
\end{equation}
This procedure ensures that each block is reconstructed to preserve the information content of its FP counterpart, reducing distributional mismatch layer by layer. 
The refined quantization parameters are then fixed for the remainder of the process, after which we apply logit-level transformation using CAT.  

\subsection{Cluster-based Affine Transformation for Error Reduction}
After refining quantization parameters, we address residual mismatches directly in the logit space through CAT (\cref{fig:CAT}). To make clustering more effective and computationally efficient, we use a \emph{principal component analysis} (PCA)~\cite{WOLD198737} to decompose the LQ logits $z_{LQ}$ . PCA reduces the dimensionality of the logits by discarding low-variance components, thereby (i) removing unnecessary complexity, (ii) improving cluster separability, and (iii) lowering clustering cost. 
Given calibration logits $\{z_{LQ}(x), z_{FP}(x)\}_{x\in\mathcal{X}}$, we apply PCA to reduce the dimensionality of $\{z_{LQ}(x)\}$ before clustering.
Note that PCA is only used to obtain clusters. 
The affine transformation itself is always applied in the original logit space of dimension $d$.
A clustering model is then fitted to the reduced features, yielding cluster assignments $c(x)\in\{1,\dots,K\}$.
For each cluster $C_k=\{x: c(x)=k\}$, we seek Affine Parameters ($\text{AP}_K$) $(\gamma_k,\beta_k)\in\mathbb{R}^d$ (with $d$ the dimensionality of logits) that best map quantized logits to their FP counterparts:
\begin{equation}
    z_{FP}(x) \;\approx\; \gamma_k \odot z_{LQ}(x) + \beta_k, \quad x \in C_k.
\label{eq:cat-affine}
\end{equation}

Where $\odot$ is the element-wise (Hadamard) product.
This can be derived in closed form by matching first- and second-order statistics of the two distributions:

\begin{align}
\mu_{LQ,k} &= \frac{1}{|C_k|}\sum_{x\in C_k} z_{LQ}(x), 
& \mu_{FP,k} &= \frac{1}{|C_k|}\sum_{x\in C_k} z_{FP}(x).
\end{align}
Here, $\mu_{LQ,k}$ and $\mu_{FP,k}$ denote the mean logits of the LQ and FP models over cluster $C_k$, respectively. \\
\begin{align}
\sigma_{LQ,k}^2 &= \frac{1}{|C_k|}\sum_{x\in C_k}\big(z_{LQ}(x)-\mu_{LQ,k}\big)^{\odot 2}.
\end{align}
 $\sigma_{LQ,k}^2$ denotes the variance of LQ logits within cluster $C_k$. \\

\begin{align}
\mathrm{cov}_{LQ,FP,k} &= \frac{1}{|C_k|}\sum_{x\in C_k}\big(z_{LQ}(x)-\mu_{LQ,k}\big)\odot\big(z_{FP}(x)-\mu_{FP,k}\big).
\end{align}
$\mathrm{cov}_{LQ,FP,k}$ denotes the element-wise covariance between LQ and FP logits over cluster $C_k$.
The elementwise affine parameters are then estimated as
\begin{equation}
\gamma_k = \frac{\mathrm{cov}_{LQ,FP,k}}{\sigma_{LQ,k}^2 + \epsilon}, 
\qquad 
\beta_k = \mu_{FP,k} - \gamma_k \odot \mu_{LQ,k},
\label{eq:gamma-beta}
\end{equation}
where division is elementwise and $\epsilon>0$ avoids division by zero.
Thus, for each cluster $k$, the transformation is obtained by aligning the cluster-wise mean and variance of quantized logits to those of FP, without requiring gradient optimization. 
We do not require backpropagation through the network to find $(\gamma_k,\beta_k)$. 
For each cluster $k$, we estimate $\gamma_k$ and  $\beta_k$ as parameters to minimize the error between $p_{LQ}$ and $p_{FP}$ using \cref{eq:cat-affine}.
 This black-box procedure fits the affine terms using only a small sample set, avoids gradient computations entirely, and naturally yields low-precision $\{\gamma_k\}_{k=1}^K$ and $\{\beta_k\}_{k=1}^K$ suitable for deployment.
Our two-stage pipeline first optimizes \eqref{eq:stage1} to refine quantization parameters using only the output-level KL loss. We then find $\gamma$ and $\beta$ through the CAT fitting phase. In inference, CAT requires only a single affine transformation per cluster assignment, introducing negligible overhead while substantially reducing the FP/LQ gap. We first assign the logits $z_{LQ}$ to a cluster $C_k$. 
We then apply the following $\alpha$-blended correction:
\[
\tilde{z} = (1 - \alpha)\,z_{\text{LQ}} \;+\; \alpha \,\big(\gamma_k \odot z_{\text{LQ}} + \beta_k\big),
\]
where $\alpha \in [0,1]$ controls the contribution of the original quantized output and the affine transformed one. 
The pseudocode for CAT fitting and inference is detailed in \cref{Section:CAT_alg}.
\section{Experiments}

\paragraph{Setup.} We use K-Means as the clustering algorithm.
We evaluate CAT on ImageNet-1K~\cite{ILSVRC15} using ResNet-18/50~\cite{He_2016_CVPR}, MobileNetV2~\cite{Sandler_2018_CVPR}, RegNetX-600MF/3.2GF~\cite{Radosavovic_2020_CVPR}, and MNasX2~\cite{Tan_2019_CVPR} under different LQ settings, denoted as $\{\text{weight bit-width}\}A\{\text{activation bit-width}\}$: W4A4, W2A4, W4A2, and W2A2. CAT is compared against strong PTQ baselines (ACIQ-Mix, LAPQ, Bit-Split, AdaRound, QDrop, PD-Quant) with identical hyperparameters to ensure fairness. Models are quantized channel-wise, calibrated with AdaRound (20k iterations, batch size 64, 1,024 samples), and follow prior work by keeping the last layer at 8-bit. CAT adopts the same calibration as PD-Quant, with KL temperature $0.4$ and learning rate $4\times 10^{-5}$. All experiments are run on NVIDIA L40 GPUs, repeated with three seeds, and we report the mean and standard deviation of Top-1 accuracy. We ablate CAT hyperparameters ($\alpha$, number of clusters (\# Clusters), PCA dimension, and clustering samples) to analyze their effect on performance.
\paragraph{Quantitative Comparison against State-of-the-Art}  
CAT provides advantages across all quantization settings and architectures (\cref{tab:pdquant_main}); however, the magnitude of improvements varies depending on the bit-width configuration and the network’s capacity. 
Overall, our results show that CAT provides the greatest benefits for networks with a larger gap between FP and LQ, such as MNasX2 (W2A2). In such settings, quantization errors accumulate heavily, and CAT’s cluster-specific affine correction substantially restores performance.  
With both weights and activations quantized to 2 bits (W2A2), CAT shows the best improvement. On ResNet-50, CAT achieves $58.08\%$, an improvement of $+1.05\%$ over PD-Quant. On MNasX2, CAT reaches $29.20\%$, outperforming PD-Quant by $+1.25\%$. Gains are also consistent across ResNet-18 ($+0.32\%$), MobileNetV2 ($+0.51\%$), and RegNetX-3.2GF ($+1.06\%$). Similarly, with 4-bit weights and 2-bit activations (W4A2), where the activation bottleneck is severe, CAT consistently delivers improvements. On ResNet-18, CAT yields $58.68\%$, exceeding PD-Quant by $+0.11\%$, while on MNasX2, CAT improves accuracy to $40.14\%$, a substantial $+0.71\%$ increase. Gains are also observed on MobileNetV2 ($+0.39\%$) and RegNetX-600MF ($+0.35\%$). These findings demonstrate that W2A2 and W4A2 are the most error-prone regimes, and CAT’s targeted corrections are especially effective at resolving their distorted feature representations.  

In another asymmetric setting, W2A4, CAT also provides consistent improvements. For instance, CAT improves over PD-Quant by $+0.19\%$ on ResNet-18, $+0.45\%$ on ResNet-50, and $+0.35\%$ on RegNetX-3.2GF. On MobileNetV2, CAT achieves $55.47\%$, improving upon PD-Quant by $+0.17\%$, while MNasX2 benefits from a $+0.64\%$ gain. Interestingly, W2A4 tends to produce more diverse outcomes compared to W2A2 or W4A2: while low-bit weights distort the learned filters and increase the risk of incorrect feature extraction, the higher activation precision (4-bit) preserves a broader dynamic range, enabling richer but more variable behavior than the severely compressed 2-bit activation cases.  
At higher precision (W4A4), where the discrepancy between full-precision (FP) and low-bit (LQ) networks is relatively small, CAT provides only marginal improvements since its corrections are designed to bridge larger gaps. In this regime, CAT achieves accuracy on par with or slightly better than prior methods. On ResNet-18, CAT obtains $69.18\%$, essentially matching PD-Quant ($69.14\%$), while on RegNetX-3.2GF, CAT improves to $76.61\%$, the best among all compared approaches. Similar trends are observed on MobileNetV2 ($68.22\%$) and ResNet-50 ($75.12\%$), confirming that when quantization noise is less severe, CAT closely mimics the FP network but cannot deliver substantial gains.
An important observation in this is that networks with a larger discrepancy between FP and LQ outcomes benefit the most from CAT in LQ settings. This effect arises because low-capacity models have limited redundancy to absorb quantization errors, making CAT’s correction particularly impactful.  
Across all architectures and bit-widths, CAT either matches or outperforms prior state-of-the-art PTQ methods. The improvements are most pronounced under LQ settings (W2A2, W4A2, W2A4) and on compact models, where quantization errors are most destructive. CAT achieves more effective reduction than plain affine mapping, thereby establishing a new state-of-the-art in PTQ. \cref{Section:Discussion} discusses a statistical analysis of CAT performance across the cross of model parameters and LQ settings.
\begin{table*}[t]
\centering
\caption{Top-1 accuracy (\%) on ImageNet-1K for various PTQ methods across architectures.}
\label{tab:pdquant_main}
\resizebox{\textwidth}{!}{%
\begin{tabular}{l c c c c c c c}
\toprule
\textbf{Methods} & \textbf{Bits (W/A)} & \textbf{ResNet-18} & \textbf{ResNet-50} & \textbf{MobileNetV2} & \textbf{RegNetX-600MF} & \textbf{RegNetX-3.2GF} & \textbf{MNasX2} \\
\midrule
Full Prec. & 32/32 & 71.01 & 76.63 & 72.62 & 73.52 & 78.46 & 76.52 \\
\midrule
ACIQ-Mix~\cite{NEURIPS2019_c0a62e13}   & \multirow{6}{*}{4/4} & 67.00 & 73.80 & --    & --    & --    & --    \\
LAPQ~\cite{nahshan2021loss}     &                       & 60.30 & 70.00 & 49.70 & 57.71 & 55.89 & 65.32 \\
Bit-Split~\cite{pmlr-v119-wang20c} &                       & 67.56 & 73.71 & --    & --    & --    & --    \\
AdaRound~\cite{pmlr-v119-nagel20a}  &                       & 67.96 & 73.88 & 61.52 & 68.20 & 73.85 & 68.86 \\
QDrop~\cite{wei2022qdrop} &        & 69.17 & 75.15 & 68.07 & 70.91 & 76.40 & 72.81 \\
PD-Quant~\cite{Liu_2023_CVPR} &                    & $69.14 \pm 0.10$ & $75.07 \pm 0.09$ &$68.18 \pm 0.02$& $70.96 \pm 0.03$ & $76.54 \pm 0.02$ & $73.24 \pm 0.02$ \\
\textbf{Ours} &                    & $\mathbf{69.18 \pm 0.09 }$& $\mathbf{75.12 \pm 0.07 }$ &  $\mathbf{68.22 \pm 0.01}$  & $\mathbf{70.98 \pm 0.01}$ &$\mathbf{76.61 \pm 0.02}$ & $\mathbf{73.31 \pm 0.05}$ \\
\midrule
LAPQ           & \multirow{4}{*}{2/4}  & 0.18 & 0.14 & 0.13 & 0.17 & 0.12 & 0.18 \\
Adaround       &                        & 0.11 & 0.12 & 0.15 & --   & --   & --   \\
QDrop &              & 64.57 & 70.09 & 53.37 & 63.18 & 71.96 & 63.23 \\
PD-Quant &                    & $65.10 \pm 0.02$ & $70.84 \pm 0.06$ & $55.30 \pm 0.24$ & $63.92 \pm 0.24$& $72.36 \pm 0.13$ & $63.32 \pm 0.24$ \\
\textbf{Ours} &                    & $\mathbf{65.26 \pm 0.06}$ & $\mathbf{71.29 \pm 0.02}$ & $\mathbf{55.47 \pm 0.22}$ & $\mathbf{64.2 \pm 0.28}$ & $\mathbf{72.71 \pm 0.12}$  & $\mathbf{63.96 \pm 0.32}$ \\
\midrule
QDrop & \multirow{2}{*}{4/2} & 57.56 & 63.26 & 17.30 & 49.73 & 62.79 & 34.12 \\
PD-Quant &                      & $58.57 \pm 0.18$ & $64.24 \pm 0.02$ & $20.14 \pm 0.52$ & $51.17 \pm 0.27$ & $62.68 \pm 0.08$ & $39.43 \pm 0.34$ \\
\textbf{Ours} &                      &  $\mathbf{58.68 \pm 0.15}$ & $\mathbf{64.38 \pm 0.06}$ & $\mathbf{20.53 \pm 0.53}$ & $\mathbf{51.52 \pm 0.26}$ &$\mathbf{63.03 \pm 0.05}$& $\mathbf{40.14 \pm 0.27}$\\
\midrule
QDrop & \multirow{2}{*}{2/2} & 51.42 & 55.45 & 10.28 & 39.01 & 54.38 & 23.59 \\
PD-Quant &                      &$52.87 \pm 0.03$ & $57.03 \pm 0.12$ & $13.65 \pm 0.63$ & $40.71 \pm 0.13$ & $55.08 \pm 0.13$ & $27.95 \pm 0.69$ \\
\textbf{Ours} &                      & $\mathbf{53.19 \pm 0.07}$ & $\mathbf{58.08 \pm 0.11}$ & $\mathbf{14.16 \pm 0.61}$ &$\mathbf{41.38 \pm 0.1}$ & $\mathbf{56.14 \pm 0.09}$ & $\mathbf{29.20 \pm 0.76}$ \\
\bottomrule
\multicolumn{8}{l}{\footnotesize{Note: PD-Quant results are reproduced from our runs.}}
\end{tabular}%
}
\end{table*}
\paragraph{Enhance PTQ methods with CAT}
\begin{figure*}[ht]
\centering

\begin{minipage}[ht]{0.35\linewidth}
\centering
\begin{tikzpicture}
\begin{axis}[
    width=\linewidth,
    height=4.0cm,
    ybar,
    bar width=0.4cm,
    ymin=0, ymax=1.1,
    ylabel={Average $\Delta$ Accuracy (\%)},
    xlabel={Quantization Setting},
    xlabel style={at={(axis description cs:0.5,1.05)},anchor=south},
    symbolic x coords={W2A2, W4A2, W2A4, W4A4},
    xtick=data,
    nodes near coords,
    nodes near coords align={vertical},
    every node near coord/.append style={
        font=\scriptsize,
        /pgf/number format/fixed,
        /pgf/number format/precision=1
    },
    grid=major,
    tick label style={font=\scriptsize},
    label style={font=\small},
    title style={font=\small}
]
\addplot[fill=blue!70] coordinates {(W2A2,0.9) (W4A2,0.5) (W2A4,0.3) (W4A4,0.1)};
\end{axis}
\end{tikzpicture}
\captionof{figure}{Average improvements $\Delta=(\text{CAT}-\text{Base})$ across quantization settings.}
\label{fig:avg-delta}
\end{minipage}%
\hfill
\begin{minipage}[ht]{0.63\linewidth}
\centering
\captionof{table}{ResNet-18 Top-1 Accuracy (\%) with/without CAT. $\Delta$ shows improvement (\upgreen{green}) or degradation (\downred{red}).}
\setlength{\tabcolsep}{3pt}
\renewcommand{\arraystretch}{1.05}
\resizebox{\linewidth}{!}{%
\begin{tabular}{c|cccc|cccc}
\toprule
 & \multicolumn{4}{c|}{\textbf{W4A4}} & \multicolumn{4}{c}{\textbf{W4A2}} \\
\cmidrule(lr){2-5}\cmidrule(lr){6-9}
 & Adaround & BRECQ & QDrop & PD-Quant 
 & Adaround & BRECQ & QDrop & PD-Quant \\
\midrule
Base 
 & 2.59   & 69.32 & 69.17 & 69.14 
 & 1.18 & 49.44 & 57.91 & 58.57 \\
+  CAT 
 & 2.58   & 69.35 & 69.24 & 69.18 
 & 1.14 & 50.49 & 58.12 & 58.68 \\
$\Delta$ 
 & \downred{-0.01}   & \upgreen{+0.03} & \upgreen{+0.07} & \upgreen{+0.04} 
 & \downred{-0.04} & \upgreen{+1.05} & \upgreen{+0.21} & \upgreen{+0.11} \\
\midrule
 & \multicolumn{4}{c|}{\textbf{W2A4}} & \multicolumn{4}{c}{\textbf{W2A2}} \\
\cmidrule(lr){2-5}\cmidrule(lr){6-9}
 & Adaround & BRECQ & QDrop & PD-Quant 
 & Adaround & BRECQ & QDrop & PD-Quant \\
\midrule
Base 
 & 18.48 & 62.69 & 64.52 & 65.10 
 & 2.59  & 40.78 & 51.47 & 52.86 \\
+  CAT 
 & 22.04 & 62.67 & 64.76 & 65.35 
 & 2.58  & 41.75 & 51.76 & 53.19 \\
$\Delta$ 
 & \upgreen{+3.56} & \downred{-0.02} & \upgreen{+0.24} & \upgreen{+0.25} 
 & \downred{-0.01} & \upgreen{+0.97} & \upgreen{+0.29} & \upgreen{+0.32} \\
\bottomrule
\end{tabular}}%

\label{tab:quant_results_diff_square}
\end{minipage}

\end{figure*}

~\cref{tab:quant_results_diff_square}~ (More comprehensive results are provided in Appendix~\cref{tab:quant_results_diff_app}) provides a direct comparison of multiple PTQ baselines with and without the proposed CAT correction across different architectures and bit-width settings. 
The values in parentheses indicate the performance difference $\Delta = (\text{CAT} - \text{Base})$, where green arrows ($\uparrow$) mark improvements and red arrows ($\downarrow$) indicate degradations.  
Across nearly all architectures and methods, CAT consistently enhances performance. The improvements are particularly pronounced in LQ settings (W2A2 and W4A2), where quantization errors are most severe. For example, CAT boosts PD-Quant on ResNet-50 (W2A2) by $+1.21\%$, on RegNetX-3.2GF (W2A2) by $+1.01\%$, and on MNasX2 (W2A2) by $+1.24\%$. Similarly, for W4A2, CAT improves QDrop on MNasX2 by more than $+1\%$ and PD-Quant on MobileNetV2 by $+0.37\%$. These gains confirm that CAT is most effective in highly error-prone regimes, where its cluster-aware affine correction recovers distorted representations.  
In asymmetric quantization (W2A4), CAT again provides steady improvements, albeit with smaller margins (e.g., ResNet-50 $+0.42\%$, RegNetX-3.2GF $+0.45\%$), while in higher-precision settings (W4A4), the effect is marginal but consistently non-negative, reflecting that the baseline already closely approximates full-precision.  
An important observation is that smaller-capacity architectures exhibit disproportionately large gains from CAT, providing new evidence that networks with a large FP–LQ gap benefit the most in LQ settings. For instance, LQ MNasX2 gains more than $+1\%$, demonstrating that CAT is particularly valuable for compact models that lack redundancy to absorb quantization errors.  
Overall, \cref{tab:quant_results_diff_square} highlights CAT’s robustness as a drop-in enhancement: it supplements diverse PTQ baselines (Adaround, BRECQ, Qdrop, PD-Quant) consistently yielding improved accuracy while never severely degrading performance. This consistency demonstrates CAT’s generality and compatibility with existing PTQ pipelines.  
\cref{fig:avg-delta} shows the average improvement of CAT over the Base method. The largest gain is observed for W2A2 quantization (0.9\%) across all PTQ methods and models. The second-highest improvement occurs for W4A2 (0.5\%), highlighting the effectiveness of CAT under very low-bit activation quantization. For W2A4 and W4A4, the accuracy increases by 0.3\% and 0.1\%, respectively, demonstrating the generalization capability of CAT. Notably, 2-bit activation settings benefit the most from CAT, since their severely limited precision reduces the ability to preserve information entropy, making them more reliant on CAT’s error reduction mechanism. \cref{Section:ViT} represents LQ ViT error reduction using CAT.

\section{Ablation Study}
\label{section:ablation}

\begin{wrapfigure}{r}{0.35\linewidth} 
\centering
\begin{tikzpicture}
\begin{axis}[
    title={W2A2},
    ymin=52.3, ymax=53.3,
    ylabel={Top-1 (\%)},
    xlabel={$\alpha$},
    width=1\linewidth,
    height=3.8cm,
    grid=major,
    legend pos=south west,
    legend cell align={left},
    tick label style={font=\small},
    label style={font=\small},
    title style={font=\small},
    legend style={font=\small}
]

\addplot[blue, mark=*] coordinates {
  (0.1,53.05) (0.2,53.12) (0.3,53.16) (0.4,53.19) (0.5,53.15)
  (0.6,53.05) (0.7,52.89) (0.8,52.75) (0.9,52.63) (1.0,52.41)
};
\addlegendentry{Mean}

\addplot[name path=W2A2upper, draw=none] coordinates {
  (0.1,53.09) (0.2,53.13) (0.3,53.16) (0.4,53.26) (0.5,53.26)
  (0.6,53.16) (0.7,52.99) (0.8,52.91) (0.9,52.74) (1.0,52.50)
};
\addplot[name path=W2A2lower, draw=none] coordinates {
  (0.1,53.01) (0.2,53.11) (0.3,53.16) (0.4,53.12) (0.5,53.04)
  (0.6,52.94) (0.7,52.79) (0.8,52.59) (0.9,52.52) (1.0,52.32)
};

\addplot[blue, opacity=0.15] fill between[of=W2A2upper and W2A2lower];
\addlegendentry{$\pm 1\sigma$}

\end{axis}
\end{tikzpicture}
\caption{Top-1 accuracy for W2A2 across different $\alpha$ values with mean and $\pm 1\sigma$ range.}
\label{fig:w2a2}
\end{wrapfigure}
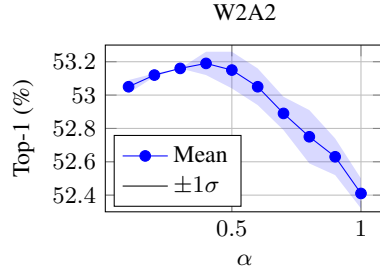

\paragraph{Ablation on Blending Coefficient $\alpha$.} 
This study ablates the effect of the blending coefficient $\alpha$, where $\alpha=0$ corresponds to using only PTQ logits without any CAT correction, and $\alpha=1$ corresponds to relying entirely on CAT-corrected logits without involving the original PTQ. As shown in Fig.~\ref{fig:w2a2}, varying $\alpha$ directly influences the final accuracy across different bit-width regimes (The comprehensive ablation of Blending Coefficient $\alpha$ for different LQ settings and architectures provided in~\cref{Section:app_alpha}). 

For the 2-bit activation quantization (W2A2 and W4A2), moderate blending ($\alpha \approx 0.3$--$0.4$) consistently provides the highest performance, indicating that CAT is most effective when used as a supplement to the baseline quantized outputs rather than a full replacement.
In the asymmetric regime (W2A4), accuracy is relatively stable across a broad range of $\alpha$, reflecting that higher-precision activations reduce sensitivity to blending, although small gains are still observed around $\alpha \approx 0.4$. At higher precision (W4A4), the performance is nearly flat across all values of $\alpha$, as the quantized model is already close to full-precision accuracy.  
Overall, these results demonstrate that CAT reliably improves PTQ performance, especially in LQ settings, but that using CAT alone ($\alpha=1$) is suboptimal. 
\begin{wrapfigure}{r}{0.55\linewidth}
\centering
\begin{tikzpicture}
\begin{axis}[
    title={W2A4},
    ymin=65.1, ymax=65.35,
    xlabel={\# Clusters},
    ylabel={Top-1 (\%)},
    width=\linewidth,
    height=4cm,
    grid=major,
    legend pos=south west,
    legend cell align={left},
    tick label style={font=\small},
    label style={font=\small},
    title style={font=\small},
    legend style={font=\small}
]

\addplot[green!60!black, mark=triangle*] coordinates {
  (1,65.22) (8,65.24) (16,65.24) (24,65.23) (32,65.23)
  (48,65.24) (64,65.26) (96,65.23) (128,65.25) (192,65.24) (256,65.21)
};
\addlegendentry{Mean}

\addplot[name path=W2A4_upper, draw=none] coordinates {
  (1,65.28) (8,65.29) (16,65.31) (24,65.31) (32,65.33)
  (48,65.24) (64,65.32) (96,65.25) (128,65.31) (192,65.31) (256,65.24)
};
\addplot[name path=W2A4_lower, draw=none] coordinates {
  (1,65.16) (8,65.19) (16,65.17) (24,65.15) (32,65.13)
  (48,65.24) (64,65.20) (96,65.21) (128,65.19) (192,65.17) (256,65.18)
};

\addplot[green!60!black, opacity=0.15] fill between[of=W2A4_upper and W2A4_lower];
\addlegendentry{$\pm 1\sigma$}

\end{axis}
\end{tikzpicture}
\caption{Top-1 accuracy for W2A4 across different number of Clusters (mean $\pm 1\sigma$).}
\label{fig:w2a4_clusters}

\end{wrapfigure}
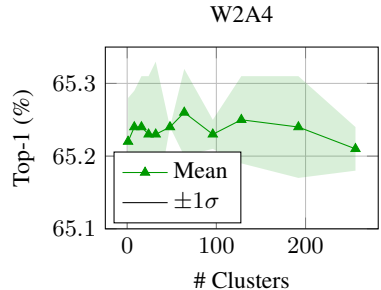
\paragraph{Ablation on the Number of Clusters.} 
We next ablate the influence of the number of clusters used in CAT, as shown in Fig.~\ref{fig:w2a4_clusters} for ResNet-18 W2A4 with PCA dimension fixed to $50$ and blending coefficient $\alpha=0.6$ (The comprehensive ablation of the number of clusters for different LQ settings and architectures is provided in~\cref{Section_app_cluster}).
For W2A2, performance is highest with a small number of clusters and gradually declines as the cluster count increases. This suggests that in extremely quantized regimes, a compact cluster partition provides stable corrections, while too many clusters lead to overfitting of noise in the heavily distorted logit space.  
For W4A2, a similar trend is observed: accuracy peaks at very low cluster counts (1–8 clusters) and then decreases slowly with more clusters, indicating diminishing returns once the coarse logit structure has been captured.  
In the asymmetric case W2A4, accuracy remains largely stable across a broad range of cluster counts, confirming that higher-precision activations mitigate sensitivity to clustering granularity.  
At higher precision (W4A4), accuracy is nearly unaffected by number of clusters, as the quantized logits already closely approximate the full-precision distribution.  
Overall, these results demonstrate that the optimal number of clusters is inherently fitted to the nature of quantization precision: lower precision reduces diversity in the logit space and thus favors fewer clusters, whereas higher precision allows richer structures that can benefit from larger cluster counts. This further suggests that clustering is not only a useful but also an essential component of the affine transformation, enabling it to adaptively reduce quantization errors according to the underlying representation capacity of the quantized network. 

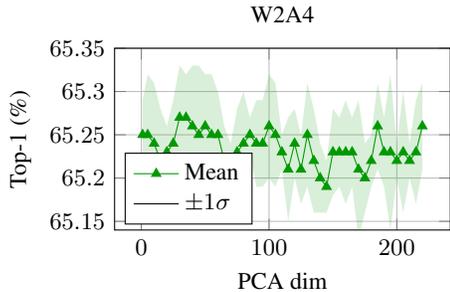
\begin{wrapfigure}{r}{0.4\linewidth}
\centering
\begin{tikzpicture}
\begin{axis}[
    title={W2A4},
    ymin=65.14, ymax=65.35,
    xlabel={PCA dim},
    ylabel={Top-1 (\%)},
    width=\linewidth,
    height=4cm,
    grid=major,
    legend pos=south west,
    legend cell align={left},
    tick label style={font=\small},
    label style={font=\small},
    title style={font=\small},
    legend style={font=\small}
]

\addplot[green!60!black, mark=triangle*] coordinates {
  (1,65.25) (5,65.25) (10,65.24) (15,65.22) (20,65.23)
  (25,65.24) (30,65.27) (35,65.27) (40,65.26) (45,65.25)
  (50,65.26) (55,65.25) (60,65.25) (65,65.22) (70,65.22)
  (75,65.23) (80,65.24) (85,65.25) (90,65.24) (95,65.24)
  (100,65.26) (105,65.25) (110,65.23) (115,65.21) (120,65.24)
  (125,65.21) (130,65.25) (135,65.22) (140,65.20) (145,65.19)
  (150,65.23) (155,65.23) (160,65.23) (165,65.23) (170,65.21)
  (175,65.20) (180,65.22) (185,65.26) (190,65.23) (195,65.23)
  (200,65.22) (205,65.23) (210,65.22) (215,65.23) (220,65.26)
};
\addlegendentry{Mean}

\addplot[name path=W2A4upper, draw=none] coordinates {
  (1,65.29) (5,65.32) (10,65.31) (15,65.28) (20,65.26)
  (25,65.29) (30,65.33) (35,65.32) (40,65.33) (45,65.33)
  (50,65.32) (55,65.32) (60,65.30) (65,65.26) (70,65.24)
  (75,65.28) (80,65.30) (85,65.27) (90,65.29) (95,65.29)
  (100,65.32) (105,65.31) (110,65.25) (115,65.27) (120,65.28)
  (125,65.25) (130,65.31) (135,65.28) (140,65.24) (145,65.22)
  (150,65.28) (155,65.27) (160,65.29) (165,65.27) (170,65.29)
  (175,65.24) (180,65.25) (185,65.31) (190,65.27) (195,65.32)
  (200,65.24) (205,65.30) (210,65.24) (215,65.29) (220,65.31)
};
\addplot[name path=W2A4lower, draw=none] coordinates {
  (1,65.21) (5,65.18) (10,65.17) (15,65.16) (20,65.20)
  (25,65.19) (30,65.21) (35,65.22) (40,65.19) (45,65.17)
  (50,65.20) (55,65.18) (60,65.20) (65,65.18) (70,65.20)
  (75,65.18) (80,65.18) (85,65.23) (90,65.19) (95,65.19)
  (100,65.20) (105,65.19) (110,65.21) (115,65.15) (120,65.20)
  (125,65.17) (130,65.19) (135,65.16) (140,65.16) (145,65.16)
  (150,65.18) (155,65.19) (160,65.17) (165,65.19) (170,65.13)
  (175,65.16) (180,65.19) (185,65.21) (190,65.19) (195,65.14)
  (200,65.20) (205,65.16) (210,65.20) (215,65.17) (220,65.21)
};

\addplot[green!60!black, opacity=0.15] fill between[of=W2A4upper and W2A4lower];
\addlegendentry{$\pm 1\sigma$}

\end{axis}
\end{tikzpicture}
\caption{Top-1 accuracy for W2A4 across different PCA dim (mean $\pm 1\sigma$).}
\label{fig:w2a4_pca}
\end{wrapfigure}

\paragraph{Ablation on PCA Dimension $k$.} 
We further study the effect of the PCA dimension $k$ used to reduce the logit space before clustering, as shown in Fig.~\ref{fig:w2a4_pca} for ResNet-18 under the W2A4 quantization setting (The comprehensive ablation of the PCA dimension for different LQ settings and architectures is provided in~\cref{Section:app_pca}).

For W2A2, performance is maximized when $k$ is very small (1–5) and gradually decreases as $k$ increases. This indicates that in extremely quantized networks, only the coarse structure of the logits can be reliably captured, and projecting onto a compact subspace avoids fitting noise.  

For W4A2, a similar but weaker trend is observed: accuracy peaks when $k$ is kept small ($\leq 10$) and remains stable for moderate values before slightly degrading with very large $k$.  
In the asymmetric W2A4 setting, accuracy is largely stable across the full range of $k$, suggesting that higher activation precision preserves sufficient feature variability to tolerate richer subspaces without significant overfitting.  

At higher precision (W4A4), accuracy is almost invariant to the PCA dimension, as the quantized logits already closely match the full-precision distribution and PCA reduction plays only a minor role.  
 Results indicate that the optimal PCA dimension depends on the severity of quantization: lower-precision networks benefit from aggressive dimensionality reduction that filters out noise, while higher-precision networks can afford larger $k$ without degradation. This highlights PCA as an essential regularization step that adapts the clustering space to the effective diversity of the quantized representations. 
\label{sec:ablation_samples}
\begin{wraptable}{l}{0.45\linewidth}
\centering
\caption{Top-1 accuracy for W2A2 across different numbers of samples CAT fitting.}
\label{tab:w2a2_calib}
\begin{tabular}{r c}
\toprule
\# Samples & Top-1 (\%) \\
\midrule
10     & 46.48 \\
50     & 51.79 \\
100    & 52.49 \\
500    & 53.02 \\
1K   & 53.04 \\
10K &53.15\\
100K&53.14\\
\bottomrule
\end{tabular}
\end{wraptable}
\paragraph{Ablation on CAT fitting sample size.}
\cref{tab:w2a2_calib} studies the impact of the number of samples for the fitting of CAT on ResNet-18 on the W2A2 setting (The comprehensive ablation of sample size on CAT's performance for different ResNet-18 LQ settings is provided in ~\cref{Section:smaples_app}). We observe a consistent monotonic trend with diminishing returns: accuracy improves rapidly when increasing samples from $10$ to $500$--$1000$, and then plateaus.
Under W2A2, the mean Top-1 rises from $46.48\%$ (10 samples) to $53.04\%$ (1{,}000), a gain of $\approx\!+6.6$ points, with only marginal changes beyond 1{,}000 (e.g., $53.15\%$ at 100{,}000). W4A2 shows a similar but smaller effect ($53.67\%\!\rightarrow\!58.54\%$, $\approx\!+4.9$). In contrast, higher-precision settings are less sensitive: W2A4 improves by $\approx\!+2.7$ ($62.60\%\!\rightarrow\!65.33\%$), while W4A4 gains only $\approx\!+1.1$ ($68.13\%\!\rightarrow\!69.26\%$).
The shaded $\pm 1\sigma$ bands narrow as sample size grows, indicating more stable calibration and reduced run-to-run variability, especially pronounced in W2A2/W4A2.
A calibration set of \mbox{$\sim$500--1{,}000} samples captures nearly all attainable gains across regimes, striking a favorable accuracy/cost trade-off. Using $\geq 10$k samples yields negligible additional improvements, particularly in W4A4, where performance is near its ceiling.
\section{Limitations}

While CAT consistently improves accuracy over PTQ baselines, it introduces additional parameters due to the clustering step and the cluster-specific affine corrections. In particular, PTQ baselines do not maintain any auxiliary parameters beyond the quantized model itself, whereas CAT requires storing the clustering model and the affine coefficients $(\gamma, \beta)$ for each cluster.
Here, we analyse additional parameters by considering the K-Means clustering algorithm with $k$ means.
This overhead grows linearly with the number of clusters $k$ and the logit dimensionality $d$ (equal to the number of classes). 
The additional parameter count of CAT can therefore be approximated as $\text{CAT}_{\#\text{params}} = (k \times d \;\text{for cluster-wise affine coefficients}) + (k \times d \;\text{for $k$-means centroids})$.
For example, for ResNet-18 with $ 11.6$ million parameters, the number of model parameters in addition to CAT with $k=50$ and $d=1000$ is $11,600,000 + 2*(50*1000) = 11,700,000$.
In this example, CAT adds only $\sim0.9\%$ parameter overhead relative to the baseline model. Nevertheless, this extra storage may be undesirable for extremely resource-constrained deployments, which we identify as a limitation of CAT compared to PTQ baselines.
\section{Conclusion}
We studied the gap between full-precision (FP) and low-bit quantized (LQ) networks, focusing on reducing quantization errors through affine transformations. Our initial findings showed that a plain affine mapping with a uniform parameter set can even worsen PTQ performance. To overcome this, we introduced Cluster-based Affine Transformation (CAT), which leverages the clusterability of quantized logits and applies cluster-specific affine corrections. CAT consistently improves accuracy under low-bit settings with only negligible parameter overhead, achieving state-of-the-art performance on ImageNet-1K. These results highlight that cluster-aware error reduction is a powerful and generalizable strategy for enhancing PTQ, particularly in extremely low-bit regimes.
\newpage
\newpage
\bibliography{iclr2026_conference}
\bibliographystyle{iclr2026_conference}
\clearpage
\appendix
\section*{Appendix for CAT}
\section{Quantization Notation}
\label{section:qn}
To better ground the discussion of CAT, we start by defining the notation used in neural network quantization.
Quantization is the process of compacting a neural network by reducing the precision of full-precision weights and activations, considering the given bit-width allowance.
For notational simplicity, we consider the quantization of a scalar value.
Let $f \in \mathbb{R}$ denote a floating-point value (weight or activation) with real range 
\[
f_{\min} = \min(f), 
\qquad 
f_{\max} = \max(f).
\]
For a given bit-width $b$, the representable integer range is defined as
\[
q_{\min} = -2^{b-1}, \quad q_{\max} = 2^{b-1} - 1,
\]
where $q_{\min}$ and $q_{\max}$ denote the minimum and maximum values of the quantized representation.
Assuming \textit{scale factor} $\delta$ and \textit{zero-point} $z$ as parameters to convert to convert floating-point value $f$ to its quantized value $q$ with $N$-bit representation, then we have, 

\[
\delta = \frac{f_{\max} - f_{\min}}{q_{\max} - q_{\min}}, 
\qquad 
z = \left\lfloor q_{\min} - \frac{f_{\min}}{\delta} \right\rceil.
\]

The quantization and dequantization mappings are then given by
\[
q = \text{clip}\!\left( \left\lfloor \frac{f}{\delta} \right\rceil + z, \; q_{\min}, \; q_{\max} \right),
\]
\[
\hat{f} = \delta \cdot (q - z),
\]
 $\hat{f}$ is the reconstructed quantized value, and 
$\text{clip}(\cdot)$ ensures $q \in [q_{\min}, q_{\max}]$.
Weights quantization is simpler since their values remain constant during PTQ, whereas activations quantization requires estimating their minimum and maximum.
In post-training quantization (PTQ), $f_{\min}$ and $f_{\max}$ are estimated from a small calibration dataset.
Different strategies include \emph{min--max calibration}, which directly uses observed extrema, 
\emph{percentile or clipping methods}, which discard outliers, and \emph{optimization-based approaches} such as 
AdaRound or BRECQ, which refine $\delta$ (and sometimes $z$) by minimizing reconstruction error between 
FP and quantized outputs.
We denote a quantization configuration by $WbAb$, where $Wb$ and $Ab$ specify the bit-widths used for 
weights and activations, respectively (e.g., $W2A4$ indicates 2-bit weights with 4-bit activations).

\section{Cluster-based Affine Transformation (CAT) Algorithm}
\label{Section:CAT_alg}

\begin{figure}[H]
\centering
\begin{algorithm}[H]
\caption{CAT Fitting}
\begin{algorithmic}[1]
\Require $\,\text{all\_q} \in \mathbb{R}^{N\times C}$, $\,\text{all\_fp} \in \mathbb{R}^{N\times C}$, $\,Clusters$, pca\_dim
\Ensure \textsc{Cluster-Model}, \textsc{PCA}, $\{\gamma_c\}_{c=1}^{K}$, $\{\beta_c\}_{c=1}^{K}$

\State $q\_features \gets \textsc{PCA}.decomposition(\text{pca\_dim},\, \text{all\_q})$
\State $\text{cluster\_ids} \gets \textsc{Cluster-Model}.fit(Clusters,q\_features)$
\For{$cid \gets 1 \to Clusters$}
    \State $\text{idxs} \gets \{ i \mid \text{cluster\_ids}[i] = cid \}$
    \If{$|\text{idxs}| = 0$}
        \State $\gamma\_{cid} \gets \mathbf{1}_C$;\ \ $\beta\_{cid} \gets \mathbf{0}_C$
        \State \textbf{continue}
    \EndIf
    \State $Q \gets \text{all\_q}[\text{idxs}]$,\quad $F \gets \text{all\_fp}[\text{idxs}]$
    \State $\mu_Q \gets \text{Mean}(Q)$,\quad $\mu_F \gets \text{Mean}(F)$
    \State $\sigma_Q^2 \gets \max(\text{Var}(Q), 10^{-8})$
    \State $\gamma_{cid} \gets \dfrac{\text{Mean}[(Q-\mu_Q)\odot(F-\mu_F)]}{\sigma_Q^2}$
    \State $\beta_{cid} \gets \mu_F - \gamma_{cid} \odot \mu_Q$
\EndFor
\State \Return \textsc{Cluster-Model}, $\{\gamma_c\}, \{\beta_c\}$
\end{algorithmic}
\end{algorithm}

\vspace{-0.5em}

\begin{algorithm}[H]
\caption{CAT Inferencing}
\begin{algorithmic}[1]
\Require $q\_logits$, \textsc{Cluster-Model}, $\{\gamma_c\}, \{\beta_c\}$, $\alpha$
\Ensure corrected logits, PCA
\State $q\_{feature} \gets \textsc{PCA}.decomposition(q\_{np})$
\State $cluster\_ids \gets \textsc{Cluster-Model}.predict(q\_{feature})$
\State $corrected \gets [\,]$
\For{$i = 1 \to |q\_logits|$}
    \State $q \gets q\_logits[i]$
    \State $cid \gets cluster\_ids[i]$
    \State $\gamma \gets \gamma\_{cid}$, $\beta \gets \beta\_{cid}$
    \State $affine\_corrected \gets q \cdot \gamma + \beta$
    \State $blended \gets q + \alpha \cdot (affine\_corrected - q)$
    \State Append $blended$ to $corrected$
\EndFor
\State \Return corrected
\end{algorithmic}
\end{algorithm}
\end{figure}
\section{Enhance PTQ methods with CAT}
\begin{table}[h!]
\centering
\caption{Comparison of Quantization Methods (with/without CAT) across Architectures on ImageNet-1K. 
Values in parentheses show $\Delta = (\text{CAT} - \text{Base})$. 
\upgreen{Green}: CAT improves. \downred{Red}: CAT worse. The comprehensive description of thís table's results is provided in \cref{sec:experiments}}
\label{tab:quant_results_diff_app}
\resizebox{\textwidth}{!}{%
\begin{tabular}{c l c c c c c c c}
\toprule
\textbf{Bits (W/A)} & \textbf{Methods} & \textbf{ResNet-18} & \textbf{ResNet-50} & \textbf{MobileNetV2} & \textbf{RegNetX-600MF} & \textbf{RegNetX-3.2GF} & \textbf{MNasX2} \\
\midrule
\multirow{2}{*}{4/4} 
 & Adaround            & 2.59 & 1.57 & 24.22 & -- & 0.95 & 0.68 \\
 & + CAT ( $\Delta$)    & 2.58 (\downred{-0.01}) & 1.57 (0.00) & 28.51 (\upgreen{+4.29}) &  & 0.98 (\upgreen{+0.03}) & 0.75 (\upgreen{+0.07}) \\
\midrule
\multirow{2}{*}{4/4} 
 & BRECQ               & 69.32 & 74.45 & 4.56 & 69.67 & 74.57 & 66.26 \\
 & + CAT ( $\Delta$)    & 69.35 (\upgreen{+0.03}) & 74.48 (\upgreen{+0.03}) & 6.48 (\upgreen{+1.92}) & 69.72 (\upgreen{+0.05}) & 74.68 (\upgreen{+0.11}) & 66.57 (\upgreen{+0.31}) \\
\midrule
\multirow{2}{*}{4/4} 
 & Qdrop               & 69.17 & 75.08 & 67.86 & 70.78 & 76.31 & 72.91 \\
 & + CAT ( $\Delta$)    & 69.24 (\upgreen{+0.07}) & 75.18 (\upgreen{+0.10}) & 68.01 (\upgreen{+0.15}) & 70.89 (\upgreen{+0.11}) & 76.46 (\upgreen{+0.15}) & 73.04 (\upgreen{+0.13}) \\
\midrule
\multirow{2}{*}{4/4} 
 & PD-Quant            & 69.17 & -- & 68.20 & 70.9 & -- & -- \\
 & + CAT ( $\Delta$)    & 69.27 (\upgreen{+0.10}) & -- & 68.28 (\upgreen{+0.08}) & 71.08 (\upgreen{+0.18}) & -- & -- \\
\midrule
\multirow{2}{*}{4/2} 
 & Adaround            & 1.18 & -- & 24.22 & -- & -- & 0.15 \\
 & + CAT ($\Delta$)    & 1.14 (\downred{-0.04}) & -- & 28.51 (\upgreen{+4.29}) & -- & -- & 0.17 (\upgreen{+0.02}) \\
\midrule
\multirow{2}{*}{4/2} 
 & BRECQ               & 49.44 & 36.59 & 0.23 & 6.79 & 6.48 & 0.82 \\
 & + CAT ($\Delta$)    & 50.49 (\upgreen{+1.05}) & 38.47 (\upgreen{+1.88}) & 0.43 (\upgreen{+0.20}) & 11.22 (\upgreen{+4.43}) & 10.16 (\upgreen{+3.68}) & 1.73 (\upgreen{+0.91}) \\
\midrule
\multirow{2}{*}{4/2} 
 & Qdrop               & 57.91 & 63.15 & 16.56 & 49.83 & 62.15 & 30.43 \\
 & + CAT ($\Delta$)    & 58.12 (\upgreen{+0.21}) & 63.54 (\upgreen{+0.39}) & 16.93 (\upgreen{+0.37}) & 50.05 (\upgreen{+0.22}) & 62.70 (\upgreen{+0.55}) & 31.44 (\upgreen{+1.01}) \\
\midrule
\multirow{2}{*}{4/2} 
 & PD-Quant            & 58.66 & 64.26 & 19.84 & 51.11 & 62.63 & 39.59 \\
 & + CAT ($\Delta$)    & 58.80 (\upgreen{+0.14}) & 64.49 (\upgreen{+0.23}) & 20.21 (\upgreen{+0.37}) & 51.52 (\upgreen{+0.41}) & 63.05 (\upgreen{+0.42}) & 40.32 (\upgreen{+0.73}) \\
\midrule
\multirow{2}{*}{2/4} 
 & Adaround            & 18.48 & -- & 0.36 & -- & -- & 0.00 \\
 & + CAT ($\Delta$)    & 22.04 (\upgreen{+3.56}) & -- & 0.74 (\upgreen{+0.38}) & -- & -- & 0.00 (0.00) \\
\midrule
\multirow{2}{*}{2/4} 
 & BRECQ               & 62.69 & 64.23 & 0.12 & 51.57 & 64.39 & 0.23 \\
 & + CAT ($\Delta$)    & 62.67 (\downred{-0.02}) & 64.44 (\upgreen{+0.21}) & 0.20 (\upgreen{+0.08}) & 51.65 (\upgreen{+0.08}) & 64.57 (\upgreen{+0.18}) & 0.32 (\upgreen{+0.09}) \\
\midrule
\multirow{2}{*}{2/4} 
 & Qdrop               & 64.52 & 69.95 & 54.10 & 63.17 & 71.59 & 63.39 \\
 & + CAT ($\Delta$)    & 64.76 (\upgreen{+0.24}) & 70.50 (\upgreen{+0.55}) & 54.41 (\upgreen{+0.31}) & 63.50 (\upgreen{+0.33}) & 72.11 (\upgreen{+0.52}) & 64.09 (\upgreen{+0.70}) \\
\midrule
\multirow{2}{*}{2/4} 
 & PD-Quant            & 65.10 & 70.87 & 55.58 & 63.90 & 72.21 & 63.06 \\
 & + CAT ($\Delta$)    & 65.35 (\upgreen{+0.25}) & 71.29 (\upgreen{+0.42}) & 55.72 (\upgreen{+0.14}) & 64.31 (\upgreen{+0.41}) & 72.66 (\upgreen{+0.45}) & 63.82 (\upgreen{+0.76}) \\
\midrule
\multirow{2}{*}{2/2} 
 & Adaround            & 2.59 & -- & -- & -- & -- & -- \\
 & + CAT ($\Delta$)    & 2.58 (\downred{-0.01}) & -- & -- & -- & -- & -- \\
\midrule
\multirow{2}{*}{2/2} 
 & BRECQ               & 40.78 & 13.78 & 0.10 & 0.68 & 1.73 & 0.12 \\
 & + CAT ($\Delta$)    & 41.75 (\upgreen{+0.97}) & 16.69 (\upgreen{+2.91}) & 0.16 (\upgreen{+0.06}) & 2.03 (\upgreen{+1.35}) & 3.45 (\upgreen{+1.72}) & 0.18 (\upgreen{+0.06}) \\
\midrule
\multirow{2}{*}{2/2} 
 & Qdrop               & 51.47 & 55.46 & 11.81 & 38.92 & 54.12 & 23.04 \\
 & + CAT ($\Delta$)    & 51.76 (\upgreen{+0.29}) & 56.81 (\upgreen{+1.35}) & 12.25 (\upgreen{+0.44}) & 39.75 (\upgreen{+0.83}) & 55.26 (\upgreen{+1.14}) & 24.11 (\upgreen{+1.07}) \\
\midrule
\multirow{2}{*}{2/2} 
 & PD-Quant            & 52.86 & 57.12 & 13.92 & 40.67 & 55.23 & 28.09 \\
 & + CAT ($\Delta$)    & 53.18 (\upgreen{+0.32}) & 58.33 (\upgreen{+1.21}) & 14.33 (\upgreen{+0.41}) & 41.32 (\upgreen{+0.65}) & 56.24 (\upgreen{+1.01}) & 29.33 (\upgreen{+1.24}) \\
\bottomrule
\end{tabular}}
\end{table}
\newpage
\section{Ablation of $\alpha$ for different Architectures}
\label{Section:app_alpha}
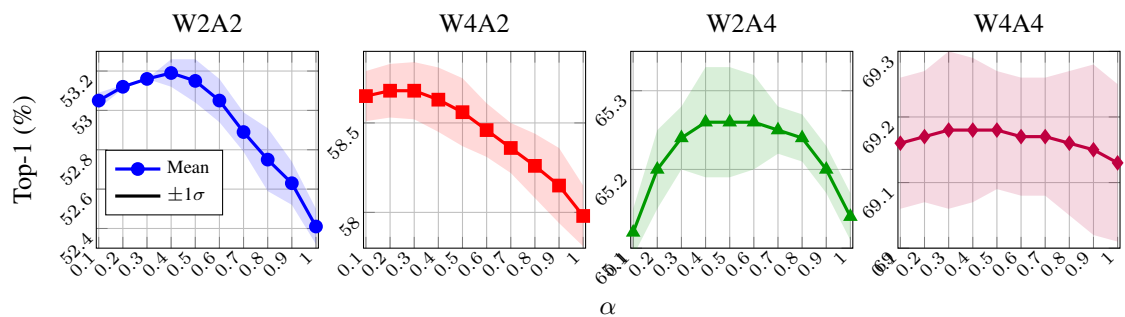
\begin{figure}[ht]
\centering
\begin{tikzpicture}
\begin{groupplot}[
  group style={group size=4 by 1, horizontal sep=0.6cm},
  width=0.3\linewidth,
  height=4.2cm,
  grid=major,
  tick label style={font=\scriptsize,/pgf/number format/fixed},
  every axis plot/.append style={very thick},
  xmin=0.09, xmax=1.01,
  xtick={0.1,0.2,0.3,0.4,0.5,0.6,0.7,0.8,0.9,1.0},
  scaled ticks=false,
  xticklabel style={rotate=45, anchor=east},
  yticklabel style={rotate=45, anchor=east},
  title style={yshift=-0.5ex},
  legend style={cells={anchor=west}, font=\scriptsize, at={(0.04,0.5)}, anchor=north west}
]

\nextgroupplot[title={W2A2}, ymin=52.3, ymax=53.3, ylabel={Top-1 (\%)}]

\addplot[blue, mark=*] coordinates {
  (0.1,53.05) (0.2,53.12) (0.3,53.16) (0.4,53.19) (0.5,53.15)
  (0.6,53.05) (0.7,52.89) (0.8,52.75) (0.9,52.63) (1.0,52.41)
};
\addlegendentry{Mean}

\addplot[name path=W2A2upper, draw=none] coordinates {
  (0.1,53.09) (0.2,53.13) (0.3,53.16) (0.4,53.26) (0.5,53.26)
  (0.6,53.16) (0.7,52.99) (0.8,52.91) (0.9,52.74) (1.0,52.50)
};
\addplot[name path=W2A2lower, draw=none] coordinates {
  (0.1,53.01) (0.2,53.11) (0.3,53.16) (0.4,53.12) (0.5,53.04)
  (0.6,52.94) (0.7,52.79) (0.8,52.59) (0.9,52.52) (1.0,52.32)
};

\addplot[blue, opacity=0.15] fill between[of=W2A2upper and W2A2lower];
\addlegendentry{$\pm$1$\sigma$}

\nextgroupplot[title={W4A2}, ymin=57.8, ymax=58.9]

\addplot[red, mark=square*] coordinates {
  (0.1,58.65) (0.2,58.68) (0.3,58.68) (0.4,58.63) (0.5,58.56)
  (0.6,58.46) (0.7,58.36) (0.8,58.26) (0.9,58.15) (1.0,57.98)
};

\addplot[name path=W4A2upper, draw=none] coordinates {
  (0.1,58.79) (0.2,58.83) (0.3,58.84) (0.4,58.81) (0.5,58.75)
  (0.6,58.61) (0.7,58.50) (0.8,58.44) (0.9,58.36) (1.0,58.15)
};
\addplot[name path=W4A2lower, draw=none] coordinates {
  (0.1,58.51) (0.2,58.53) (0.3,58.52) (0.4,58.45) (0.5,58.37)
  (0.6,58.31) (0.7,58.22) (0.8,58.08) (0.9,57.94) (1.0,57.81)
};

\addplot[red, opacity=0.15] fill between[of=W4A2upper and W4A2lower];

\nextgroupplot[title={W2A4}, ymin=65.1, ymax=65.35]

\addplot[green!60!black, mark=triangle*] coordinates {
  (0.1,65.12) (0.2,65.20) (0.3,65.24) (0.4,65.26) (0.5,65.26)
  (0.6,65.26) (0.7,65.25) (0.8,65.24) (0.9,65.20) (1.0,65.14)
};

\addplot[name path=W2A4upper, draw=none] coordinates {
  (0.1,65.15) (0.2,65.25) (0.3,65.28) (0.4,65.33) (0.5,65.33)
  (0.6,65.32) (0.7,65.28) (0.8,65.27) (0.9,65.23) (1.0,65.17)
};
\addplot[name path=W2A4lower, draw=none] coordinates {
  (0.1,65.09) (0.2,65.15) (0.3,65.20) (0.4,65.19) (0.5,65.19)
  (0.6,65.20) (0.7,65.22) (0.8,65.21) (0.9,65.17) (1.0,65.11)
};

\addplot[green!60!black, opacity=0.15] fill between[of=W2A4upper and W2A4lower];

\nextgroupplot[title={W4A4}, ymin=69.0, ymax=69.3]

\addplot[purple, mark=diamond*] coordinates {
  (0.1,69.16) (0.2,69.17) (0.3,69.18) (0.4,69.18) (0.5,69.18)
  (0.6,69.17) (0.7,69.17) (0.8,69.16) (0.9,69.15) (1.0,69.13)
};

\addplot[name path=W4A4upper, draw=none] coordinates {
  (0.1,69.26) (0.2,69.27) (0.3,69.30) (0.4,69.29) (0.5,69.27)
  (0.6,69.26) (0.7,69.26) (0.8,69.27) (0.9,69.28) (1.0,69.25)
};
\addplot[name path=W4A4lower, draw=none] coordinates {
  (0.1,69.06) (0.2,69.07) (0.3,69.06) (0.4,69.07) (0.5,69.09)
  (0.6,69.08) (0.7,69.08) (0.8,69.05) (0.9,69.02) (1.0,69.01)
};

\addplot[purple, opacity=0.15] fill between[of=W4A4upper and W4A4lower];

\end{groupplot}

\node at ($(group c1r1.south)!0.5!(group c4r1.south)$)
      [yshift=-0.8cm, font=\normalsize] {$\alpha$};

\end{tikzpicture}
\caption{Ablation of $\alpha$ for ResNet-18 under different quantization precisions 
(W2A2, W2A4, W4A2, W4A4). The solid line denotes the mean Top-1 accuracy across three runs, and the shaded region indicates $\pm 1\sigma$. The comprehensive description of this ablation is provided in \cref{section:ablation} }
\label{fig:ablation-alpha}
\end{figure}

\begin{figure}[H]
\centering
\begin{tikzpicture}
\begin{groupplot}[
  group style={group size=4 by 1, horizontal sep=0.6cm},
  width=0.3\linewidth,
  height=4.2cm,
  grid=major,
  tick label style={font=\scriptsize,/pgf/number format/fixed},
  every axis plot/.append style={very thick},
  xmin=0.09, xmax=1.01,
  xtick={0.2,0.4,...,1.0},
  scaled ticks=false,
  xticklabel style={rotate=45, anchor=east},
  yticklabel style={rotate=45, anchor=east},
  title style={yshift=-0.5ex},
  legend style={cells={anchor=west}, font=\scriptsize, at={(0.2,0.5)}, anchor=north west}
]

\nextgroupplot[title={W2A2}, ymin=57.3, ymax=58.2, ylabel={Top-1 (\%)}]

\addplot[blue, mark=*] coordinates {
  (0.1,57.44) (0.2,57.71) (0.3,57.93) (0.4,58.04) (0.5,58.08)
  (0.6,58.04) (0.7,57.93) (0.8,57.85) (0.9,57.73) (1.0,57.59)
};
\addlegendentry{Mean}

\addplot[name path=W2A2upper, draw=none] coordinates {
  (0.1,57.53) (0.2,57.81) (0.3,58.05) (0.4,58.17) (0.5,58.19)
  (0.6,58.19) (0.7,58.10) (0.8,57.98) (0.9,57.86) (1.0,57.72)
};
\addplot[name path=W2A2lower, draw=none] coordinates {
  (0.1,57.35) (0.2,57.61) (0.3,57.81) (0.4,57.91) (0.5,57.97)
  (0.6,57.89) (0.7,57.76) (0.8,57.72) (0.9,57.60) (1.0,57.46)
};
\addplot[blue, opacity=0.15] fill between[of=W2A2upper and W2A2lower];
\addlegendentry{$\pm$1$\sigma$}

\nextgroupplot[title={W4A2}, ymin=63.7, ymax=64.5]

\addplot[red, mark=square*] coordinates {
  (0.1,64.31) (0.2,64.38) (0.3,64.36) (0.4,64.34) (0.5,64.35)
  (0.6,64.28) (0.7,64.22) (0.8,64.09) (0.9,63.97) (1.0,63.82)
};
\addplot[name path=W4A2upper, draw=none] coordinates {
  (0.1,64.34) (0.2,64.44) (0.3,64.41) (0.4,64.42) (0.5,64.40)
  (0.6,64.37) (0.7,64.38) (0.8,64.22) (0.9,64.12) (1.0,63.96)
};
\addplot[name path=W4A2lower, draw=none] coordinates {
  (0.1,64.28) (0.2,64.32) (0.3,64.31) (0.4,64.26) (0.5,64.30)
  (0.6,64.19) (0.7,64.06) (0.8,63.96) (0.9,63.82) (1.0,63.68)
};
\addplot[red, opacity=0.15] fill between[of=W4A2upper and W4A2lower];

\nextgroupplot[title={W2A4}, ymin=70.9, ymax=71.4]

\addplot[green!60!black, mark=triangle*] coordinates {
  (0.1,70.98) (0.2,71.10) (0.3,71.18) (0.4,71.25) (0.5,71.28)
  (0.6,71.28) (0.7,71.29) (0.8,71.28) (0.9,71.26) (1.0,71.23)
};
\addplot[name path=W2A4upper, draw=none] coordinates {
  (0.1,71.05) (0.2,71.13) (0.3,71.21) (0.4,71.26) (0.5,71.30)
  (0.6,71.31) (0.7,71.31) (0.8,71.32) (0.9,71.29) (1.0,71.26)
};
\addplot[name path=W2A4lower, draw=none] coordinates {
  (0.1,70.91) (0.2,71.07) (0.3,71.15) (0.4,71.24) (0.5,71.26)
  (0.6,71.25) (0.7,71.27) (0.8,71.24) (0.9,71.23) (1.0,71.20)
};
\addplot[green!60!black, opacity=0.15] fill between[of=W2A4upper and W2A4lower];

\nextgroupplot[title={W4A4}, ymin=74.9, ymax=75.3]

\addplot[purple, mark=diamond*] coordinates {
  (0.1,75.07) (0.2,75.08) (0.3,75.11) (0.4,75.11) (0.5,75.12)
  (0.6,75.11) (0.7,75.10) (0.8,75.10) (0.9,75.09) (1.0,75.10)
};
\addplot[name path=W4A4upper, draw=none] coordinates {
  (0.1,75.14) (0.2,75.15) (0.3,75.20) (0.4,75.19) (0.5,75.19)
  (0.6,75.18) (0.7,75.17) (0.8,75.14) (0.9,75.15) (1.0,75.14)
};
\addplot[name path=W4A4lower, draw=none] coordinates {
  (0.1,75.00) (0.2,75.01) (0.3,75.02) (0.4,75.03) (0.5,75.05)
  (0.6,75.04) (0.7,75.03) (0.8,75.06) (0.9,75.03) (1.0,75.06)
};
\addplot[purple, opacity=0.15] fill between[of=W4A4upper and W4A4lower];

\end{groupplot}

\node at ($(group c1r1.south)!0.5!(group c4r1.south)$)
      [yshift=-0.8cm, font=\normalsize] {$\alpha$};

\end{tikzpicture}
\caption{Ablation of $\alpha$ for ResNet-50 under different quantization precisions 
(W2A2, W2A4, W4A2, W4A4). The solid line denotes the mean Top-1 accuracy across three runs, and the shaded region indicates $\pm 1\sigma$. Across quantization settings, $\alpha$ controls the balance between the base quantized output and CAT correction. 
Performance peaks at moderate $\alpha$ ($\approx0.3–0.5$) for W2A2 and W4A2, remains stable for W2A4, and is nearly flat for W4A4, indicating that CAT provides the most benefit in severely quantized regimes with low-bit activations.}
\label{fig:ablation-alpha-rn50}
\end{figure}
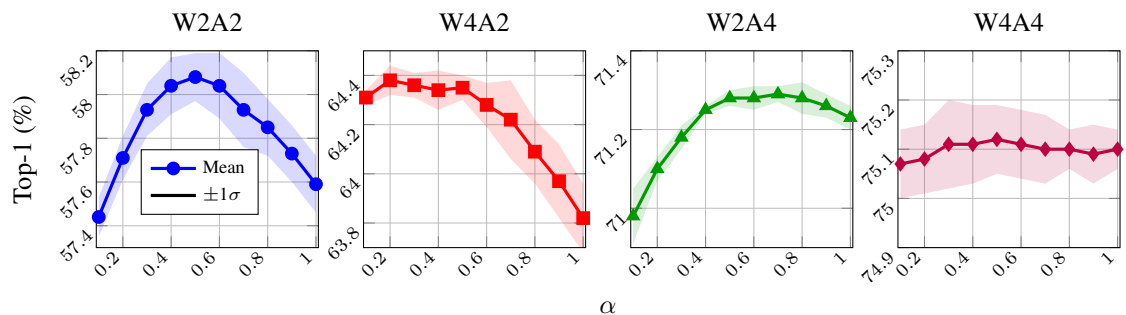


\begin{figure}[H]
\centering
\begin{tikzpicture}
\begin{groupplot}[
  group style={group size=4 by 1, horizontal sep=0.6cm},
  width=0.3\linewidth,
  height=4.2cm,
  grid=major,
  tick label style={font=\scriptsize,/pgf/number format/fixed},
  every axis plot/.append style={very thick},
  xmin=0.09, xmax=1.01,
  xtick={0.2,0.4,...,1.0},
  scaled ticks=false,
  xticklabel style={rotate=45, anchor=east},
  yticklabel style={rotate=45, anchor=east},
  title style={yshift=-0.5ex},
  legend style={cells={anchor=west}, font=\scriptsize, at={(0.2,0.5)}, anchor=north west}
]

\nextgroupplot[title={W2A2}, ymin=12.3, ymax=14.9, ylabel={Top-1 (\%)}]

\addplot[blue, mark=*] coordinates {
  (0.1,13.88) (0.2,14.02) (0.3,14.13) (0.4,14.16) (0.5,14.10)
  (0.6,14.00) (0.7,13.88) (0.8,13.66) (0.9,13.42) (1.0,13.17)
};
\addlegendentry{Mean}

\addplot[name path=MBV2_W2A2_upper, draw=none] coordinates {
  (0.1,14.53) (0.2,14.68) (0.3,14.72) (0.4,14.77) (0.5,14.71)
  (0.6,14.61) (0.7,14.52) (0.8,14.34) (0.9,14.09) (1.0,13.89)
};
\addplot[name path=MBV2_W2A2_lower, draw=none] coordinates {
  (0.1,13.23) (0.2,13.36) (0.3,13.54) (0.4,13.55) (0.5,13.49)
  (0.6,13.39) (0.7,13.24) (0.8,12.98) (0.9,12.75) (1.0,12.45)
};
\addplot[blue, opacity=0.15] fill between[of=MBV2_W2A2_upper and MBV2_W2A2_lower];
\addlegendentry{$\pm$1$\sigma$}

\nextgroupplot[title={W4A2}, ymin=18.6, ymax=21.2]

\addplot[red, mark=square*] coordinates {
  (0.1,20.30) (0.2,20.44) (0.3,20.53) (0.4,20.50) (0.5,20.47)
  (0.6,20.32) (0.7,20.15) (0.8,19.92) (0.9,19.64) (1.0,19.31)
};
\addplot[name path=MBV2_W4A2_upper, draw=none] coordinates {
  (0.1,20.83) (0.2,20.96) (0.3,21.06) (0.4,21.00) (0.5,20.96)
  (0.6,20.85) (0.7,20.69) (0.8,20.47) (0.9,20.13) (1.0,19.78)
};
\addplot[name path=MBV2_W4A2_lower, draw=none] coordinates {
  (0.1,19.77) (0.2,19.92) (0.3,20.00) (0.4,20.00) (0.5,19.98)
  (0.6,19.79) (0.7,19.61) (0.8,19.37) (0.9,19.15) (1.0,18.84)
};
\addplot[red, opacity=0.15] fill between[of=MBV2_W4A2_upper and MBV2_W4A2_lower];

\nextgroupplot[title={W2A4}, ymin=54.3, ymax=55.8]

\addplot[green!60!black, mark=triangle*] coordinates {
  (0.1,55.39) (0.2,55.45) (0.3,55.47) (0.4,55.42) (0.5,55.33)
  (0.6,55.24) (0.7,55.11) (0.8,54.97) (0.9,54.82) (1.0,54.63)
};
\addplot[name path=MBV2_W2A4_upper, draw=none] coordinates {
  (0.1,55.58) (0.2,55.69) (0.3,55.69) (0.4,55.66) (0.5,55.59)
  (0.6,55.52) (0.7,55.37) (0.8,55.24) (0.9,55.11) (1.0,54.88)
};
\addplot[name path=MBV2_W2A4_lower, draw=none] coordinates {
  (0.1,55.20) (0.2,55.21) (0.3,55.25) (0.4,55.18) (0.5,55.07)
  (0.6,54.96) (0.7,54.85) (0.8,54.70) (0.9,54.53) (1.0,54.38)
};
\addplot[green!60!black, opacity=0.15] fill between[of=MBV2_W2A4_upper and MBV2_W2A4_lower];

\nextgroupplot[title={W4A4}, ymin=68.0, ymax=68.3]

\addplot[purple, mark=diamond*] coordinates {
  (0.1,68.19) (0.2,68.21) (0.3,68.20) (0.4,68.22) (0.5,68.20)
  (0.6,68.18) (0.7,68.15) (0.8,68.12) (0.9,68.07) (1.0,68.03)
};
\addplot[name path=MBV2_W4A4_upper, draw=none] coordinates {
  (0.1,68.22) (0.2,68.22) (0.3,68.21) (0.4,68.23) (0.5,68.22)
  (0.6,68.19) (0.7,68.17) (0.8,68.13) (0.9,68.09) (1.0,68.04)
};
\addplot[name path=MBV2_W4A4_lower, draw=none] coordinates {
  (0.1,68.16) (0.2,68.20) (0.3,68.19) (0.4,68.21) (0.5,68.18)
  (0.6,68.17) (0.7,68.13) (0.8,68.11) (0.9,68.05) (1.0,68.02)
};
\addplot[purple, opacity=0.15] fill between[of=MBV2_W4A4_upper and MBV2_W4A4_lower];

\end{groupplot}

\node at ($(group c1r1.south)!0.5!(group c4r1.south)$)
      [yshift=-0.8cm, font=\normalsize] {$\alpha$};

\end{tikzpicture}
\caption{Ablation of $\alpha$ for MobileNetV2 under different quantization precisions (W2A2, W2A4, W4A2, W4A4). The solid line denotes the mean Top-1 accuracy across three runs, and the shaded region indicates $\pm 1\sigma$. For MobileNetV2, CAT gains are largest under severe activation quantization: 
W2A2 and W4A2 peak at moderate $\alpha$ ($\approx0.3–0.4$) before dropping sharply, 
while W2A4 shows only mild variation and W4A4 remains essentially flat, 
indicating that $\alpha$ is most critical when activations are 2-bit.
}
\label{fig:ablation-alpha-mbv2}
\end{figure}
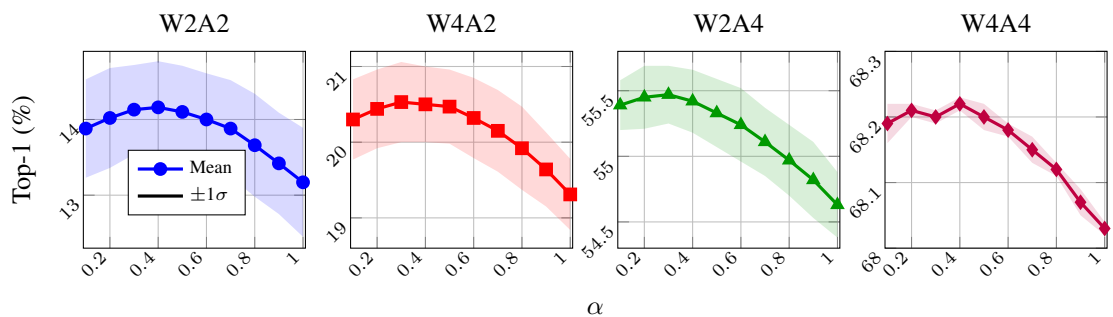


\begin{figure}[H]
\centering
\begin{tikzpicture}
\begin{groupplot}[
  group style={group size=4 by 1, horizontal sep=0.6cm},
  width=0.3\linewidth,
  height=4.2cm,
  grid=major,
  tick label style={font=\scriptsize,/pgf/number format/fixed},
  every axis plot/.append style={very thick},
  xmin=0.09, xmax=1.01,
  xtick={0.2,0.4,...,1.0},
  scaled ticks=false,
  xticklabel style={rotate=45, anchor=east},
  yticklabel style={rotate=45, anchor=east},
  title style={yshift=-0.5ex},
  legend style={cells={anchor=west}, font=\scriptsize, at={(0.2,0.5)}, anchor=north west}
]

\nextgroupplot[title={W2A2}, ymin=40.5, ymax=41.6, ylabel={Top-1 (\%)}]

\addplot[blue, mark=*] coordinates {
  (0.1,41.02) (0.2,41.19) (0.3,41.29) (0.4,41.38) (0.5,41.37)
  (0.6,41.38) (0.7,41.30) (0.8,41.15) (0.9,41.00) (1.0,40.79)
};
\addlegendentry{Mean}

\addplot[name path=RX600_W2A2_upper, draw=none] coordinates {
  (0.1,41.10) (0.2,41.28) (0.3,41.42) (0.4,41.52) (0.5,41.51)
  (0.6,41.48) (0.7,41.37) (0.8,41.20) (0.9,41.04) (1.0,40.84)
};
\addplot[name path=RX600_W2A2_lower, draw=none] coordinates {
  (0.1,40.94) (0.2,41.10) (0.3,41.16) (0.4,41.24) (0.5,41.23)
  (0.6,41.28) (0.7,41.23) (0.8,41.10) (0.9,40.96) (1.0,40.74)
};
\addplot[blue, opacity=0.15] fill between[of=RX600_W2A2_upper and RX600_W2A2_lower];
\addlegendentry{$\pm$1$\sigma$}

\nextgroupplot[title={W4A2}, ymin=50.4, ymax=51.9]

\addplot[red, mark=square*] coordinates {
  (0.1,51.33) (0.2,51.43) (0.3,51.52) (0.4,51.51) (0.5,51.48)
  (0.6,51.40) (0.7,51.24) (0.8,51.11) (0.9,50.95) (1.0,50.77)
};

\addplot[name path=RX600_W4A2_upper, draw=none] coordinates {
  (0.1,51.56) (0.2,51.69) (0.3,51.78) (0.4,51.79) (0.5,51.78)
  (0.6,51.63) (0.7,51.47) (0.8,51.36) (0.9,51.16) (1.0,50.99)
};
\addplot[name path=RX600_W4A2_lower, draw=none] coordinates {
  (0.1,51.10) (0.2,51.17) (0.3,51.26) (0.4,51.23) (0.5,51.18)
  (0.6,51.17) (0.7,51.01) (0.8,50.86) (0.9,50.74) (1.0,50.55)
};
\addplot[red, opacity=0.15] fill between[of=RX600_W4A2_upper and RX600_W4A2_lower];

\nextgroupplot[title={W2A4}, ymin=63.6, ymax=64.6]

\addplot[green!60!black, mark=triangle*] coordinates {
  (0.1,64.01) (0.2,64.05) (0.3,64.10) (0.4,64.17) (0.5,64.20)
  (0.6,64.20) (0.7,64.19) (0.8,64.20) (0.9,64.16) (1.0,64.12)
};

\addplot[name path=RX600_W2A4_upper, draw=none] coordinates {
  (0.1,64.29) (0.2,64.39) (0.3,64.47) (0.4,64.51) (0.5,64.54)
  (0.6,64.49) (0.7,64.45) (0.8,64.48) (0.9,64.44) (1.0,64.42)
};
\addplot[name path=RX600_W2A4_lower, draw=none] coordinates {
  (0.1,63.73) (0.2,63.71) (0.3,63.73) (0.4,63.83) (0.5,63.86)
  (0.6,63.91) (0.7,63.93) (0.8,63.92) (0.9,63.88) (1.0,63.82)
};
\addplot[green!60!black, opacity=0.15] fill between[of=RX600_W2A4_upper and RX600_W2A4_lower];

\nextgroupplot[title={W4A4}, ymin=70.9, ymax=71.05]

\addplot[purple, mark=diamond*] coordinates {
  (0.1,70.98) (0.2,70.98) (0.3,70.99) (0.4,70.97) (0.5,70.98)
  (0.6,70.98) (0.7,70.96) (0.8,70.95) (0.9,70.95) (1.0,70.94)
};

\addplot[name path=RX600_W4A4_upper, draw=none] coordinates {
  (0.1,71.00) (0.2,71.00) (0.3,71.01) (0.4,70.98) (0.5,70.99)
  (0.6,71.00) (0.7,71.00) (0.8,70.99) (0.9,71.00) (1.0,70.98)
};
\addplot[name path=RX600_W4A4_lower, draw=none] coordinates {
  (0.1,70.96) (0.2,70.96) (0.3,70.97) (0.4,70.96) (0.5,70.97)
  (0.6,70.96) (0.7,70.92) (0.8,70.91) (0.9,70.90) (1.0,70.90)
};
\addplot[purple, opacity=0.15] fill between[of=RX600_W4A4_upper and RX600_W4A4_lower];

\end{groupplot}

\node at ($(group c1r1.south)!0.5!(group c4r1.south)$)
      [yshift=-0.8cm, font=\normalsize] {$\alpha$};

\end{tikzpicture}
\caption{Ablation of $\alpha$ for RegNetX-600M under different quantization precisions (W2A2, W2A4, W4A2, W4A4). The solid line denotes the mean Top-1 accuracy across three runs, and the shaded region indicates $\pm 1\sigma$. For RegNetX-600M, $\alpha$ is most influential in 2-bit activation regimes: 
W2A2 and W4A2 peak at moderate $\alpha$ ($\approx0.3–0.5$) and decline at larger values, 
whereas W2A4 shows only mild variation and W4A4 remains flat around 71\%.
}
\label{fig:ablation-alpha-regnetx600m}
\end{figure}
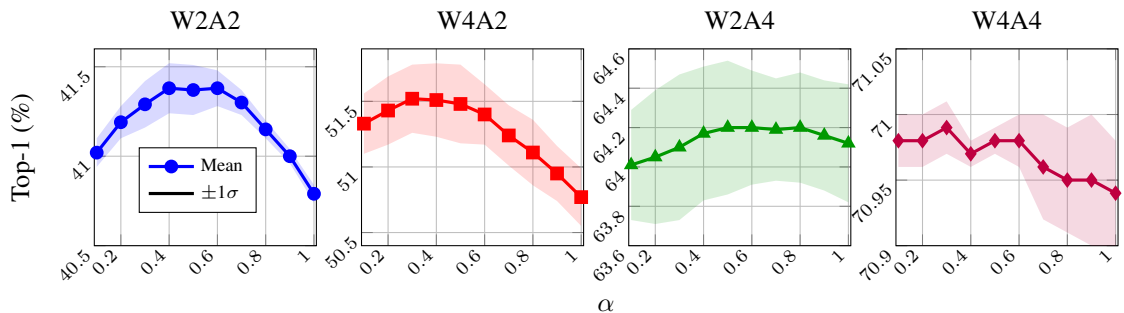

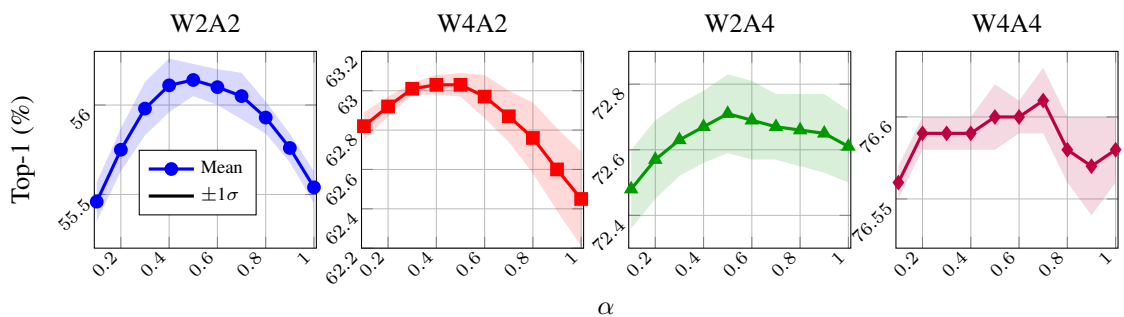
\begin{figure}[H]
\centering
\begin{tikzpicture}
\begin{groupplot}[
  group style={group size=4 by 1, horizontal sep=0.6cm},
  width=0.3\linewidth,
  height=4.2cm,
  grid=major,
  tick label style={font=\scriptsize,/pgf/number format/fixed},
  every axis plot/.append style={very thick},
  xmin=0.09, xmax=1.01,
  xtick={0.2,0.4,...,1.0},
  scaled ticks=false,
  xticklabel style={rotate=45, anchor=east},
  yticklabel style={rotate=45, anchor=east},
  title style={yshift=-0.5ex},
  legend style={cells={anchor=west}, font=\scriptsize, at={(0.2,0.5)}, anchor=north west}
]

\nextgroupplot[title={W2A2}, ymin=55.2, ymax=56.3, ylabel={Top-1 (\%)}]

\addplot[blue, mark=*] coordinates {
  (0.1,55.46) (0.2,55.75) (0.3,55.98) (0.4,56.11) (0.5,56.14)
  (0.6,56.10) (0.7,56.05) (0.8,55.93) (0.9,55.76) (1.0,55.54)
};
\addlegendentry{Mean}

\addplot[name path=RG32_W2A2_upper, draw=none] coordinates {
  (0.1,55.57) (0.2,55.87) (0.3,56.13) (0.4,56.26) (0.5,56.23)
  (0.6,56.20) (0.7,56.18) (0.8,56.02) (0.9,55.84) (1.0,55.63)
};
\addplot[name path=RG32_W2A2_lower, draw=none] coordinates {
  (0.1,55.35) (0.2,55.63) (0.3,55.83) (0.4,55.96) (0.5,56.05)
  (0.6,56.00) (0.7,55.92) (0.8,55.84) (0.9,55.68) (1.0,55.45)
};
\addplot[blue, opacity=0.15] fill between[of=RG32_W2A2_upper and RG32_W2A2_lower];
\addlegendentry{$\pm$1$\sigma$}

\nextgroupplot[title={W4A2}, ymin=62.2, ymax=63.2]

\addplot[red, mark=square*] coordinates {
  (0.1,62.82) (0.2,62.92) (0.3,63.01) (0.4,63.03) (0.5,63.03)
  (0.6,62.97) (0.7,62.87) (0.8,62.76) (0.9,62.60) (1.0,62.45)
};

\addplot[name path=RG32_W4A2_upper, draw=none] coordinates {
  (0.1,62.89) (0.2,62.97) (0.3,63.04) (0.4,63.08) (0.5,63.09)
  (0.6,63.08) (0.7,63.00) (0.8,62.94) (0.9,62.81) (1.0,62.69)
};
\addplot[name path=RG32_W4A2_lower, draw=none] coordinates {
  (0.1,62.75) (0.2,62.87) (0.3,62.98) (0.4,62.98) (0.5,62.97)
  (0.6,62.86) (0.7,62.74) (0.8,62.58) (0.9,62.39) (1.0,62.21)
};
\addplot[red, opacity=0.15] fill between[of=RG32_W4A2_upper and RG32_W4A2_lower];

\nextgroupplot[title={W2A4}, ymin=72.3, ymax=72.9]

\addplot[green!60!black, mark=triangle*] coordinates {
  (0.1,72.48) (0.2,72.57) (0.3,72.63) (0.4,72.67) (0.5,72.71)
  (0.6,72.69) (0.7,72.67) (0.8,72.66) (0.9,72.65) (1.0,72.61)
};

\addplot[name path=RG32_W2A4_upper, draw=none] coordinates {
  (0.1,72.60) (0.2,72.69) (0.3,72.74) (0.4,72.78) (0.5,72.83)
  (0.6,72.81) (0.7,72.77) (0.8,72.77) (0.9,72.77) (1.0,72.72)
};
\addplot[name path=RG32_W2A4_lower, draw=none] coordinates {
  (0.1,72.36) (0.2,72.45) (0.3,72.52) (0.4,72.56) (0.5,72.59)
  (0.6,72.57) (0.7,72.57) (0.8,72.55) (0.9,72.53) (1.0,72.50)
};
\addplot[green!60!black, opacity=0.15] fill between[of=RG32_W2A4_upper and RG32_W2A4_lower];

\nextgroupplot[title={W4A4}, ymin=76.52, ymax=76.64]

\addplot[purple, mark=diamond*] coordinates {
  (0.1,76.56) (0.2,76.59) (0.3,76.59) (0.4,76.59) (0.5,76.60)
  (0.6,76.60) (0.7,76.61) (0.8,76.58) (0.9,76.57) (1.0,76.58)
};

\addplot[name path=RG32_W4A4_upper, draw=none] coordinates {
  (0.1,76.57) (0.2,76.60) (0.3,76.60) (0.4,76.60) (0.5,76.62)
  (0.6,76.61) (0.7,76.63) (0.8,76.60) (0.9,76.60) (1.0,76.60)
};
\addplot[name path=RG32_W4A4_lower, draw=none] coordinates {
  (0.1,76.55) (0.2,76.58) (0.3,76.58) (0.4,76.58) (0.5,76.58)
  (0.6,76.59) (0.7,76.59) (0.8,76.56) (0.9,76.54) (1.0,76.56)
};
\addplot[purple, opacity=0.15] fill between[of=RG32_W4A4_upper and RG32_W4A4_lower];

\end{groupplot}

\node at ($(group c1r1.south)!0.5!(group c4r1.south)$)
      [yshift=-0.8cm, font=\normalsize] {$\alpha$};

\end{tikzpicture}
\caption{Ablation of $\alpha$ for RegNetX-3.2G under different quantization precisions (W2A2, W2A4, W4A2, W4A4). The solid line denotes the mean Top-1 accuracy across three runs, and the shaded region indicates $\pm 1\sigma$. For RegNetX-3.2G, $\alpha$ tuning is most beneficial in 2-bit activation regimes: 
W2A2 and W4A2 peak at moderate values ($\alpha\approx0.3$–$0.5$) before declining, 
while W2A4 shows only mild variation and W4A4 remains flat near 76.6\%.
}
\label{fig:ablation-alpha-regnetx32g}
\end{figure}

\begin{figure}[H]
\centering
\begin{tikzpicture}
\begin{groupplot}[
  group style={group size=4 by 1, horizontal sep=0.6cm},
  width=0.3\linewidth,
  height=4.2cm,
  grid=major,
  tick label style={font=\scriptsize,/pgf/number format/fixed},
  every axis plot/.append style={very thick},
  xmin=0.09, xmax=1.01,
  xtick={0.2,0.4,...,1.0},
  scaled ticks=false,
  xticklabel style={rotate=45, anchor=east},
  yticklabel style={rotate=45, anchor=east},
  title style={yshift=-0.5ex},
  legend style={cells={anchor=west}, font=\scriptsize, at={(0.2,0.5)}, anchor=north west}
]

\nextgroupplot[title={W2A2}, ymin=27.2, ymax=30.0, ylabel={Top-1 (\%)}]

\addplot[blue, mark=*] coordinates {
  (0.1,28.40) (0.2,28.70) (0.3,28.93) (0.4,29.10) (0.5,29.19)
  (0.6,29.20) (0.7,29.07) (0.8,28.83) (0.9,28.56) (1.0,28.14)
};
\addlegendentry{Mean}

\addplot[name path=MNasX2_W2A2_upper, draw=none] coordinates {
  (0.1,29.14) (0.2,29.50) (0.3,29.76) (0.4,29.89) (0.5,29.98)
  (0.6,29.96) (0.7,29.83) (0.8,29.60) (0.9,29.39) (1.0,28.97)
};
\addplot[name path=MNasX2_W2A2_lower, draw=none] coordinates {
  (0.1,27.66) (0.2,27.90) (0.3,28.10) (0.4,28.31) (0.5,28.40)
  (0.6,28.44) (0.7,28.31) (0.8,28.06) (0.9,27.73) (1.0,27.31)
};
\addplot[blue, opacity=0.15] fill between[of=MNasX2_W2A2_upper and MNasX2_W2A2_lower];
\addlegendentry{$\pm$1$\sigma$}

\nextgroupplot[title={W4A2}, ymin=38.6, ymax=40.5]

\addplot[red, mark=square*] coordinates {
  (0.1,39.74) (0.2,39.93) (0.3,40.05) (0.4,40.14) (0.5,40.14)
  (0.6,40.07) (0.7,39.86) (0.8,39.60) (0.9,39.33) (1.0,39.02)
};

\addplot[name path=MNasX2_W4A2_upper, draw=none] coordinates {
  (0.1,40.07) (0.2,40.29) (0.3,40.36) (0.4,40.43) (0.5,40.41)
  (0.6,40.37) (0.7,40.18) (0.8,39.91) (0.9,39.62) (1.0,39.33)
};
\addplot[name path=MNasX2_W4A2_lower, draw=none] coordinates {
  (0.1,39.41) (0.2,39.57) (0.3,39.74) (0.4,39.85) (0.5,39.87)
  (0.6,39.77) (0.7,39.54) (0.8,39.29) (0.9,39.04) (1.0,38.71)
};
\addplot[red, opacity=0.15] fill between[of=MNasX2_W4A2_upper and MNasX2_W4A2_lower];

\nextgroupplot[title={W2A4}, ymin=63.1, ymax=64.4]

\addplot[green!60!black, mark=triangle*] coordinates {
  (0.1,63.55) (0.2,63.70) (0.3,63.84) (0.4,63.92) (0.5,63.95)
  (0.6,63.96) (0.7,63.94) (0.8,63.84) (0.9,63.70) (1.0,63.56)
};

\addplot[name path=MNasX2_W2A4_upper, draw=none] coordinates {
  (0.1,63.78) (0.2,63.96) (0.3,64.09) (0.4,64.22) (0.5,64.22)
  (0.6,64.28) (0.7,64.25) (0.8,64.17) (0.9,64.05) (1.0,63.90)
};
\addplot[name path=MNasX2_W2A4_lower, draw=none] coordinates {
  (0.1,63.32) (0.2,63.44) (0.3,63.59) (0.4,63.62) (0.5,63.68)
  (0.6,63.64) (0.7,63.63) (0.8,63.51) (0.9,63.35) (1.0,63.22)
};
\addplot[green!60!black, opacity=0.15] fill between[of=MNasX2_W2A4_upper and MNasX2_W2A4_lower];

\nextgroupplot[title={W4A4}, ymin=73.0, ymax=73.4]

\addplot[purple, mark=diamond*] coordinates {
  (0.1,73.29) (0.2,73.30) (0.3,73.31) (0.4,73.31) (0.5,73.30)
  (0.6,73.27) (0.7,73.24) (0.8,73.20) (0.9,73.16) (1.0,73.11)
};

\addplot[name path=MNasX2_W4A4_upper, draw=none] coordinates {
  (0.1,73.30) (0.2,73.34) (0.3,73.35) (0.4,73.36) (0.5,73.36)
  (0.6,73.34) (0.7,73.34) (0.8,73.30) (0.9,73.26) (1.0,73.19)
};
\addplot[name path=MNasX2_W4A4_lower, draw=none] coordinates {
  (0.1,73.28) (0.2,73.26) (0.3,73.27) (0.4,73.26) (0.5,73.24)
  (0.6,73.20) (0.7,73.14) (0.8,73.10) (0.9,73.06) (1.0,73.03)
};
\addplot[purple, opacity=0.15] fill between[of=MNasX2_W4A4_upper and MNasX2_W4A4_lower];

\end{groupplot}

\node at ($(group c1r1.south)!0.5!(group c4r1.south)$)
      [yshift=-0.8cm, font=\normalsize] {$\alpha$};

\end{tikzpicture}
\caption{Ablation of $\alpha$ for MNasX2 under different quantization precisions (W2A2, W2A4, W4A2, W4A4). The solid line denotes the mean Top-1 accuracy across three runs, and the shaded region indicates $\pm 1\sigma$. For MNasX2, $\alpha$ tuning is most beneficial in 2-bit activation regimes: 
W2A2 and W4A2 peak at moderate values ($\alpha\approx0.3$–$0.5$) before degrading, 
while W2A4 shows only mild variation and W4A4 remains flat around 73.3\%.
}
\label{fig:ablation-alpha-mnasnet10}
\end{figure}
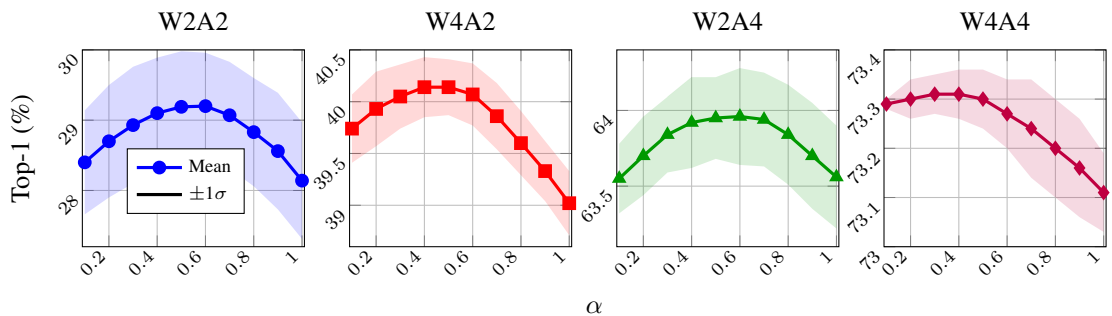

\newpage
\section{Ablation of the number of clusters for different Architectures}
\label{Section_app_cluster}
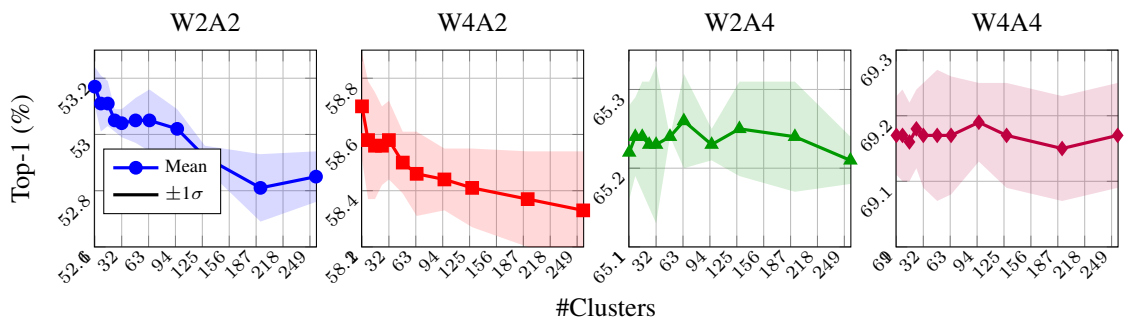
\begin{figure}[ht]
\centering
\begin{tikzpicture}
\begin{groupplot}[
  group style={group size=4 by 1, horizontal sep=0.6cm},
  width=0.3\linewidth,
  height=4.2cm,
  grid=major,
  tick label style={font=\scriptsize,/pgf/number format/fixed},
  every axis plot/.append style={very thick},
  xmin=0.5, xmax=256.5,
  xtick={1,32,...,256},
  scaled ticks=false,
  xticklabel style={rotate=45, anchor=east},
  yticklabel style={rotate=45, anchor=east},
  title style={yshift=-0.5ex},
  legend style={cells={anchor=west}, font=\scriptsize, at={(0.04,0.5)}, anchor=north west}
]

\nextgroupplot[title={W2A2}, ymin=52.6, ymax=53.3, ylabel={Top-1 (\%)}]

\addplot[blue, mark=*] coordinates {
  (1,53.17) (8,53.11) (16,53.11) (24,53.05) (32,53.04)
  (48,53.05) (64,53.05) (96,53.02) (128,52.92) (192,52.81) (256,52.85)
};
\addlegendentry{Mean}

\addplot[name path=W2A2_upper, draw=none] coordinates {
  (1,53.24) (8,53.21) (16,53.19) (24,53.09) (32,53.09)
  (48,53.13) (64,53.16) (96,53.09) (128,52.96) (192,52.93) (256,52.94)
};
\addplot[name path=W2A2_lower, draw=none] coordinates {
  (1,53.10) (8,53.01) (16,53.03) (24,53.01) (32,52.99)
  (48,52.97) (64,52.94) (96,52.95) (128,52.88) (192,52.69) (256,52.76)
};
\addplot[blue, opacity=0.15] fill between[of=W2A2_upper and W2A2_lower];
\addlegendentry{$\pm$1$\sigma$}

\nextgroupplot[title={W4A2}, ymin=58.2, ymax=58.9]

\addplot[red, mark=square*] coordinates {
  (1,58.70) (8,58.58) (16,58.56) (24,58.56) (32,58.58)
  (48,58.50) (64,58.46) (96,58.44) (128,58.41) (192,58.37) (256,58.33)
};

\addplot[name path=W4A2_upper, draw=none] coordinates {
  (1,58.89) (8,58.79) (16,58.75) (24,58.70) (32,58.72)
  (48,58.61) (64,58.61) (96,58.55) (128,58.55) (192,58.54) (256,58.54)
};
\addplot[name path=W4A2_lower, draw=none] coordinates {
  (1,58.51) (8,58.37) (16,58.37) (24,58.42) (32,58.44)
  (48,58.39) (64,58.31) (96,58.33) (128,58.27) (192,58.20) (256,58.12)
};
\addplot[red, opacity=0.15] fill between[of=W4A2_upper and W4A2_lower];

\nextgroupplot[title={W2A4}, ymin=65.1, ymax=65.35]

\addplot[green!60!black, mark=triangle*] coordinates {
  (1,65.22) (8,65.24) (16,65.24) (24,65.23) (32,65.23)
  (48,65.24) (64,65.26) (96,65.23) (128,65.25) (192,65.24) (256,65.21)
};

\addplot[name path=W2A4_upper, draw=none] coordinates {
  (1,65.28) (8,65.29) (16,65.31) (24,65.31) (32,65.33)
  (48,65.24) (64,65.32) (96,65.25) (128,65.31) (192,65.31) (256,65.24)
};
\addplot[name path=W2A4_lower, draw=none] coordinates {
  (1,65.16) (8,65.19) (16,65.17) (24,65.15) (32,65.13)
  (48,65.24) (64,65.20) (96,65.21) (128,65.19) (192,65.17) (256,65.18)
};
\addplot[green!60!black, opacity=0.15] fill between[of=W2A4_upper and W2A4_lower];

\nextgroupplot[title={W4A4}, ymin=69.0, ymax=69.3]

\addplot[purple, mark=diamond*] coordinates {
  (1,69.17) (8,69.17) (16,69.16) (24,69.18) (32,69.17)
  (48,69.17) (64,69.17) (96,69.19) (128,69.17) (192,69.15) (256,69.17)
};

\addplot[name path=W4A4_upper, draw=none] coordinates {
  (1,69.23) (8,69.24) (16,69.22) (24,69.24) (32,69.25)
  (48,69.27) (64,69.26) (96,69.25) (128,69.25) (192,69.23) (256,69.25)
};
\addplot[name path=W4A4_lower, draw=none] coordinates {
  (1,69.11) (8,69.10) (16,69.10) (24,69.12) (32,69.09)
  (48,69.07) (64,69.08) (96,69.13) (128,69.09) (192,69.07) (256,69.09)
};

\addplot[purple, opacity=0.15] fill between[of=W4A4_upper and W4A4_lower];

\end{groupplot}

\node at ($(group c1r1.south)!0.5!(group c4r1.south)$)
      [yshift=-0.8cm, font=\normalsize] {\#Clusters};

\end{tikzpicture}
\caption{Ablation of the number of clusters for ResNet-18 (PCA=50, $\alpha=0.6$) under different quantization precisions (W2A2, W4A2, W2A4; W4A4 pending). Solid line is the mean Top-1 over three runs; shaded region is $\pm 1\sigma$. The comprehensive description of this ablation is provided in \cref{section:ablation}}
\label{fig:ablation-clusters-rn18-one-row}
\end{figure}

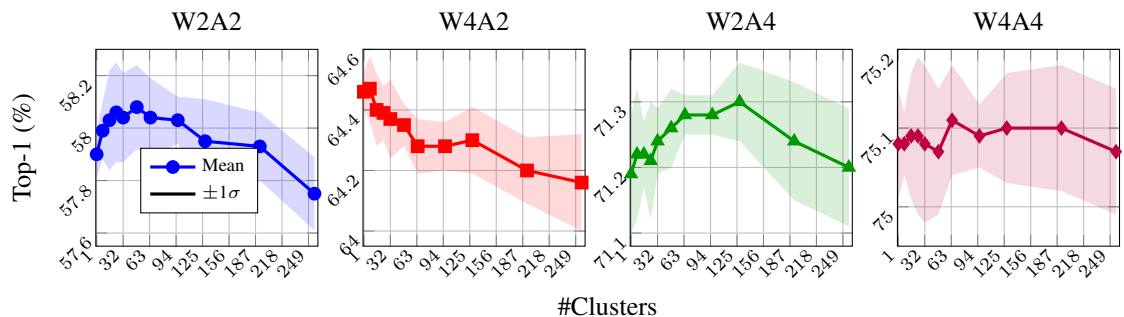
\begin{figure}[H]
\centering
\begin{tikzpicture}
\begin{groupplot}[
  group style={group size=4 by 1, horizontal sep=0.6cm},
  width=0.3\linewidth,
  height=4.2cm,
  grid=major,
  tick label style={font=\scriptsize,/pgf/number format/fixed},
  every axis plot/.append style={very thick},
  xmin=0, xmax=260,
  xtick={1,32,...,256},
  scaled ticks=false,
  xticklabel style={rotate=45, anchor=east},
  yticklabel style={rotate=45, anchor=east},
  title style={yshift=-0.5ex},
  legend style={cells={anchor=west}, font=\scriptsize, at={(0.2,0.5)}, anchor=north west}
]

\nextgroupplot[title={W2A2}, ymin=57.55, ymax=58.30, ylabel={Top-1 (\%)}]

\addplot[blue, mark=*] coordinates {
  (1,57.90) (8,57.99) (16,58.03) (24,58.06) (32,58.04)
  (48,58.08) (64,58.04) (96,58.03) (128,57.95) (192,57.93) (256,57.75)
};
\addlegendentry{Mean}

\addplot[name path=RN50C_W2A2_upper, draw=none] coordinates {
  (1,57.99) (8,58.10) (16,58.22) (24,58.25) (32,58.21)
  (48,58.24) (64,58.19) (96,58.12) (128,58.11) (192,58.06) (256,57.89)
};
\addplot[name path=RN50C_W2A2_lower, draw=none] coordinates {
  (1,57.81) (8,57.88) (16,57.84) (24,57.87) (32,57.87)
  (48,57.92) (64,57.89) (96,57.94) (128,57.79) (192,57.80) (256,57.61)
};
\addplot[blue, opacity=0.15] fill between[of=RN50C_W2A2_upper and RN50C_W2A2_lower];
\addlegendentry{$\pm$1$\sigma$}

\nextgroupplot[title={W4A2}, ymin=63.95, ymax=64.60]

\addplot[red, mark=square*] coordinates {
  (1,64.46) (8,64.47) (16,64.40) (24,64.39) (32,64.37)
  (48,64.35) (64,64.28) (96,64.28) (128,64.30) (192,64.20) (256,64.16)
};

\addplot[name path=RN50C_W4A2_upper, draw=none] coordinates {
  (1,64.52) (8,64.58) (16,64.52) (24,64.48) (32,64.50)
  (48,64.42) (64,64.37) (96,64.36) (128,64.41) (192,64.31) (256,64.32)
};
\addplot[name path=RN50C_W4A2_lower, draw=none] coordinates {
  (1,64.40) (8,64.36) (16,64.28) (24,64.30) (32,64.24)
  (48,64.28) (64,64.19) (96,64.20) (128,64.19) (192,64.09) (256,64.00)
};
\addplot[red, opacity=0.15] fill between[of=RN50C_W4A2_upper and RN50C_W4A2_lower];

\nextgroupplot[title={W2A4}, ymin=71.08, ymax=71.38]

\addplot[green!60!black, mark=triangle*] coordinates {
  (1,71.19) (8,71.22) (16,71.22) (24,71.21) (32,71.24)
  (48,71.26) (64,71.28) (96,71.28) (128,71.30) (192,71.24) (256,71.20)
};

\addplot[name path=RN50C_W2A4_upper, draw=none] coordinates {
  (1,71.28) (8,71.32) (16,71.27) (24,71.30) (32,71.29)
  (48,71.32) (64,71.31) (96,71.31) (128,71.36) (192,71.33) (256,71.29)
};
\addplot[name path=RN50C_W2A4_lower, draw=none] coordinates {
  (1,71.10) (8,71.12) (16,71.17) (24,71.12) (32,71.19)
  (48,71.20) (64,71.25) (96,71.25) (128,71.24) (192,71.15) (256,71.11)
};
\addplot[green!60!black, opacity=0.15] fill between[of=RN50C_W2A4_upper and RN50C_W2A4_lower];

\nextgroupplot[title={W4A4}, ymin=74.95, ymax=75.20]

\addplot[purple, mark=diamond*] coordinates {
  (1,75.08) (8,75.08) (16,75.09) (24,75.09) (32,75.08)
  (48,75.07) (64,75.11) (96,75.09) (128,75.10) (192,75.10) (256,75.07)
};

\addplot[name path=RN50C_W4A4_upper, draw=none] coordinates {
  (1,75.15) (8,75.12) (16,75.17) (24,75.19) (32,75.18)
  (48,75.15) (64,75.18) (96,75.13) (128,75.17) (192,75.18) (256,75.15)
};
\addplot[name path=RN50C_W4A4_lower, draw=none] coordinates {
  (1,75.01) (8,75.04) (16,75.01) (24,74.99) (32,74.98)
  (48,74.99) (64,75.04) (96,75.05) (128,75.03) (192,75.02) (256,74.99)
};
\addplot[purple, opacity=0.15] fill between[of=RN50C_W4A4_upper and RN50C_W4A4_lower];

\end{groupplot}

\node at ($(group c1r1.south)!0.5!(group c4r1.south)$)
      [yshift=-0.8cm, font=\normalsize] {\#Clusters};

\end{tikzpicture}
\caption{Ablation of number of clusters for ResNet-50 under different quantization precisions (W2A2, W2A4, W4A2, W4A4). The solid line denotes the mean Top-1 accuracy across three runs, and the shaded region indicates $\pm 1\sigma$. On ResNet-50, W2A2 benefits slightly from moderate $K$ (peak $\sim$58.08\% near $K{\approx}48$),
whereas W4A2 declines with larger $K$, W2A4 shows a shallow peak near $K{\approx}128$, and W4A4 is flat.
Small-to-moderate $K$ suffices; very large $K$ offers no consistent gains.
}
\label{fig:ablation-clusters-rn50}
\end{figure}

\begin{figure}[H]
\centering
\begin{tikzpicture}
\begin{groupplot}[
  group style={group size=4 by 1, horizontal sep=0.6cm},
  width=0.3\linewidth,
  height=4.2cm,
  grid=major,
  tick label style={font=\scriptsize,/pgf/number format/fixed},
  every axis plot/.append style={very thick},
  xmin=0, xmax=260,
  xtick={1,32,...,256},
  scaled ticks=false,
  xticklabel style={rotate=45, anchor=east},
  yticklabel style={rotate=45, anchor=east},
  title style={yshift=-0.5ex},
  legend style={cells={anchor=west}, font=\scriptsize, at={(0.2,0.5)}, anchor=north west}
]

\nextgroupplot[title={W2A2}, ymin=13.0, ymax=15.0, ylabel={Top-1 (\%)}]

\addplot[blue, mark=*] coordinates {
  (1,13.63) (8,13.73) (16,13.81) (24,13.83) (32,13.95)
  (48,13.99) (64,14.01) (96,14.14) (128,14.14) (192,14.20) (256,14.27)
};
\addlegendentry{Mean}

\addplot[name path=MBV2C_W2A2_upper, draw=none] coordinates {
  (1,14.23) (8,14.32) (16,14.46) (24,14.48) (32,14.55)
  (48,14.65) (64,14.59) (96,14.88) (128,14.79) (192,14.79) (256,14.88)
};
\addplot[name path=MBV2C_W2A2_lower, draw=none] coordinates {
  (1,13.03) (8,13.14) (16,13.16) (24,13.18) (32,13.35)
  (48,13.33) (64,13.43) (96,13.40) (128,13.49) (192,13.61) (256,13.66)
};
\addplot[blue, opacity=0.15] fill between[of=MBV2C_W2A2_upper and MBV2C_W2A2_lower];
\addlegendentry{$\pm$1$\sigma$}

\nextgroupplot[title={W4A2}, ymin=19.4, ymax=21.2]

\addplot[red, mark=square*] coordinates {
  (1,20.15) (8,20.18) (16,20.19) (24,20.24) (32,20.26)
  (48,20.27) (64,20.33) (96,20.36) (128,20.41) (192,20.43) (256,20.49)
};

\addplot[name path=MBV2C_W4A2_upper, draw=none] coordinates {
  (1,20.61) (8,20.68) (16,20.70) (24,20.72) (32,20.70)
  (48,20.70) (64,20.86) (96,20.87) (128,20.84) (192,20.89) (256,21.08)
};
\addplot[name path=MBV2C_W4A2_lower, draw=none] coordinates {
  (1,19.69) (8,19.68) (16,19.68) (24,19.76) (32,19.82)
  (48,19.84) (64,19.80) (96,19.85) (128,19.98) (192,19.97) (256,19.90)
};
\addplot[red, opacity=0.15] fill between[of=MBV2C_W4A2_upper and MBV2C_W4A2_lower];

\nextgroupplot[title={W2A4}, ymin=54.9, ymax=55.7]

\addplot[green!60!black, mark=triangle*] coordinates {
  (1,55.34) (8,55.30) (16,55.28) (24,55.26) (32,55.24)
  (48,55.21) (64,55.24) (96,55.21) (128,55.12) (192,55.17) (256,55.23)
};

\addplot[name path=MBV2C_W2A4_upper, draw=none] coordinates {
  (1,55.57) (8,55.54) (16,55.52) (24,55.51) (32,55.54)
  (48,55.46) (64,55.52) (96,55.49) (128,55.30) (192,55.37) (256,55.51)
};
\addplot[name path=MBV2C_W2A4_lower, draw=none] coordinates {
  (1,55.11) (8,55.06) (16,55.04) (24,55.01) (32,54.94)
  (48,54.96) (64,54.96) (96,54.93) (128,54.94) (192,54.97) (256,54.95)
};
\addplot[green!60!black, opacity=0.15] fill between[of=MBV2C_W2A4_upper and MBV2C_W2A4_lower];

\nextgroupplot[title={W4A4}, ymin=68.0, ymax=68.32]

\addplot[purple, mark=diamond*] coordinates {
  (1,68.21) (8,68.19) (16,68.16) (24,68.16) (32,68.16)
  (48,68.19) (64,68.18) (96,68.17) (128,68.15) (192,68.12) (256,68.13)
};

\addplot[name path=MBV2C_W4A4_upper, draw=none] coordinates {
  (1,68.22) (8,68.25) (16,68.21) (24,68.20) (32,68.20)
  (48,68.24) (64,68.19) (96,68.19) (128,68.17) (192,68.13) (256,68.22)
};
\addplot[name path=MBV2C_W4A4_lower, draw=none] coordinates {
  (1,68.20) (8,68.13) (16,68.11) (24,68.12) (32,68.12)
  (48,68.14) (64,68.17) (96,68.15) (128,68.13) (192,68.11) (256,68.04)
};
\addplot[purple, opacity=0.15] fill between[of=MBV2C_W4A4_upper and MBV2C_W4A4_lower];

\end{groupplot}

\node at ($(group c1r1.south)!0.5!(group c4r1.south)$)
      [yshift=-0.8cm, font=\normalsize] {\#Clusters};

\end{tikzpicture}
\caption{Ablation of number of clusters for MobileNetV2 under different quantization precisions (W2A2, W2A4, W4A2, W4A4). The solid line denotes the mean Top-1 accuracy across three runs, and the shaded region indicates $\pm 1\sigma$. On MobileNetV2, W2A2 and W4A2 improve steadily with larger $K$, peaking at $K{=}256$, while W2A4
declines slightly, and W4A4 is flat. Thus, more clusters help only in severe activation quantization.
}
\label{fig:ablation-clusters-mbv2}
\end{figure}
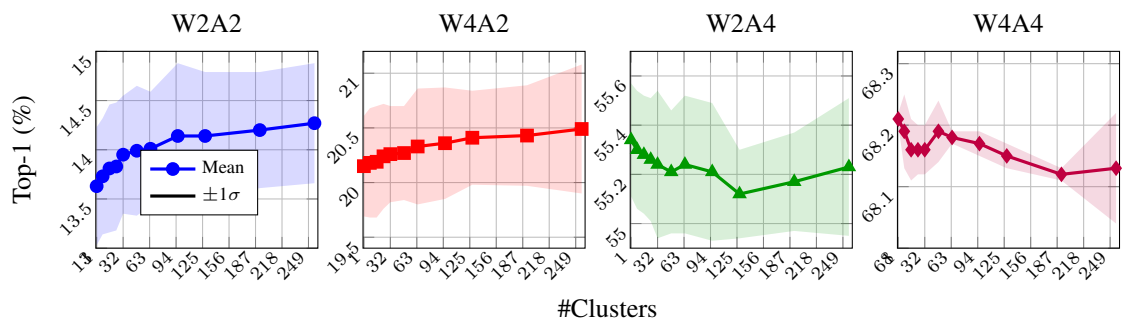

\newpage
\section{Ablation of PCA dimensions for different Architectures}
\label{Section:app_pca}

\begin{figure}[ht]
\centering
\begin{tikzpicture}
\begin{groupplot}[
  group style={group size=4 by 1, horizontal sep=0.6cm},
  width=0.3\linewidth,
  height=4.2cm,
  grid=major,
  tick label style={font=\scriptsize,/pgf/number format/fixed},
  every axis plot/.append style={very thick},
  xmin=0, xmax=222,
  xtick={1,20,40,60,80,100,120,140,160,180,200,220},
  scaled ticks=false,
  xticklabel style={rotate=45, anchor=east},
  yticklabel style={rotate=45, anchor=east},
  title style={yshift=-0.5ex},
  legend style={cells={anchor=west}, font=\scriptsize, at={(0.2,0.50)}, anchor=north west}
]

\nextgroupplot[title={W2A2}, ymin=52.8, ymax=53.2, ylabel={Top-1 (\%)}]

\addplot[blue, mark=*] coordinates {
  (1,53.13) (5,53.14) (10,53.07) (15,53.05) (20,53.05)
  (25,53.01) (30,53.02) (35,53.07) (40,53.01) (45,53.04)
  (50,53.05) (55,53.03) (60,53.03) (65,53.02) (70,53.00)
  (75,53.03) (80,53.03) (85,52.98) (90,53.04) (95,53.03)
  (100,52.95) (105,53.01) (110,52.99) (115,53.04) (120,53.06)
  (125,52.99) (130,53.02) (135,53.02) (140,53.03) (145,53.03)
  (150,53.01) (155,53.04) (160,53.03) (165,53.03) (170,52.96)
  (175,53.00) (180,53.01) (185,53.02) (190,53.04) (195,53.02)
  (200,53.01) (205,52.98) (210,53.02) (215,52.99) (220,53.00)
};
\addlegendentry{Mean}

\addplot[name path=W2A2upper, draw=none] coordinates {
  (1,53.19) (5,53.21) (10,53.13) (15,53.15) (20,53.11)
  (25,53.08) (30,53.15) (35,53.14) (40,53.10) (45,53.18)
  (50,53.17) (55,53.16) (60,53.07) (65,53.12) (70,53.10)
  (75,53.12) (80,53.11) (85,53.08) (90,53.10) (95,53.11)
  (100,53.06) (105,53.09) (110,53.12) (115,53.17) (120,53.12)
  (125,53.09) (130,53.06) (135,53.07) (140,53.12) (145,53.11)
  (150,53.13) (155,53.12) (160,53.13) (165,53.10) (170,53.07)
  (175,53.06) (180,53.13) (185,53.09) (190,53.10) (195,53.09)
  (200,53.11) (205,53.05) (210,53.06) (215,53.05) (220,53.05)
};
\addplot[name path=W2A2lower, draw=none] coordinates {
  (1,53.07) (5,53.07) (10,53.01) (15,52.95) (20,52.99)
  (25,52.94) (30,52.89) (35,53.00) (40,52.92) (45,52.90)
  (50,52.93) (55,52.90) (60,52.99) (65,52.92) (70,52.90)
  (75,52.94) (80,52.95) (85,52.88) (90,52.98) (95,52.95)
  (100,52.84) (105,52.93) (110,52.86) (115,52.91) (120,53.00)
  (125,52.89) (130,52.98) (135,52.97) (140,52.94) (145,52.95)
  (150,52.89) (155,52.96) (160,52.93) (165,52.96) (170,52.85)
  (175,52.94) (180,52.89) (185,52.95) (190,52.98) (195,52.95)
  (200,52.91) (205,52.91) (210,52.98) (215,52.93) (220,52.95)
};

\addplot[blue, opacity=0.15] fill between[of=W2A2upper and W2A2lower];
\addlegendentry{$\pm$1$\sigma$}

\nextgroupplot[title={W4A2}, ymin=58.23, ymax=58.78]

\addplot[red, mark=square*] coordinates {
  (1,58.68) (5,58.62) (10,58.57) (15,58.54) (20,58.55)
  (25,58.56) (30,58.53) (35,58.50) (40,58.51) (45,58.48)
  (50,58.46) (55,58.49) (60,58.43) (65,58.48) (70,58.47)
  (75,58.44) (80,58.46) (85,58.45) (90,58.48) (95,58.43)
  (100,58.48) (105,58.48) (110,58.44) (115,58.48) (120,58.48)
  (125,58.45) (130,58.46) (135,58.45) (140,58.43) (145,58.45)
  (150,58.46) (155,58.46) (160,58.46) (165,58.48) (170,58.43)
  (175,58.44) (180,58.44) (185,58.46) (190,58.45) (195,58.45)
  (200,58.45) (205,58.46) (210,58.44) (215,58.46) (220,58.47)
};

\addplot[name path=W4A2upper, draw=none] coordinates {
  (1,58.85) (5,58.82) (10,58.74) (15,58.78) (20,58.74)
  (25,58.76) (30,58.69) (35,58.66) (40,58.68) (45,58.61)
  (50,58.61) (55,58.66) (60,58.59) (65,58.63) (70,58.63)
  (75,58.60) (80,58.60) (85,58.56) (90,58.65) (95,58.61)
  (100,58.63) (105,58.66) (110,58.61) (115,58.64) (120,58.66)
  (125,58.59) (130,58.64) (135,58.60) (140,58.57) (145,58.59)
  (150,58.61) (155,58.67) (160,58.60) (165,58.66) (170,58.60)
  (175,58.61) (180,58.57) (185,58.61) (190,58.62) (195,58.63)
  (200,58.61) (205,58.64) (210,58.61) (215,58.59) (220,58.58)
};
\addplot[name path=W4A2lower, draw=none] coordinates {
  (1,58.51) (5,58.42) (10,58.40) (15,58.30) (20,58.36)
  (25,58.36) (30,58.37) (35,58.34) (40,58.34) (45,58.35)
  (50,58.31) (55,58.32) (60,58.27) (65,58.33) (70,58.31)
  (75,58.28) (80,58.32) (85,58.34) (90,58.31) (95,58.25)
  (100,58.33) (105,58.30) (110,58.27) (115,58.32) (120,58.30)
  (125,58.31) (130,58.28) (135,58.30) (140,58.29) (145,58.31)
  (150,58.31) (155,58.25) (160,58.32) (165,58.30) (170,58.26)
  (175,58.27) (180,58.31) (185,58.31) (190,58.28) (195,58.27)
  (200,58.29) (205,58.28) (210,58.27) (215,58.33) (220,58.36)
};

\addplot[red, opacity=0.15] fill between[of=W4A2upper and W4A2lower];

\nextgroupplot[title={W2A4}, ymin=65.14, ymax=65.35]

\addplot[green!60!black, mark=triangle*] coordinates {
  (1,65.25) (5,65.25) (10,65.24) (15,65.22) (20,65.23)
  (25,65.24) (30,65.27) (35,65.27) (40,65.26) (45,65.25)
  (50,65.26) (55,65.25) (60,65.25) (65,65.22) (70,65.22)
  (75,65.23) (80,65.24) (85,65.25) (90,65.24) (95,65.24)
  (100,65.26) (105,65.25) (110,65.23) (115,65.21) (120,65.24)
  (125,65.21) (130,65.25) (135,65.22) (140,65.20) (145,65.19)
  (150,65.23) (155,65.23) (160,65.23) (165,65.23) (170,65.21)
  (175,65.20) (180,65.22) (185,65.26) (190,65.23) (195,65.23)
  (200,65.22) (205,65.23) (210,65.22) (215,65.23) (220,65.26)
};

\addplot[name path=W2A4upper, draw=none] coordinates {
  (1,65.29) (5,65.32) (10,65.31) (15,65.28) (20,65.26)
  (25,65.29) (30,65.33) (35,65.32) (40,65.33) (45,65.33)
  (50,65.32) (55,65.32) (60,65.30) (65,65.26) (70,65.24)
  (75,65.28) (80,65.30) (85,65.27) (90,65.29) (95,65.29)
  (100,65.32) (105,65.31) (110,65.25) (115,65.27) (120,65.28)
  (125,65.25) (130,65.31) (135,65.28) (140,65.24) (145,65.22)
  (150,65.28) (155,65.27) (160,65.29) (165,65.27) (170,65.29)
  (175,65.24) (180,65.25) (185,65.31) (190,65.27) (195,65.32)
  (200,65.24) (205,65.30) (210,65.24) (215,65.29) (220,65.31)
};
\addplot[name path=W2A4lower, draw=none] coordinates {
  (1,65.21) (5,65.18) (10,65.17) (15,65.16) (20,65.20)
  (25,65.19) (30,65.21) (35,65.22) (40,65.19) (45,65.17)
  (50,65.20) (55,65.18) (60,65.20) (65,65.18) (70,65.20)
  (75,65.18) (80,65.18) (85,65.23) (90,65.19) (95,65.19)
  (100,65.20) (105,65.19) (110,65.21) (115,65.15) (120,65.20)
  (125,65.17) (130,65.19) (135,65.16) (140,65.16) (145,65.16)
  (150,65.18) (155,65.19) (160,65.17) (165,65.19) (170,65.13)
  (175,65.16) (180,65.19) (185,65.21) (190,65.19) (195,65.14)
  (200,65.20) (205,65.16) (210,65.20) (215,65.17) (220,65.21)
};

\addplot[green!60!black, opacity=0.15] fill between[of=W2A4upper and W2A4lower];

\nextgroupplot[title={W4A4}, ymin=69.02, ymax=69.33]

\addplot[purple, mark=diamond*] coordinates {
  (1,69.17) (5,69.17) (10,69.16) (15,69.15) (20,69.16)
  (25,69.15) (30,69.20) (35,69.19) (40,69.18) (45,69.19)
  (50,69.18) (55,69.19) (60,69.19) (65,69.17) (70,69.19)
  (75,69.17) (80,69.17) (85,69.18) (90,69.20) (95,69.18)
  (100,69.18) (105,69.17) (110,69.17) (115,69.18) (120,69.16)
  (125,69.17) (130,69.18) (135,69.15) (140,69.17) (145,69.16)
  (150,69.18) (155,69.18) (160,69.18) (165,69.18) (170,69.19)
  (175,69.17) (180,69.18) (185,69.18) (190,69.15) (195,69.17)
  (200,69.18) (205,69.17) (210,69.20) (215,69.18) (220,69.18)
};

\addplot[name path=W4A4upper, draw=none] coordinates {
  (1,69.25) (5,69.24) (10,69.22) (15,69.26) (20,69.25)
  (25,69.24) (30,69.28) (35,69.28) (40,69.27) (45,69.27)
  (50,69.27) (55,69.27) (60,69.29) (65,69.27) (70,69.28)
  (75,69.26) (80,69.26) (85,69.27) (90,69.28) (95,69.27)
  (100,69.28) (105,69.26) (110,69.26) (115,69.27) (120,69.26)
  (125,69.28) (130,69.29) (135,69.24) (140,69.25) (145,69.26)
  (150,69.25) (155,69.27) (160,69.29) (165,69.30) (170,69.29)
  (175,69.27) (180,69.27) (185,69.28) (190,69.25) (195,69.27)
  (200,69.27) (205,69.27) (210,69.28) (215,69.26) (220,69.28)
};
\addplot[name path=W4A4lower, draw=none] coordinates {
  (1,69.09) (5,69.10) (10,69.10) (15,69.04) (20,69.07)
  (25,69.06) (30,69.12) (35,69.10) (40,69.09) (45,69.11)
  (50,69.09) (55,69.11) (60,69.09) (65,69.07) (70,69.10)
  (75,69.08) (80,69.08) (85,69.09) (90,69.12) (95,69.09)
  (100,69.08) (105,69.08) (110,69.08) (115,69.09) (120,69.06)
  (125,69.06) (130,69.07) (135,69.06) (140,69.09) (145,69.06)
  (150,69.11) (155,69.09) (160,69.07) (165,69.06) (170,69.09)
  (175,69.07) (180,69.09) (185,69.08) (190,69.05) (195,69.07)
  (200,69.09) (205,69.07) (210,69.12) (215,69.10) (220,69.08)
};

\addplot[purple, opacity=0.15] fill between[of=W4A4upper and W4A4lower];

\end{groupplot}

\node at ($(group c1r1.south)!0.5!(group c4r1.south)$)
      [yshift=-0.8cm, font=\normalsize] {PCA dimension};

\end{tikzpicture}
\caption{Ablation of PCA dimension $k$ for ResNet-18 under different quantization precisions (W2A2, W4A2, W2A4, W4A4). Solid lines show mean Top-1 accuracy over three runs; the shaded region is $\pm 1\sigma$. The comprehensive description of this ablation is provided in \cref{section:ablation}}
\label{fig:ablation-pca-rn18}
\end{figure}
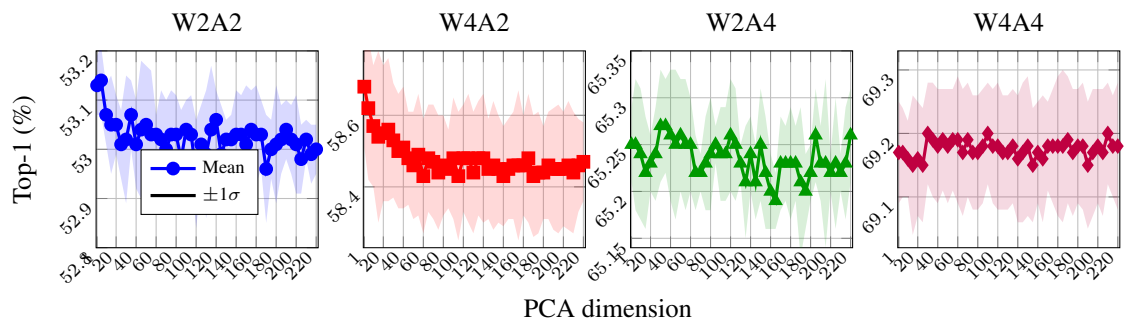

\begin{figure}[H]
\centering
\begin{tikzpicture}
\begin{groupplot}[
  group style={group size=4 by 1, horizontal sep=0.6cm},
  width=0.30\linewidth,
  height=4.2cm,
  grid=major,
  tick label style={font=\scriptsize,/pgf/number format/fixed},
  every axis plot/.append style={very thick},
  xmin=0.5, xmax=220.5,
  xtick={1,20,...,210}, 
  scaled ticks=false,
  xticklabel style={rotate=45, anchor=east},
  yticklabel style={rotate=45, anchor=east},
  title style={yshift=-0.5ex},
  legend style={cells={anchor=west}, font=\scriptsize, at={(0.02,0.98)}, anchor=north west}
]

\nextgroupplot[title={W2A2}, ymin=57.90, ymax=58.38, ylabel={Top-1 (\%)}]
\addplot[blue, mark=*] coordinates {
  (1,57.94) (5,58.01) (10,58.09) (15,58.09) (20,58.04) (25,57.99) (30,58.12) (35,58.10) (40,58.03) (45,58.09)
  (50,58.04) (55,58.06) (60,58.06) (65,58.06) (70,58.08) (75,57.98) (80,58.02) (85,58.02) (90,58.05) (95,58.02)
  (100,58.04) (105,58.01) (110,57.98) (115,58.04) (120,58.07) (125,57.97) (130,58.05) (135,58.06) (140,58.05) (145,58.05)
  (150,57.99) (155,58.00) (160,58.03) (165,58.05) (170,57.98) (175,58.02) (180,57.95) (185,58.04) (190,58.05) (195,58.05)
  (200,57.99) (205,58.04) (210,58.05) (215,57.98) (220,58.04)
};
\addlegendentry{Mean}
\addplot[name path=RN50_W2A2_upper, draw=none] coordinates {
  (1,58.05) (5,58.13) (10,58.21) (15,58.24) (20,58.24) (25,58.18) (30,58.31) (35,58.34) (40,58.21) (45,58.24)
  (50,58.19) (55,58.18) (60,58.19) (65,58.19) (70,58.19) (75,58.11) (80,58.13) (85,58.12) (90,58.22) (95,58.15)
  (100,58.16) (105,58.17) (110,58.13) (115,58.17) (120,58.27) (125,58.19) (130,58.21) (135,58.20) (140,58.21) (145,58.20)
  (150,58.14) (155,58.15) (160,58.11) (165,58.19) (170,58.15) (175,58.20) (180,58.10) (185,58.11) (190,58.17) (195,58.16)
  (200,58.16) (205,58.20) (210,58.19) (215,58.09) (220,58.15)
};
\addplot[name path=RN50_W2A2_lower, draw=none] coordinates {
  (1,57.83) (5,57.89) (10,57.97) (15,57.94) (20,57.84) (25,57.80) (30,57.93) (35,57.86) (40,57.85) (45,57.94)
  (50,57.89) (55,57.94) (60,57.93) (65,57.93) (70,57.97) (75,57.85) (80,57.91) (85,57.92) (90,57.88) (95,57.89)
  (100,57.92) (105,57.85) (110,57.83) (115,57.91) (120,57.87) (125,57.75) (130,57.89) (135,57.92) (140,57.89) (145,57.90)
  (150,57.84) (155,57.85) (160,57.95) (165,57.91) (170,57.81) (175,57.84) (180,57.80) (185,57.97) (190,57.93) (195,57.94)
  (200,57.82) (205,57.88) (210,57.91) (215,57.87) (220,57.93)
};
\addplot[blue, opacity=0.15] fill between[of=RN50_W2A2_upper and RN50_W2A2_lower];
\addlegendentry{$\pm$1$\sigma$}

\nextgroupplot[title={W4A2}, ymin=64.1, ymax=64.62]
\addplot[red, mark=square*] coordinates {
  (1,64.44) (5,64.45) (10,64.40) (15,64.36) (20,64.33) (25,64.29) (30,64.30) (35,64.36) (40,64.31) (45,64.31)
  (50,64.28) (55,64.31) (60,64.30) (65,64.35) (70,64.34) (75,64.35) (80,64.29) (85,64.33) (90,64.36) (95,64.32)
  (100,64.32) (105,64.31) (110,64.34) (115,64.34) (120,64.34) (125,64.34) (130,64.31) (135,64.32) (140,64.26) (145,64.32)
  (150,64.32) (155,64.35) (160,64.32) (165,64.34) (170,64.33) (175,64.35) (180,64.28) (185,64.30) (190,64.31) (195,64.33)
  (200,64.31) (205,64.34) (210,64.34) (215,64.29) (220,64.30)
};
\addplot[name path=RN50_W4A2_upper, draw=none] coordinates {
  (1,64.59) (5,64.53) (10,64.49) (15,64.37) (20,64.36) (25,64.35) (30,64.35) (35,64.40) (40,64.40) (45,64.37)
  (50,64.37) (55,64.42) (60,64.41) (65,64.44) (70,64.42) (75,64.40) (80,64.34) (85,64.40) (90,64.39) (95,64.37)
  (100,64.44) (105,64.34) (110,64.45) (115,64.45) (120,64.40) (125,64.42) (130,64.35) (135,64.39) (140,64.33) (145,64.37)
  (150,64.35) (155,64.41) (160,64.41) (165,64.45) (170,64.47) (175,64.44) (180,64.36) (185,64.36) (190,64.35) (195,64.42)
  (200,64.39) (205,64.41) (210,64.44) (215,64.39) (220,64.32)
};
\addplot[name path=RN50_W4A2_lower, draw=none] coordinates {
  (1,64.29) (5,64.37) (10,64.31) (15,64.35) (20,64.30) (25,64.23) (30,64.25) (35,64.32) (40,64.22) (45,64.25)
  (50,64.19) (55,64.20) (60,64.19) (65,64.26) (70,64.26) (75,64.30) (80,64.24) (85,64.26) (90,64.33) (95,64.27)
  (100,64.20) (105,64.28) (110,64.23) (115,64.23) (120,64.28) (125,64.26) (130,64.27) (135,64.25) (140,64.19) (145,64.27)
  (150,64.29) (155,64.29) (160,64.23) (165,64.23) (170,64.19) (175,64.26) (180,64.20) (185,64.24) (190,64.27) (195,64.24)
  (200,64.23) (205,64.27) (210,64.24) (215,64.19) (220,64.28)
};
\addplot[red, opacity=0.15] fill between[of=RN50_W4A2_upper and RN50_W4A2_lower];

\nextgroupplot[title={W2A4}, ymin=71.10, ymax=71.40]
\addplot[green!60!black, mark=triangle*] coordinates {
  (1,71.20) (5,71.21) (10,71.27) (15,71.25) (20,71.27) (25,71.27) (30,71.27) (35,71.25) (40,71.27) (45,71.30)
  (50,71.28) (55,71.27) (60,71.26) (65,71.28) (70,71.28) (75,71.26) (80,71.30) (85,71.28) (90,71.29) (95,71.25)
  (100,71.26) (105,71.30) (110,71.27) (115,71.29) (120,71.25) (125,71.28) (130,71.26) (135,71.28) (140,71.28) (145,71.29)
  (150,71.26) (155,71.28) (160,71.25) (165,71.28) (170,71.29) (175,71.27) (180,71.26) (185,71.32) (190,71.28) (195,71.26)
  (200,71.26) (205,71.28) (210,71.27) (215,71.28) (220,71.29)
};
\addplot[name path=RN50_W2A4_upper, draw=none] coordinates {
  (1,71.27) (5,71.28) (10,71.31) (15,71.31) (20,71.32) (25,71.28) (30,71.31) (35,71.28) (40,71.31) (45,71.36)
  (50,71.31) (55,71.30) (60,71.29) (65,71.34) (70,71.31) (75,71.30) (80,71.34) (85,71.32) (90,71.31) (95,71.33)
  (100,71.30) (105,71.35) (110,71.31) (115,71.34) (120,71.30) (125,71.36) (130,71.30) (135,71.33) (140,71.37) (145,71.39)
  (150,71.29) (155,71.31) (160,71.29) (165,71.33) (170,71.33) (175,71.30) (180,71.29) (185,71.39) (190,71.34) (195,71.34)
  (200,71.30) (205,71.31) (210,71.33) (215,71.36) (220,71.35)
};
\addplot[name path=RN50_W2A4_lower, draw=none] coordinates {
  (1,71.13) (5,71.14) (10,71.23) (15,71.19) (20,71.22) (25,71.26) (30,71.23) (35,71.22) (40,71.23) (45,71.24)
  (50,71.25) (55,71.24) (60,71.23) (65,71.22) (70,71.25) (75,71.22) (80,71.26) (85,71.24) (90,71.27) (95,71.17)
  (100,71.22) (105,71.25) (110,71.23) (115,71.24) (120,71.20) (125,71.20) (130,71.22) (135,71.23) (140,71.19) (145,71.19)
  (150,71.23) (155,71.25) (160,71.21) (165,71.23) (170,71.25) (175,71.24) (180,71.18) (185,71.25) (190,71.22) (195,71.18)
  (200,71.22) (205,71.25) (210,71.21) (215,71.20) (220,71.23)
};
\addplot[green!60!black, opacity=0.15] fill between[of=RN50_W2A4_upper and RN50_W2A4_lower];

\nextgroupplot[title={W4A4}, ymin=74.94, ymax=75.24]
\addplot[purple, mark=diamond*] coordinates {
  (1,75.07) (5,75.05) (10,75.08) (15,75.07) (20,75.07) (25,75.11) (30,75.09) (35,75.09) (40,75.09) (45,75.10)
  (50,75.11) (55,75.10) (60,75.10) (65,75.09) (70,75.11) (75,75.09) (80,75.09) (85,75.09) (90,75.09) (95,75.09)
  (100,75.11) (105,75.07) (110,75.09) (115,75.08) (120,75.09) (125,75.09) (130,75.09) (135,75.11) (140,75.09) (145,75.08)
  (150,75.09) (155,75.10) (160,75.07) (165,75.08) (170,75.08) (175,75.09) (180,75.10) (185,75.11) (190,75.10) (195,75.09)
  (200,75.09) (205,75.08) (210,75.12) (215,75.08) (220,75.08)
};
\addplot[name path=RN50_W4A4_upper, draw=none] coordinates {
  (1,75.15) (5,75.15) (10,75.16) (15,75.13) (20,75.16) (25,75.19) (30,75.16) (35,75.18) (40,75.17) (45,75.15)
  (50,75.18) (55,75.18) (60,75.17) (65,75.17) (70,75.20) (75,75.16) (80,75.17) (85,75.18) (90,75.18) (95,75.18)
  (100,75.21) (105,75.17) (110,75.19) (115,75.15) (120,75.17) (125,75.19) (130,75.18) (135,75.20) (140,75.17) (145,75.18)
  (150,75.20) (155,75.18) (160,75.16) (165,75.18) (170,75.18) (175,75.20) (180,75.18) (185,75.19) (190,75.20) (195,75.19)
  (200,75.19) (205,75.19) (210,75.22) (215,75.16) (220,75.18)
};
\addplot[name path=RN50_W4A4_lower, draw=none] coordinates {
  (1,74.99) (5,74.95) (10,75.00) (15,75.01) (20,74.98) (25,75.03) (30,75.02) (35,75.00) (40,75.01) (45,75.05)
  (50,75.04) (55,75.02) (60,75.03) (65,75.01) (70,75.02) (75,75.02) (80,75.01) (85,75.00) (90,75.00) (95,75.00)
  (100,75.01) (105,74.97) (110,74.99) (115,75.01) (120,75.01) (125,74.99) (130,75.00) (135,75.02) (140,75.01) (145,74.98)
  (150,74.98) (155,75.02) (160,74.98) (165,74.98) (170,74.98) (175,74.98) (180,75.02) (185,75.03) (190,75.00) (195,74.99)
  (200,74.99) (205,74.97) (210,75.02) (215,75.00) (220,74.98)
};
\addplot[purple, opacity=0.15] fill between[of=RN50_W4A4_upper and RN50_W4A4_lower];

\end{groupplot}

\node at ($(group c1r1.south)!0.5!(group c4r1.south)$)
      [yshift=-0.8cm, font=\normalsize] {PCA dimension};
\end{tikzpicture}
\caption{Ablation of PCA dimension $d$ for ResNet-50 under different quantization precisions (W2A2, W2A4, W4A2, W4A4). The solid line is the mean Top-1 across three runs; the shaded band shows $\pm1\sigma$. For ResNet-50, the number of clusters $K$ controls how finely CAT partitions the logit space.
Under 2-bit activations, performance improves rapidly as $K$ increases from small values and peaks
at a moderate range ($K\!\approx\!8$–$32$), after which it plateaus or slightly declines due to
over-partitioning and noisier parameter estimates. In W2A2, Top-1 rises steadily up to mid-range $K$
and then saturates; W4A2 follows a similar pattern with a slightly earlier plateau. By contrast,
W2A4 shows only mild gains across $K$, while W4A4 is essentially flat, reflecting the smaller FP–LQ
gap in higher-precision regimes. Overall, these trends indicate that a moderate cluster count strikes
the best balance between expressivity and robustness, especially when activations are quantized to 2-bit.
}
\label{fig:ablation-pca-rn50}
\end{figure}
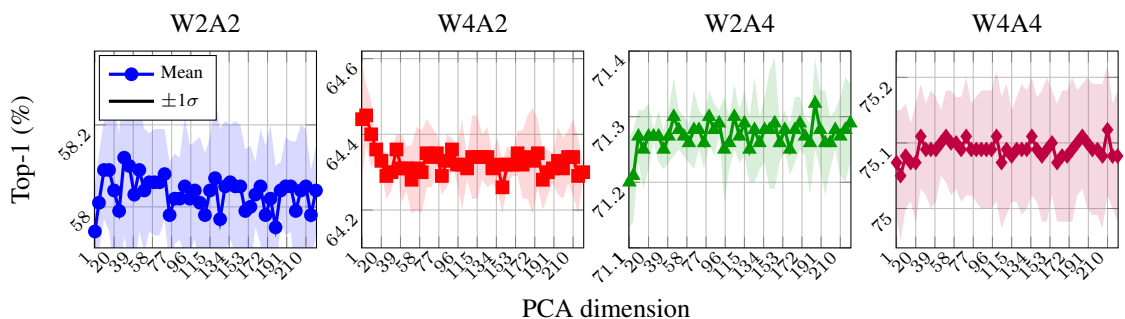


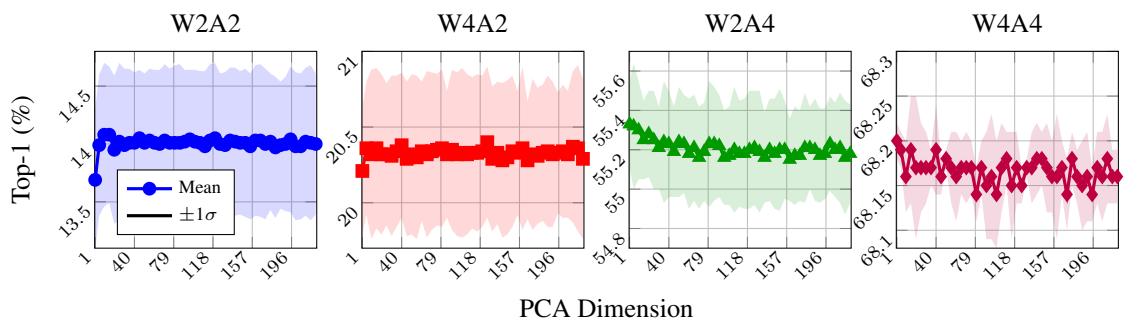
\begin{figure}[H]
\centering
\begin{tikzpicture}
\begin{groupplot}[
  group style={group size=4 by 1, horizontal sep=0.6cm},
  width=0.3\linewidth,
  height=4.2cm,
  grid=major,
  tick label style={font=\scriptsize,/pgf/number format/fixed},
  every axis plot/.append style={very thick},
  xmin=0.5, xmax=221,
  xtick={1,40,...,210},
  scaled ticks=false,
  xticklabel style={rotate=45, anchor=east},
  yticklabel style={rotate=45, anchor=east},
  title style={yshift=-0.5ex},
  legend style={cells={anchor=west}, font=\scriptsize, at={(0.1,0.4)}, anchor=north west}
]

\nextgroupplot[title={W2A2}, ymin=13.1, ymax=14.8, ylabel={Top-1 (\%)}]

\addplot[blue, mark=*] coordinates {
  (1,13.69) (5,13.99) (10,14.08) (15,14.08) (20,13.95) (25,14.02) (30,13.99) (35,14.01) (40,14.01) (45,14.05)
  (50,14.01) (55,14.03) (60,14.01) (65,14.00) (70,14.03) (75,14.01) (80,14.01) (85,14.01) (90,14.02) (95,14.04)
  (100,14.02) (105,14.01) (110,13.98) (115,14.03) (120,14.05) (125,14.00) (130,13.99) (135,14.03) (140,14.02) (145,14.01)
  (150,14.01) (155,13.98) (160,14.03) (165,14.03) (170,14.00) (175,14.02) (180,13.97) (185,13.99) (190,14.00) (195,14.04)
  (200,13.98) (205,13.98) (210,14.02) (215,14.01) (220,14.00)
};
\addlegendentry{Mean}

\addplot[name path=MBV2PCA_W2A2_upper, draw=none] coordinates {
  (1,14.30) (5,14.60) (10,14.70) (15,14.67) (20,14.59) (25,14.70) (30,14.65) (35,14.65) (40,14.62) (45,14.64)
  (50,14.61) (55,14.65) (60,14.64) (65,14.64) (70,14.69) (75,14.65) (80,14.68) (85,14.69) (90,14.64) (95,14.64)
  (100,14.64) (105,14.63) (110,14.63) (115,14.69) (120,14.70) (125,14.61) (130,14.62) (135,14.66) (140,14.66) (145,14.63)
  (150,14.65) (155,14.61) (160,14.68) (165,14.69) (170,14.65) (175,14.61) (180,14.58) (185,14.54) (190,14.58) (195,14.67)
  (200,14.62) (205,14.61) (210,14.66) (215,14.68) (220,14.61)
};
\addplot[name path=MBV2PCA_W2A2_lower, draw=none] coordinates {
  (1,13.08) (5,13.38) (10,13.46) (15,13.49) (20,13.31) (25,13.34) (30,13.33) (35,13.37) (40,13.39) (45,13.46)
  (50,13.41) (55,13.41) (60,13.38) (65,13.36) (70,13.37) (75,13.37) (80,13.34) (85,13.33) (90,13.40) (95,13.44)
  (100,13.40) (105,13.39) (110,13.33) (115,13.37) (120,13.40) (125,13.39) (130,13.36) (135,13.40) (140,13.38) (145,13.39)
  (150,13.37) (155,13.35) (160,13.38) (165,13.37) (170,13.35) (175,13.43) (180,13.36) (185,13.44) (190,13.42) (195,13.41)
  (200,13.34) (205,13.35) (210,13.38) (215,13.34) (220,13.39)
};
\addplot[blue, opacity=0.15] fill between[of=MBV2PCA_W2A2_upper and MBV2PCA_W2A2_lower];
\addlegendentry{$\pm$1$\sigma$}

\nextgroupplot[title={W4A2}, ymin=19.7, ymax=21.0]

\addplot[red, mark=square*] coordinates {
  (1,20.21) (5,20.36) (10,20.32) (15,20.36) (20,20.32) (25,20.32) (30,20.31) (35,20.34) (40,20.38) (45,20.29)
  (50,20.32) (55,20.30) (60,20.34) (65,20.32) (70,20.34) (75,20.35) (80,20.36) (85,20.32) (90,20.35) (95,20.32)
  (100,20.33) (105,20.32) (110,20.33) (115,20.33) (120,20.34) (125,20.40) (130,20.29) (135,20.33) (140,20.28) (145,20.30)
  (150,20.34) (155,20.33) (160,20.36) (165,20.28) (170,20.32) (175,20.31) (180,20.33) (185,20.34) (190,20.32) (195,20.33)
  (200,20.34) (205,20.32) (210,20.37) (215,20.36) (220,20.29)
};

\addplot[name path=MBV2PCA_W4A2_upper, draw=none] coordinates {
  (1,20.63) (5,20.80) (10,20.88) (15,20.89) (20,20.85) (25,20.79) (30,20.84) (35,20.84) (40,20.86) (45,20.80)
  (50,20.85) (55,20.84) (60,20.80) (65,20.87) (70,20.86) (75,20.85) (80,20.88) (85,20.74) (90,20.88) (95,20.86)
  (100,20.89) (105,20.85) (110,20.83) (115,20.85) (120,20.84) (125,20.90) (130,20.77) (135,20.80) (140,20.79) (145,20.81)
  (150,20.84) (155,20.74) (160,20.90) (165,20.67) (170,20.82) (175,20.76) (180,20.78) (185,20.84) (190,20.80) (195,20.73)
  (200,20.78) (205,20.74) (210,20.87) (215,20.91) (220,20.82)
};
\addplot[name path=MBV2PCA_W4A2_lower, draw=none] coordinates {
  (1,19.79) (5,19.92) (10,19.76) (15,19.83) (20,19.79) (25,19.85) (30,19.78) (35,19.84) (40,19.90) (45,19.78)
  (50,19.79) (55,19.76) (60,19.88) (65,19.77) (70,19.82) (75,19.85) (80,19.84) (85,19.90) (90,19.82) (95,19.78)
  (100,19.77) (105,19.79) (110,19.83) (115,19.81) (120,19.84) (125,19.90) (130,19.81) (135,19.86) (140,19.77) (145,19.79)
  (150,19.84) (155,19.92) (160,19.82) (165,19.89) (170,19.82) (175,19.86) (180,19.88) (185,19.84) (190,19.84) (195,19.93)
  (200,19.90) (205,19.90) (210,19.87) (215,19.81) (220,19.76)
};
\addplot[red, opacity=0.15] fill between[of=MBV2PCA_W4A2_upper and MBV2PCA_W4A2_lower];

\nextgroupplot[title={W2A4}, ymin=54.7, ymax=55.7]

\addplot[green!60!black, mark=triangle*] coordinates {
  (1,55.33) (5,55.32) (10,55.30) (15,55.25) (20,55.28) (25,55.25) (30,55.21) (35,55.24) (40,55.23) (45,55.19)
  (50,55.24) (55,55.23) (60,55.18) (65,55.21) (70,55.16) (75,55.19) (80,55.23) (85,55.23) (90,55.22) (95,55.16)
  (100,55.17) (105,55.20) (110,55.18) (115,55.18) (120,55.20) (125,55.21) (130,55.19) (135,55.16) (140,55.20) (145,55.18)
  (150,55.20) (155,55.20) (160,55.15) (165,55.17) (170,55.17) (175,55.21) (180,55.21) (185,55.20) (190,55.17) (195,55.18)
  (200,55.23) (205,55.20) (210,55.21) (215,55.16) (220,55.18)
};

\addplot[name path=MBV2PCA_W2A4_upper, draw=none] coordinates {
  (1,55.55) (5,55.64) (10,55.55) (15,55.45) (20,55.50) (25,55.50) (30,55.42) (35,55.55) (40,55.50) (45,55.42)
  (50,55.52) (55,55.48) (60,55.46) (65,55.48) (70,55.42) (75,55.44) (80,55.55) (85,55.52) (90,55.48) (95,55.40)
  (100,55.42) (105,55.52) (110,55.45) (115,55.46) (120,55.50) (125,55.50) (130,55.50) (135,55.39) (140,55.46) (145,55.46)
  (150,55.51) (155,55.45) (160,55.36) (165,55.42) (170,55.40) (175,55.49) (180,55.42) (185,55.46) (190,55.41) (195,55.41)
  (200,55.47) (205,55.44) (210,55.49) (215,55.45) (220,55.43)
};
\addplot[name path=MBV2PCA_W2A4_lower, draw=none] coordinates {
  (1,55.11) (5,55.00) (10,55.05) (15,55.05) (20,55.06) (25,55.00) (30,55.00) (35,54.93) (40,54.96) (45,54.96)
  (50,54.96) (55,54.98) (60,54.90) (65,54.94) (70,54.90) (75,54.94) (80,54.91) (85,54.94) (90,54.96) (95,54.92)
  (100,54.92) (105,54.88) (110,54.91) (115,54.90) (120,54.90) (125,54.92) (130,54.88) (135,54.93) (140,54.94) (145,54.90)
  (150,54.89) (155,54.95) (160,54.94) (165,54.92) (170,54.94) (175,54.93) (180,55.00) (185,54.94) (190,54.93) (195,54.95)
  (200,54.99) (205,54.96) (210,54.93) (215,54.87) (220,54.93)
};
\addplot[green!60!black, opacity=0.15] fill between[of=MBV2PCA_W2A4_upper and MBV2PCA_W2A4_lower];

\nextgroupplot[title={W4A4}, ymin=68.08, ymax=68.3]

\addplot[purple, mark=diamond*] coordinates {
  (1,68.20) (5,68.19) (10,68.16) (15,68.19) (20,68.17) (25,68.17) (30,68.17) (35,68.17) (40,68.19) (45,68.16)
  (50,68.18) (55,68.17) (60,68.16) (65,68.17) (70,68.17) (75,68.17) (80,68.14) (85,68.17) (90,68.15) (95,68.16)
  (100,68.14) (105,68.17) (110,68.18) (115,68.15) (120,68.17) (125,68.15) (130,68.17) (135,68.17) (140,68.18) (145,68.18)
  (150,68.17) (155,68.16) (160,68.16) (165,68.17) (170,68.14) (175,68.18) (180,68.16) (185,68.15) (190,68.16) (195,68.14)
  (200,68.17) (205,68.16) (210,68.18) (215,68.16) (220,68.16)
};

\addplot[name path=MBV2PCA_W4A4_upper, draw=none] coordinates {
  (1,68.22) (5,68.21) (10,68.20) (15,68.25) (20,68.25) (25,68.22) (30,68.21) (35,68.19) (40,68.24) (45,68.18)
  (50,68.19) (55,68.21) (60,68.21) (65,68.20) (70,68.21) (75,68.20) (80,68.17) (85,68.20) (90,68.21) (95,68.23)
  (100,68.20) (105,68.22) (110,68.21) (115,68.21) (120,68.21) (125,68.17) (130,68.18) (135,68.21) (140,68.22) (145,68.21)
  (150,68.22) (155,68.20) (160,68.23) (165,68.20) (170,68.18) (175,68.21) (180,68.19) (185,68.20) (190,68.21) (195,68.17)
  (200,68.21) (205,68.21) (210,68.23) (215,68.19) (220,68.19)
};
\addplot[name path=MBV2PCA_W4A4_lower, draw=none] coordinates {
  (1,68.18) (5,68.17) (10,68.12) (15,68.13) (20,68.09) (25,68.12) (30,68.13) (35,68.15) (40,68.14) (45,68.14)
  (50,68.17) (55,68.13) (60,68.11) (65,68.14) (70,68.13) (75,68.14) (80,68.11) (85,68.14) (90,68.09) (95,68.09)
  (100,68.08) (105,68.12) (110,68.15) (115,68.09) (120,68.13) (125,68.13) (130,68.16) (135,68.13) (140,68.14) (145,68.15)
  (150,68.12) (155,68.12) (160,68.09) (165,68.14) (170,68.10) (175,68.15) (180,68.13) (185,68.10) (190,68.11) (195,68.11)
  (200,68.13) (205,68.11) (210,68.13) (215,68.13) (220,68.13)
};
\addplot[purple, opacity=0.15] fill between[of=MBV2PCA_W4A4_upper and MBV2PCA_W4A4_lower];

\end{groupplot}

\node at ($(group c1r1.south)!0.5!(group c4r1.south)$)
      [yshift=-0.8cm, font=\normalsize] {PCA Dimension};

\end{tikzpicture}
\caption{Effect of PCA dimension on \textbf{MobileNetV2} Top-1 accuracy under different quantization settings (W2A2, W4A2, W2A4, W4A4). The solid line is the mean over three runs; the shaded area is $\pm 1\sigma$. On MobileNetV2, PCA dimension matters mainly with 2-bit activations: W2A2 peaks at small $p$
($\sim$10–15) and then plateaus, while W4A2/W2A4 change little and W4A4 is flat. A compact
subspace ($p{\approx}10$–$40$) offers a good robustness–accuracy trade-off.
}
\label{fig:ablation-pca-mbv2}
\end{figure}

\begin{figure}[H]
\centering
\begin{tikzpicture}
\begin{groupplot}[
  group style={group size=4 by 1, horizontal sep=0.6cm},
  width=0.3\linewidth,
  height=4.2cm,
  grid=major,
  tick label style={font=\scriptsize,/pgf/number format/fixed},
  every axis plot/.append style={very thick},
  xmin=0.5, xmax=221,
  xtick={1,40,...,210},
  scaled ticks=false,
  xticklabel style={rotate=45, anchor=east},
  yticklabel style={rotate=45, anchor=east},
  title style={yshift=-0.5ex},
  legend style={cells={anchor=west}, font=\scriptsize, at={(0.1,0.4)}, anchor=north west}
]

\nextgroupplot[title={W2A2}, ymin=41.0, ymax=41.6, ylabel={Top-1 (\%)}]

\addplot[blue, mark=*] coordinates {
  (1,41.19) (5,41.38) (10,41.43) (15,41.38) (20,41.43) (25,41.42) (30,41.39) (35,41.41) (40,41.36) (45,41.37)
  (50,41.38) (55,41.37) (60,41.35) (65,41.36) (70,41.37) (75,41.40) (80,41.39) (85,41.40) (90,41.33) (95,41.31)
  (100,41.32) (105,41.36) (110,41.38) (115,41.34) (120,41.36) (125,41.36) (130,41.38) (135,41.37) (140,41.31) (145,41.36)
  (150,41.38) (155,41.36) (160,41.33) (165,41.30) (170,41.43) (175,41.37) (180,41.36) (185,41.39) (190,41.39) (195,41.33)
  (200,41.36) (205,41.36) (210,41.34) (215,41.38) (220,41.38)
};
\addlegendentry{Mean}

\addplot[name path=MBV2PCA_W2A2_upper, draw=none] coordinates {
  (1,41.29) (5,41.39) (10,41.57) (15,41.54) (20,41.50) (25,41.51) (30,41.50) (35,41.48) (40,41.44) (45,41.46)
  (50,41.47) (55,41.42) (60,41.43) (65,41.46) (70,41.46) (75,41.49) (80,41.46) (85,41.48) (90,41.42) (95,41.41)
  (100,41.44) (105,41.41) (110,41.47) (115,41.39) (120,41.45) (125,41.44) (130,41.40) (135,41.50) (140,41.41) (145,41.45)
  (150,41.46) (155,41.47) (160,41.50) (165,41.40) (170,41.51) (175,41.42) (180,41.46) (185,41.51) (190,41.46) (195,41.41)
  (200,41.46) (205,41.42) (210,41.42) (215,41.50) (220,41.45)
};
\addplot[name path=MBV2PCA_W2A2_lower, draw=none] coordinates {
  (1,41.09) (5,41.37) (10,41.29) (15,41.22) (20,41.36) (25,41.33) (30,41.28) (35,41.34) (40,41.28) (45,41.28)
  (50,41.29) (55,41.32) (60,41.27) (65,41.26) (70,41.28) (75,41.31) (80,41.32) (85,41.32) (90,41.24) (95,41.21)
  (100,41.20) (105,41.31) (110,41.29) (115,41.29) (120,41.27) (125,41.28) (130,41.36) (135,41.24) (140,41.21) (145,41.27)
  (150,41.30) (155,41.25) (160,41.16) (165,41.20) (170,41.35) (175,41.32) (180,41.26) (185,41.27) (190,41.32) (195,41.25)
  (200,41.26) (205,41.30) (210,41.26) (215,41.26) (220,41.31)
};
\addplot[blue, opacity=0.15] fill between[of=MBV2PCA_W2A2_upper and MBV2PCA_W2A2_lower];
\addlegendentry{$\pm$1$\sigma$}

\nextgroupplot[title={W4A2}, ymin=50.9, ymax=51.8]

\addplot[red, mark=square*] coordinates {
  (1,51.44) (5,51.49) (10,51.45) (15,51.46) (20,51.38) (25,51.38) (30,51.36) (35,51.40) (40,51.33) (45,51.38)
  (50,51.40) (55,51.38) (60,51.38) (65,51.38) (70,51.32) (75,51.35) (80,51.31) (85,51.35) (90,51.36) (95,51.36)
  (100,51.35) (105,51.33) (110,51.37) (115,51.35) (120,51.31) (125,51.32) (130,51.35) (135,51.34) (140,51.35) (145,51.31)
  (150,51.35) (155,51.30) (160,51.34) (165,51.33) (170,51.36) (175,51.28) (180,51.33) (185,51.36) (190,51.36) (195,51.31)
  (200,51.36) (205,51.34) (210,51.29) (215,51.37) (220,51.35)
};

\addplot[name path=MBV2PCA_W4A2_upper, draw=none] coordinates {
  (1,51.72) (5,51.66) (10,51.63) (15,51.71) (20,51.64) (25,51.65) (30,51.66) (35,51.69) (40,51.63) (45,51.65)
  (50,51.63) (55,51.68) (60,51.58) (65,51.67) (70,51.61) (75,51.67) (80,51.56) (85,51.57) (90,51.61) (95,51.61)
  (100,51.65) (105,51.67) (110,51.63) (115,51.55) (120,51.59) (125,51.57) (130,51.63) (135,51.64) (140,51.59) (145,51.56)
  (150,51.67) (155,51.58) (160,51.62) (165,51.59) (170,51.64) (175,51.59) (180,51.57) (185,51.63) (190,51.63) (195,51.59)
  (200,51.67) (205,51.62) (210,51.54) (215,51.64) (220,51.62)
};
\addplot[name path=MBV2PCA_W4A2_lower, draw=none] coordinates {
  (1,51.16) (5,51.32) (10,51.27) (15,51.21) (20,51.12) (25,51.11) (30,51.06) (35,51.11) (40,51.03) (45,51.11)
  (50,51.17) (55,51.08) (60,51.18) (65,51.09) (70,51.03) (75,51.03) (80,51.06) (85,51.13) (90,51.11) (95,51.11)
  (100,51.05) (105,50.99) (110,51.11) (115,51.15) (120,51.03) (125,51.07) (130,51.07) (135,51.04) (140,51.11) (145,51.06)
  (150,51.03) (155,51.02) (160,51.06) (165,51.07) (170,51.08) (175,50.97) (180,51.09) (185,51.09) (190,51.09) (195,51.03)
  (200,51.05) (205,51.06) (210,51.04) (215,51.10) (220,51.08)
};
\addplot[red, opacity=0.15] fill between[of=MBV2PCA_W4A2_upper and MBV2PCA_W4A2_lower];

\nextgroupplot[title={W2A4}, ymin=63.6, ymax=64.7]

\addplot[green!60!black, mark=triangle*] coordinates {
  (1,64.13) (5,64.18) (10,64.22) (15,64.22) (20,64.22) (25,64.23) (30,64.25) (35,64.25) (40,64.23) (45,64.24)
  (50,64.19) (55,64.22) (60,64.21) (65,64.24) (70,64.23) (75,64.23) (80,64.21) (85,64.19) (90,64.20) (95,64.20)
  (100,64.20) (105,64.20) (110,64.20) (115,64.22) (120,64.19) (125,64.19) (130,64.20) (135,64.26) (140,64.16) (145,64.20)
  (150,64.20) (155,64.22) (160,64.21) (165,64.21) (170,64.21) (175,64.22) (180,64.18) (185,64.19) (190,64.18) (195,64.19)
  (200,64.21) (205,64.21) (210,64.19) (215,64.22) (220,64.25)
};

\addplot[name path=MBV2PCA_W2A4_upper, draw=none] coordinates {
  (1,64.44) (5,64.47) (10,64.56) (15,64.52) (20,64.53) (25,64.54) (30,64.57) (35,64.54) (40,64.55) (45,64.60)
  (50,64.48) (55,64.57) (60,64.56) (65,64.59) (70,64.56) (75,64.54) (80,64.53) (85,64.49) (90,64.47) (95,64.51)
  (100,64.46) (105,64.48) (110,64.51) (115,64.60) (120,64.51) (125,64.54) (130,64.52) (135,64.52) (140,64.47) (145,64.49)
  (150,64.50) (155,64.53) (160,64.50) (165,64.48) (170,64.54) (175,64.51) (180,64.51) (185,64.51) (190,64.51) (195,64.52)
  (200,64.55) (205,64.52) (210,64.51) (215,64.54) (220,64.54)
};
\addplot[name path=MBV2PCA_W2A4_lower, draw=none] coordinates {
  (1,63.82) (5,63.89) (10,63.88) (15,63.92) (20,63.91) (25,63.92) (30,63.93) (35,63.96) (40,63.91) (45,63.88)
  (50,63.90) (55,63.87) (60,63.86) (65,63.89) (70,63.90) (75,63.92) (80,63.89) (85,63.89) (90,63.93) (95,63.89)
  (100,63.94) (105,63.92) (110,63.89) (115,63.84) (120,63.87) (125,63.84) (130,63.88) (135,64.00) (140,63.85) (145,63.91)
  (150,63.90) (155,63.91) (160,63.92) (165,63.94) (170,63.88) (175,63.93) (180,63.85) (185,63.87) (190,63.85) (195,63.86)
  (200,63.87) (205,63.90) (210,63.87) (215,63.90) (220,63.96)
};
\addplot[green!60!black, opacity=0.15] fill between[of=MBV2PCA_W2A4_upper and MBV2PCA_W2A4_lower];

\nextgroupplot[title={W4A4}, ymin=70.92, ymax=71.1]

\addplot[purple, mark=diamond*] coordinates {
  (1,70.97) (5,71.02) (10,71.04) (15,71.01) (20,71.00) (25,71.02) (30,71.00) (35,71.01) (40,71.00) (45,70.98)
  (50,70.97) (55,70.97) (60,70.98) (65,70.98) (70,70.96) (75,70.99) (80,71.01) (85,70.98) (90,70.99) (95,70.99)
  (100,70.99) (105,70.99) (110,70.97) (115,71.00) (120,70.98) (125,70.98) (130,70.98) (135,70.97) (140,70.98) (145,70.98)
  (150,70.99) (155,70.99) (160,71.01) (165,71.00) (170,70.99) (175,70.98) (180,70.99) (185,70.99) (190,70.98) (195,70.96)
  (200,70.98) (205,70.97) (210,70.99) (215,70.98) (220,70.99)
};

\addplot[name path=MBV2PCA_W4A4_upper, draw=none] coordinates {
  (1,70.99) (5,71.06) (10,71.06) (15,71.06) (20,71.04) (25,71.06) (30,71.04) (35,71.07) (40,71.05) (45,71.03)
  (50,70.99) (55,71.02) (60,71.03) (65,71.03) (70,70.99) (75,71.02) (80,71.04) (85,71.01) (90,71.03) (95,71.01)
  (100,71.01) (105,71.02) (110,70.97) (115,71.02) (120,70.98) (125,71.01) (130,71.00) (135,71.01) (140,71.03) (145,71.02)
  (150,71.04) (155,71.01) (160,71.03) (165,71.04) (170,71.03) (175,70.99) (180,71.03) (185,71.02) (190,71.01) (195,70.99)
  (200,71.03) (205,71.00) (210,71.01) (215,71.02) (220,71.02)
};
\addplot[name path=MBV2PCA_W4A4_lower, draw=none] coordinates {
  (1,70.95) (5,70.98) (10,71.02) (15,70.96) (20,70.96) (25,70.98) (30,70.96) (35,70.95) (40,70.95) (45,70.93)
  (50,70.95) (55,70.92) (60,70.93) (65,70.93) (70,70.93) (75,70.96) (80,70.98) (85,70.95) (90,70.95) (95,70.97)
  (100,70.97) (105,70.96) (110,70.97) (115,70.98) (120,70.98) (125,70.95) (130,70.96) (135,70.93) (140,70.93) (145,70.94)
  (150,70.94) (155,70.97) (160,70.99) (165,70.96) (170,70.95) (175,70.97) (180,70.95) (185,70.96) (190,70.95) (195,70.93)
  (200,70.93) (205,70.94) (210,70.97) (215,70.94) (220,70.96)
};
\addplot[purple, opacity=0.15] fill between[of=MBV2PCA_W4A4_upper and MBV2PCA_W4A4_lower];

\end{groupplot}

\node at ($(group c1r1.south)!0.5!(group c4r1.south)$)
      [yshift=-0.8cm, font=\normalsize] {PCA Dimension};

\end{tikzpicture}
\caption{Effect of PCA dimension on \textbf{RegNetX-600MF} Top-1 accuracy under different quantization settings (W2A2, W4A2, W2A4, W4A4). The solid line is the mean over three runs; the shaded area is $\pm 1\sigma$. On RegNetX-600MF, PCA dimension matters only in W2A2, which peaks around $p{=}10$–20 and then plateaus;
W4A2, W2A4, and W4A4 remain essentially flat. Thus, small subspaces suffice for this model.
}
\label{fig:ablation-pca-regnetx600m}
\end{figure}
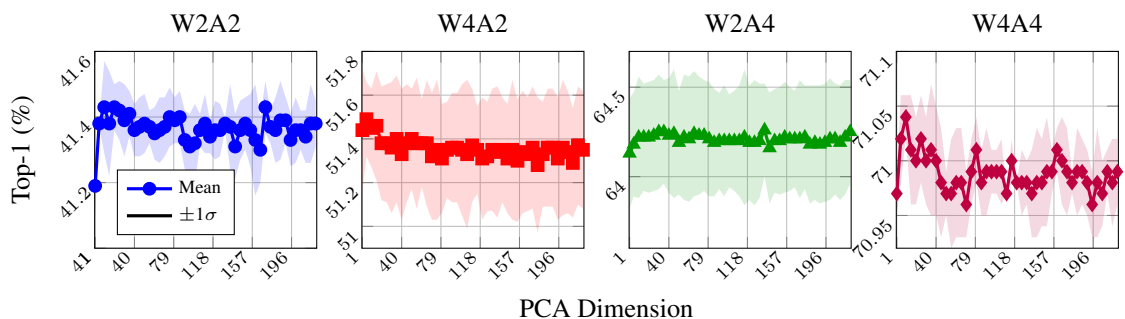

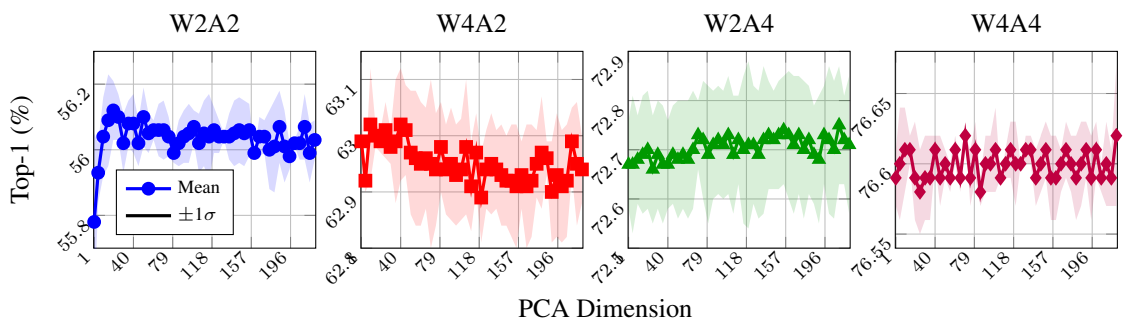
\begin{figure}[ht]
\centering
\begin{tikzpicture}
\begin{groupplot}[
  group style={group size=4 by 1, horizontal sep=0.6cm},
  width=0.3\linewidth,
  height=4.2cm,
  grid=major,
  tick label style={font=\scriptsize,/pgf/number format/fixed},
  every axis plot/.append style={very thick},
  xmin=0.5, xmax=221,
  xtick={1,40,...,210},
  scaled ticks=false,
  xticklabel style={rotate=45, anchor=east},
  yticklabel style={rotate=45, anchor=east},
  title style={yshift=-0.5ex},
  legend style={cells={anchor=west}, font=\scriptsize, at={(0.1,0.4)}, anchor=north west}
]

\nextgroupplot[title={W2A2}, ymin=55.7, ymax=56.3, ylabel={Top-1 (\%)}]

\addplot[blue, mark=*] coordinates {
  (1,55.78) (5,55.93) (10,56.04) (15,56.09) (20,56.12) (25,56.10) (30,56.02) (35,56.08) (40,56.08) (45,56.02)
  (50,56.10) (55,56.05) (60,56.06) (65,56.06) (70,56.06) (75,56.04) (80,55.99) (85,56.02) (90,56.04) (95,56.05)
  (100,56.07) (105,56.02) (110,56.05) (115,56.04) (120,56.06) (125,56.04) (130,56.04) (135,56.04) (140,56.05) (145,56.06)
  (150,56.05) (155,56.06) (160,55.99) (165,56.04) (170,56.04) (175,56.00) (180,56.01) (185,56.07) (190,56.01) (195,55.98)
  (200,56.02) (205,56.02) (210,56.07) (215,55.99) (220,56.03)
};
\addlegendentry{Mean}

\addplot[name path=MBV2PCA_W2A2_upper, draw=none] coordinates {
  (1,56.05) (5,56.02) (10,56.16) (15,56.23) (20,56.21) (25,56.17) (30,56.11) (35,56.15) (40,56.17) (45,56.09)
  (50,56.19) (55,56.11) (60,56.09) (65,56.14) (70,56.15) (75,56.10) (80,56.02) (85,56.09) (90,56.12) (95,56.14)
  (100,56.12) (105,56.11) (110,56.15) (115,56.17) (120,56.13) (125,56.09) (130,56.05) (135,56.11) (140,56.15) (145,56.16)
  (150,56.15) (155,56.14) (160,55.99) (165,56.10) (170,56.09) (175,56.12) (180,56.15) (185,56.18) (190,56.12) (195,56.02)
  (200,56.03) (205,56.07) (210,56.17) (215,56.08) (220,56.09)
};
\addplot[name path=MBV2PCA_W2A2_lower, draw=none] coordinates {
  (1,55.51) (5,55.84) (10,55.92) (15,55.95) (20,56.03) (25,56.03) (30,55.93) (35,56.01) (40,55.99) (45,55.95)
  (50,56.01) (55,55.99) (60,56.03) (65,55.98) (70,55.97) (75,55.98) (80,55.96) (85,55.95) (90,55.96) (95,55.96)
  (100,56.02) (105,55.93) (110,55.95) (115,55.91) (120,55.99) (125,55.99) (130,56.03) (135,55.97) (140,55.95) (145,55.96)
  (150,55.95) (155,55.98) (160,55.99) (165,55.98) (170,55.99) (175,55.88) (180,55.87) (185,55.96) (190,55.90) (195,55.94)
  (200,56.01) (205,55.97) (210,55.97) (215,55.90) (220,55.97)
};
\addplot[blue, opacity=0.15] fill between[of=MBV2PCA_W2A2_upper and MBV2PCA_W2A2_lower];
\addlegendentry{$\pm$1$\sigma$}

\nextgroupplot[title={W4A2}, ymin=62.8, ymax=63.15]

\addplot[red, mark=square*] coordinates {
  (1,62.99) (5,62.92) (10,63.02) (15,63.00) (20,62.99) (25,63.01) (30,62.98) (35,62.99) (40,63.02) (45,63.01)
  (50,62.97) (55,62.96) (60,62.95) (65,62.96) (70,62.95) (75,62.94) (80,62.98) (85,62.94) (90,62.95) (95,62.93)
  (100,62.94) (105,62.98) (110,62.91) (115,62.97) (120,62.89) (125,62.94) (130,62.95) (135,62.94) (140,62.93) (145,62.92)
  (150,62.92) (155,62.91) (160,62.93) (165,62.91) (170,62.92) (175,62.95) (180,62.97) (185,62.96) (190,62.90) (195,62.93)
  (200,62.91) (205,62.92) (210,62.99) (215,62.95) (220,62.94)
};

\addplot[name path=MBV2PCA_W4A2_upper, draw=none] coordinates {
  (1,63.04) (5,62.94) (10,63.12) (15,63.02) (20,63.01) (25,63.06) (30,63.01) (35,63.10) (40,63.12) (45,63.09)
  (50,63.08) (55,63.08) (60,63.06) (65,63.05) (70,63.10) (75,63.04) (80,63.06) (85,63.02) (90,63.05) (95,63.02)
  (100,63.03) (105,63.10) (110,62.99) (115,63.04) (120,62.97) (125,63.01) (130,63.04) (135,63.01) (140,63.02) (145,63.00)
  (150,63.02) (155,63.02) (160,62.99) (165,63.02) (170,63.02) (175,63.01) (180,63.07) (185,62.99) (190,62.99) (195,63.04)
  (200,63.01) (205,63.02) (210,63.05) (215,63.06) (220,63.00)
};
\addplot[name path=MBV2PCA_W4A2_lower, draw=none] coordinates {
  (1,62.94) (5,62.90) (10,62.92) (15,62.98) (20,62.97) (25,62.96) (30,62.95) (35,62.88) (40,62.92) (45,62.93)
  (50,62.86) (55,62.84) (60,62.84) (65,62.87) (70,62.80) (75,62.84) (80,62.90) (85,62.86) (90,62.85) (95,62.84)
  (100,62.85) (105,62.86) (110,62.83) (115,62.90) (120,62.81) (125,62.87) (130,62.86) (135,62.87) (140,62.84) (145,62.84)
  (150,62.82) (155,62.80) (160,62.87) (165,62.80) (170,62.82) (175,62.89) (180,62.87) (185,62.93) (190,62.81) (195,62.82)
  (200,62.81) (205,62.82) (210,62.93) (215,62.84) (220,62.88)
};
\addplot[red, opacity=0.15] fill between[of=MBV2PCA_W4A2_upper and MBV2PCA_W4A2_lower];

\nextgroupplot[title={W2A4}, ymin=72.5, ymax=72.9]

\addplot[green!60!black, mark=triangle*] coordinates {
  (1,72.67) (5,72.67) (10,72.68) (15,72.69) (20,72.70) (25,72.66) (30,72.69) (35,72.67) (40,72.67) (45,72.69)
  (50,72.68) (55,72.69) (60,72.68) (65,72.70) (70,72.73) (75,72.72) (80,72.69) (85,72.71) (90,72.72) (95,72.71)
  (100,72.73) (105,72.69) (110,72.72) (115,72.70) (120,72.71) (125,72.71) (130,72.69) (135,72.72) (140,72.73) (145,72.72)
  (150,72.73) (155,72.74) (160,72.73) (165,72.71) (170,72.73) (175,72.70) (180,72.72) (185,72.69) (190,72.68) (195,72.73)
  (200,72.72) (205,72.70) (210,72.75) (215,72.72) (220,72.71)
};

\addplot[name path=MBV2PCA_W2A4_upper, draw=none] coordinates {
  (1,72.76) (5,72.81) (10,72.77) (15,72.81) (20,72.82) (25,72.77) (30,72.80) (35,72.79) (40,72.76) (45,72.79)
  (50,72.80) (55,72.80) (60,72.80) (65,72.82) (70,72.80) (75,72.84) (80,72.83) (85,72.82) (90,72.83) (95,72.82)
  (100,72.85) (105,72.81) (110,72.81) (115,72.85) (120,72.88) (125,72.83) (130,72.82) (135,72.87) (140,72.84) (145,72.82)
  (150,72.84) (155,72.85) (160,72.86) (165,72.85) (170,72.85) (175,72.84) (180,72.83) (185,72.79) (190,72.81) (195,72.83)
  (200,72.87) (205,72.84) (210,72.87) (215,72.81) (220,72.85)
};
\addplot[name path=MBV2PCA_W2A4_lower, draw=none] coordinates {
  (1,72.58) (5,72.53) (10,72.59) (15,72.57) (20,72.58) (25,72.55) (30,72.58) (35,72.55) (40,72.58) (45,72.59)
  (50,72.56) (55,72.58) (60,72.56) (65,72.58) (70,72.66) (75,72.60) (80,72.55) (85,72.60) (90,72.61) (95,72.60)
  (100,72.61) (105,72.57) (110,72.63) (115,72.55) (120,72.54) (125,72.59) (130,72.56) (135,72.57) (140,72.62) (145,72.62)
  (150,72.62) (155,72.63) (160,72.60) (165,72.57) (170,72.61) (175,72.56) (180,72.61) (185,72.59) (190,72.55) (195,72.63)
  (200,72.57) (205,72.56) (210,72.63) (215,72.63) (220,72.57)
};
\addplot[green!60!black, opacity=0.15] fill between[of=MBV2PCA_W2A4_upper and MBV2PCA_W2A4_lower];

\nextgroupplot[title={W4A4}, ymin=76.54, ymax=76.68]

\addplot[purple, mark=diamond*] coordinates {
  (1,76.59) (5,76.60) (10,76.61) (15,76.61) (20,76.59) (25,76.58) (30,76.59) (35,76.59) (40,76.61) (45,76.59)
  (50,76.60) (55,76.59) (60,76.61) (65,76.59) (70,76.62) (75,76.59) (80,76.61) (85,76.58) (90,76.60) (95,76.60)
  (100,76.61) (105,76.59) (110,76.60) (115,76.61) (120,76.60) (125,76.60) (130,76.61) (135,76.61) (140,76.59) (145,76.60)
  (150,76.61) (155,76.59) (160,76.59) (165,76.60) (170,76.61) (175,76.60) (180,76.59) (185,76.60) (190,76.61) (195,76.59)
  (200,76.61) (205,76.59) (210,76.60) (215,76.59) (220,76.62)
};

\addplot[name path=MBV2PCA_W4A4_upper, draw=none] coordinates {
  (1,76.61) (5,76.64) (10,76.64) (15,76.62) (20,76.63) (25,76.61) (30,76.62) (35,76.62) (40,76.62) (45,76.62)
  (50,76.61) (55,76.60) (60,76.62) (65,76.60) (70,76.62) (75,76.61) (80,76.63) (85,76.59) (90,76.62) (95,76.62)
  (100,76.64) (105,76.59) (110,76.61) (115,76.63) (120,76.63) (125,76.60) (130,76.62) (135,76.62) (140,76.61) (145,76.62)
  (150,76.63) (155,76.62) (160,76.62) (165,76.61) (170,76.63) (175,76.62) (180,76.62) (185,76.63) (190,76.63) (195,76.62)
  (200,76.63) (205,76.60) (210,76.63) (215,76.61) (220,76.67)
};
\addplot[name path=MBV2PCA_W4A4_lower, draw=none] coordinates {
  (1,76.57) (5,76.56) (10,76.58) (15,76.60) (20,76.56) (25,76.55) (30,76.56) (35,76.56) (40,76.60) (45,76.57)
  (50,76.59) (55,76.58) (60,76.60) (65,76.58) (70,76.62) (75,76.57) (80,76.59) (85,76.57) (90,76.58) (95,76.58)
  (100,76.58) (105,76.59) (110,76.59) (115,76.59) (120,76.57) (125,76.60) (130,76.60) (135,76.60) (140,76.57) (145,76.58)
  (150,76.59) (155,76.56) (160,76.56) (165,76.58) (170,76.59) (175,76.58) (180,76.56) (185,76.57) (190,76.59) (195,76.56)
  (200,76.59) (205,76.58) (210,76.57) (215,76.57) (220,76.57)
};
\addplot[purple, opacity=0.15] fill between[of=MBV2PCA_W4A4_upper and MBV2PCA_W4A4_lower];

\end{groupplot}

\node at ($(group c1r1.south)!0.5!(group c4r1.south)$)
      [yshift=-0.8cm, font=\normalsize] {PCA Dimension};

\end{tikzpicture}
\caption{Effect of PCA dimension on \textbf{RegNetX-3.2GF} Top-1 accuracy under different quantization settings (W2A2, W4A2, W2A4, W4A4). The solid line is the mean over three runs; the shaded area is $\pm 1\sigma$. On RegNetX-3.2GF, W2A2 benefits slightly from small $p$ (peaking near $p{\approx}20$), while W4A2, W2A4,
and W4A4 are essentially flat. A compact PCA subspace ($p{\approx}10$–$40$) suffices for robust performance.
}
\label{fig:ablation-pca-regnetx3200m}
\end{figure}
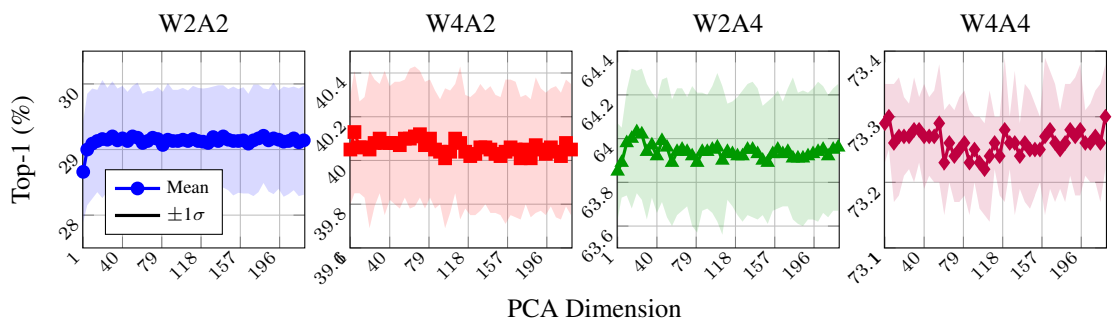
\begin{figure}[H]
\centering
\begin{tikzpicture}
\begin{groupplot}[
  group style={group size=4 by 1, horizontal sep=0.6cm},
  width=0.3\linewidth,
  height=4.2cm,
  grid=major,
  tick label style={font=\scriptsize,/pgf/number format/fixed},
  every axis plot/.append style={very thick},
  xmin=0.5, xmax=221,
  xtick={1,40,...,210},
  scaled ticks=false,
  xticklabel style={rotate=45, anchor=east},
  yticklabel style={rotate=45, anchor=east},
  title style={yshift=-0.5ex},
  legend style={cells={anchor=west}, font=\scriptsize, at={(0.1,0.4)}, anchor=north west}
]

\nextgroupplot[title={W2A2}, ymin=27.5, ymax=30.5, ylabel={Top-1 (\%)}]

\addplot[blue, mark=*] coordinates {%
  (1,28.66) (5,29.00) (10,29.09) (15,29.13) (20,29.16) (25,29.15) (30,29.20) (35,29.14) (40,29.17) (45,29.13)
  (50,29.20) (55,29.18) (60,29.10) (65,29.13) (70,29.18) (75,29.16) (80,29.07) (85,29.15) (90,29.13) (95,29.13)
  (100,29.15) (105,29.13) (110,29.16) (115,29.13) (120,29.12) (125,29.10) (130,29.19) (135,29.13) (140,29.20) (145,29.15)
  (150,29.13) (155,29.13) (160,29.14) (165,29.09) (170,29.14) (175,29.17) (180,29.21) (185,29.14) (190,29.17) (195,29.15)
  (200,29.12) (205,29.13) (210,29.17) (215,29.11) (220,29.14)
};
\addlegendentry{Mean}

\addplot[name path=MBV2PCA_W2A2_upper, draw=none] coordinates {%
  (1,29.43) (5,29.85) (10,29.93) (15,29.91) (20,30.06) (25,29.94) (30,30.04) (35,29.93) (40,29.97) (45,29.88)
  (50,29.97) (55,30.02) (60,29.85) (65,29.89) (70,29.89) (75,29.95) (80,29.88) (85,30.01) (90,29.91) (95,29.91)
  (100,29.94) (105,29.90) (110,29.97) (115,29.94) (120,29.93) (125,29.94) (130,29.96) (135,29.98) (140,29.98) (145,29.92)
  (150,29.96) (155,29.92) (160,29.95) (165,29.85) (170,29.99) (175,29.99) (180,29.93) (185,29.92) (190,30.02) (195,29.90)
  (200,29.96) (205,29.94) (210,29.96) (215,29.93) (220,29.97)
};
\addplot[name path=MBV2PCA_W2A2_lower, draw=none] coordinates {%
  (1,27.89) (5,28.15) (10,28.25) (15,28.35) (20,28.26) (25,28.36) (30,28.36) (35,28.35) (40,28.37) (45,28.38)
  (50,28.43) (55,28.34) (60,28.35) (65,28.37) (70,28.47) (75,28.37) (80,28.26) (85,28.29) (90,28.35) (95,28.35)
  (100,28.36) (105,28.36) (110,28.35) (115,28.32) (120,28.31) (125,28.26) (130,28.42) (135,28.28) (140,28.42) (145,28.38)
  (150,28.30) (155,28.34) (160,28.33) (165,28.33) (170,28.29) (175,28.35) (180,28.49) (185,28.36) (190,28.32) (195,28.40)
  (200,28.28) (205,28.32) (210,28.38) (215,28.29) (220,28.31)
};
\addplot[blue, opacity=0.15] fill between[of=MBV2PCA_W2A2_upper and MBV2PCA_W2A2_lower];
\addlegendentry{$\pm$1$\sigma$}

\nextgroupplot[title={W4A2}, ymin=39.6, ymax=40.5]

\addplot[red, mark=square*] coordinates {%
  (1,40.05) (5,40.13) (10,40.06) (15,40.06) (20,40.05) (25,40.08) (30,40.10) (35,40.08) (40,40.08) (45,40.08)
  (50,40.07) (55,40.10) (60,40.10) (65,40.11) (70,40.12) (75,40.07) (80,40.10) (85,40.05) (90,40.04) (95,40.01)
  (100,40.04) (105,40.10) (110,40.08) (115,40.04) (120,40.02) (125,40.03) (130,40.06) (135,40.06) (140,40.05) (145,40.03)
  (150,40.02) (155,40.04) (160,40.06) (165,40.05) (170,40.01) (175,40.05) (180,40.01) (185,40.07) (190,40.05) (195,40.03)
  (200,40.06) (205,40.04) (210,40.02) (215,40.08) (220,40.05)
};

\addplot[name path=MBV2PCA_W4A2_upper, draw=none] coordinates {%
  (1,40.30) (5,40.41) (10,40.27) (15,40.29) (20,40.41) (25,40.37) (30,40.35) (35,40.41) (40,40.34) (45,40.36)
  (50,40.36) (55,40.37) (60,40.42) (65,40.43) (70,40.41) (75,40.36) (80,40.37) (85,40.32) (90,40.28) (95,40.33)
  (100,40.32) (105,40.35) (110,40.43) (115,40.30) (120,40.29) (125,40.31) (130,40.35) (135,40.32) (140,40.34) (145,40.30)
  (150,40.29) (155,40.37) (160,40.37) (165,40.36) (170,40.29) (175,40.30) (180,40.28) (185,40.38) (190,40.33) (195,40.31)
  (200,40.33) (205,40.34) (210,40.31) (215,40.37) (220,40.35)
};
\addplot[name path=MBV2PCA_W4A2_lower, draw=none] coordinates {%
  (1,39.80) (5,39.85) (10,39.85) (15,39.83) (20,39.69) (25,39.79) (30,39.85) (35,39.75) (40,39.82) (45,39.80)
  (50,39.78) (55,39.83) (60,39.78) (65,39.79) (70,39.83) (75,39.78) (80,39.83) (85,39.78) (90,39.80) (95,39.69)
  (100,39.76) (105,39.85) (110,39.73) (115,39.78) (120,39.75) (125,39.75) (130,39.77) (135,39.80) (140,39.76) (145,39.76)
  (150,39.75) (155,39.71) (160,39.75) (165,39.74) (170,39.73) (175,39.80) (180,39.74) (185,39.76) (190,39.77) (195,39.75)
  (200,39.79) (205,39.74) (210,39.73) (215,39.79) (220,39.75)
};
\addplot[red, opacity=0.15] fill between[of=MBV2PCA_W4A2_upper and MBV2PCA_W4A2_lower];

\nextgroupplot[title={W2A4}, ymin=63.5, ymax=64.4]

\addplot[green!60!black, mark=triangle*] coordinates {%
  (1,63.85) (5,63.89) (10,63.98) (15,64.00) (20,64.03) (25,64.02) (30,63.94) (35,63.97) (40,63.92) (45,63.99)
  (50,63.96) (55,63.89) (60,63.94) (65,63.95) (70,63.94) (75,63.92) (80,63.89) (85,63.94) (90,63.94) (95,63.95)
  (100,63.96) (105,63.90) (110,63.94) (115,63.93) (120,63.92) (125,63.92) (130,63.95) (135,63.95) (140,63.93) (145,63.90)
  (150,63.89) (155,63.93) (160,63.95) (165,63.93) (170,63.94) (175,63.91) (180,63.91) (185,63.91) (190,63.92) (195,63.93)
  (200,63.94) (205,63.95) (210,63.92) (215,63.95) (220,63.96)
};

\addplot[name path=MBV2PCA_W2A4_upper, draw=none] coordinates {%
  (1,64.07) (5,64.05) (10,64.22) (15,64.32) (20,64.31) (25,64.32) (30,64.25) (35,64.21) (40,64.18) (45,64.27)
  (50,64.28) (55,64.18) (60,64.23) (65,64.23) (70,64.19) (75,64.24) (80,64.21) (85,64.21) (90,64.20) (95,64.24)
  (100,64.30) (105,64.24) (110,64.16) (115,64.17) (120,64.19) (125,64.22) (130,64.25) (135,64.26) (140,64.23) (145,64.15)
  (150,64.17) (155,64.19) (160,64.22) (165,64.17) (170,64.23) (175,64.28) (180,64.21) (185,64.22) (190,64.21) (195,64.23)
  (200,64.26) (205,64.24) (210,64.20) (215,64.23) (220,64.25)
};
\addplot[name path=MBV2PCA_W2A4_lower, draw=none] coordinates {%
  (1,63.63) (5,63.73) (10,63.74) (15,63.68) (20,63.75) (25,63.72) (30,63.63) (35,63.73) (40,63.66) (45,63.71)
  (50,63.64) (55,63.60) (60,63.65) (65,63.67) (70,63.69) (75,63.60) (80,63.57) (85,63.67) (90,63.68) (95,63.66)
  (100,63.62) (105,63.56) (110,63.72) (115,63.69) (120,63.65) (125,63.62) (130,63.65) (135,63.64) (140,63.63) (145,63.65)
  (150,63.61) (155,63.67) (160,63.68) (165,63.69) (170,63.69) (175,63.63) (180,63.61) (185,63.60) (190,63.63) (195,63.63)
  (200,63.62) (205,63.66) (210,63.64) (215,63.67) (220,63.67)
};
\addplot[green!60!black, opacity=0.15] fill between[of=MBV2PCA_W2A4_upper and MBV2PCA_W2A4_lower];

\nextgroupplot[title={W4A4}, ymin=73.1, ymax=73.4]

\addplot[purple, mark=diamond*] coordinates {%
  (1,73.29) (5,73.30) (10,73.26) (15,73.27) (20,73.27) (25,73.27) (30,73.28) (35,73.28) (40,73.27) (45,73.27)
  (50,73.27) (55,73.29) (60,73.23) (65,73.26) (70,73.24) (75,73.25) (80,73.26) (85,73.23) (90,73.25) (95,73.23)
  (100,73.22) (105,73.24) (110,73.26) (115,73.24) (120,73.28) (125,73.26) (130,73.26) (135,73.24) (140,73.26) (145,73.25)
  (150,73.25) (155,73.25) (160,73.27) (165,73.28) (170,73.26) (175,73.25) (180,73.26) (185,73.28) (190,73.27) (195,73.28)
  (200,73.26) (205,73.26) (210,73.27) (215,73.26) (220,73.30)
};

\addplot[name path=MBV2PCA_W4A4_upper, draw=none] coordinates {%
  (1,73.35) (5,73.35) (10,73.35) (15,73.36) (20,73.33) (25,73.32) (30,73.36) (35,73.33) (40,73.35) (45,73.33)
  (50,73.34) (55,73.38) (60,73.31) (65,73.32) (70,73.31) (75,73.33) (80,73.34) (85,73.31) (90,73.30) (95,73.30)
  (100,73.29) (105,73.34) (110,73.30) (115,73.30) (120,73.39) (125,73.30) (130,73.37) (135,73.29) (140,73.35) (145,73.34)
  (150,73.32) (155,73.30) (160,73.39) (165,73.37) (170,73.35) (175,73.31) (180,73.35) (185,73.35) (190,73.36) (195,73.38)
  (200,73.31) (205,73.32) (210,73.34) (215,73.35) (220,73.38)
};
\addplot[name path=MBV2PCA_W4A4_lower, draw=none] coordinates {%
  (1,73.23) (5,73.25) (10,73.17) (15,73.18) (20,73.21) (25,73.22) (30,73.20) (35,73.23) (40,73.19) (45,73.21)
  (50,73.20) (55,73.20) (60,73.15) (65,73.20) (70,73.17) (75,73.17) (80,73.18) (85,73.15) (90,73.20) (95,73.16)
  (100,73.15) (105,73.14) (110,73.22) (115,73.18) (120,73.17) (125,73.22) (130,73.15) (135,73.19) (140,73.17) (145,73.16)
  (150,73.18) (155,73.20) (160,73.15) (165,73.19) (170,73.17) (175,73.19) (180,73.17) (185,73.21) (190,73.18) (195,73.18)
  (200,73.21) (205,73.20) (210,73.20) (215,73.17) (220,73.22)
};
\addplot[purple, opacity=0.15] fill between[of=MBV2PCA_W4A4_upper and MBV2PCA_W4A4_lower];

\end{groupplot}

\node at ($(group c1r1.south)!0.5!(group c4r1.south)$)
      [yshift=-0.8cm, font=\normalsize] {PCA Dimension};

\end{tikzpicture}
\caption{Effect of PCA dimension on \textbf{MNasX2} Top-1 accuracy under different quantization settings (W2A2, W4A2, W2A4, W4A4). The solid line is the mean over three runs; the shaded area is $\pm 1\sigma$. On MNasX2, W2A2 gains slightly up to $p{\approx}30$ and then plateaus; W4A2, W2A4, and W4A4 are
essentially flat. Thus, a compact PCA subspace ($p{\approx}10$–$40$) suffices. 
}
\label{fig:ablation-pca-MNasX2}
\end{figure}
\newpage
\section{Ablation on CAT fitting sample size}
\label{Section:smaples_app}
\begin{figure}[ht]
\centering
\begin{tikzpicture}
\begin{groupplot}[
  group style={group size=4 by 1, horizontal sep=0.6cm},
  width=0.30\linewidth,
  height=4.2cm,
  grid=major,
  tick label style={font=\scriptsize,/pgf/number format/fixed},
  every axis plot/.append style={very thick},
  xmin=-10, xmax=1100,
  xtick={10,100,500,1000},
  scaled ticks=false,
  xticklabel style={rotate=45, anchor=east},
  yticklabel style={rotate=45, anchor=east},
  title style={yshift=-0.5ex},
  legend style={cells={anchor=west}, font=\scriptsize, at={(0.2,0.5)}, anchor=north west}
]

\nextgroupplot[title={W2A2}, ymin=46.2, ymax=53.5, ylabel={Top-1 (\%)}]

\addplot[blue, mark=*] coordinates {
  (10,46.48) (50,51.79) (100,52.49) (500,53.02) (1000,53.04)(10000,53.14) (100000,53.15)
};
\addlegendentry{Mean}

\addplot[name path=W2A2U, draw=none] coordinates {
  (10,46.66) (50,51.89) (100,52.53) (500,53.10) (1000,53.11) (10000,53.04)(100000,53.04)
};
\addplot[name path=W2A2L, draw=none] coordinates {
  (10,46.30) (50,51.69) (100,52.45) (500,52.94) (1000,52.97) (10000,53.24)(100000, 53.26)
};
\addplot[blue, opacity=0.15] fill between[of=W2A2U and W2A2L];
\addlegendentry{$\pm 1\sigma$}

\nextgroupplot[title={W4A2}, ymin=53.2, ymax=59]

\addplot[red, mark=square*] coordinates {
  (10,53.67) (50,57.61) (100,58.09) (500,58.48) (1000,58.54)(10000,58.57)
};

\addplot[name path=W4A2U, draw=none] coordinates {
  (10,54.11) (50,57.77) (100,58.29) (500,58.67) (1000,58.71)(10000, 58.39)
};
\addplot[name path=W4A2L, draw=none] coordinates {
  (10,53.23) (50,57.45) (100,57.89) (500,58.29) (1000,58.37)(10000,58.75)
};
\addplot[red, opacity=0.15] fill between[of=W4A2U and W4A2L];

\nextgroupplot[title={W2A4}, ymin=62.4, ymax=65.5]

\addplot[green!60!black, mark=triangle*] coordinates {
  (10,62.60) (50,64.85) (100,65.03) (500,65.20) (1000,65.33)(10000,65.28)
};

\addplot[name path=W2A4U, draw=none] coordinates {
  (10,62.62) (50,64.91) (100,65.13) (500,65.30) (1000,65.41)(10000,65.19)
};
\addplot[name path=W2A4L, draw=none] coordinates {
  (10,62.58) (50,64.79) (100,64.93) (500,65.10) (1000,65.25)(10000,65.37)
};
\addplot[green!60!black, opacity=0.15] fill between[of=W2A4U and W2A4L];

\nextgroupplot[title={W4A4}, ymin=67.9, ymax=69.4]

\addplot[purple, mark=diamond*] coordinates {
  (10,68.13) (50,69.12) (100,69.20) (500,69.27) (1000,69.26)(10000,69.3)
};

\addplot[name path=W4A4U, draw=none] coordinates {
  (10,68.22) (50,69.18) (100,69.25) (500,69.33) (1000,69.32)(10000,69.21)
};
\addplot[name path=W4A4L, draw=none] coordinates {
  (10,68.04) (50,69.06) (100,69.15) (500,69.21) (1000,69.20)(10000,69.29)
};
\addplot[purple, opacity=0.15] fill between[of=W4A4U and W4A4L];

\end{groupplot}

\node at ($(group c1r1.south)!0.5!(group c4r1.south)$)
      [yshift=-0.8cm, font=\normalsize] {\#Samples };

\end{tikzpicture}
\caption{Ablation on the number of samples for ResNet-18 under different quantization settings (W2A2, W4A2, W2A4, W4A4). Solid lines show mean Top-1 accuracy; the shaded region is $\pm 1\sigma$. For ResNet-18, increasing the number of fitting samples yields clear gains that saturate
once a few hundred to a thousand samples are used. In W2A2, Top-1 rises sharply from
46.5\% at 10 samples to 53.0–53.1\% by 500–1{,}000 samples, with only marginal changes
thereafter (53.14–53.15\% at 10k–100k). W4A2 follows the same trend, improving from
53.7\% (10) to 58.5\% (1k–10k) with diminishing returns beyond 500 samples. In W2A4,
accuracy increases from 62.6\% (10) to 65.3\% (1k) and flattens around 65.2–65.3\%.
W4A4 similarly moves from 68.1\% (10) to 69.3\% (0.5k–10k) and then saturates. Overall,
CAT achieves most of its benefit with \(\sim\)500–1{,}000 samples, indicating that modest
data budgets suffice for effective post-training error restoration.
}
\label{fig:samples-ablation-rn18}
\end{figure}
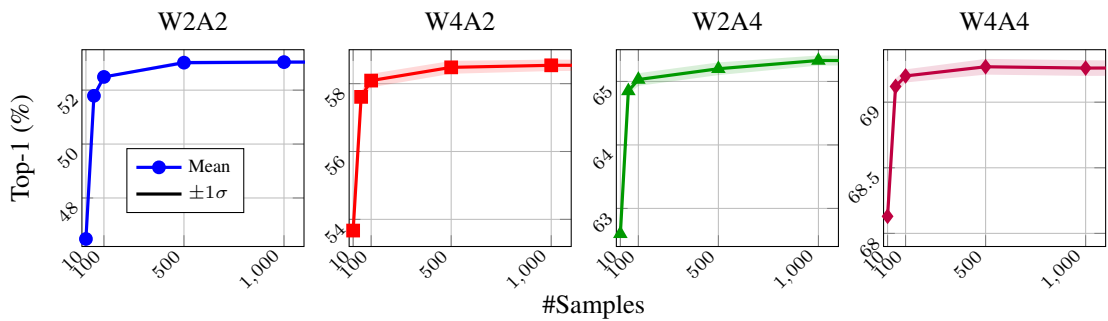

\newpage
\newpage
\section{CAT improvement on ViT quantization  via different PTQ methods}
\label{Section:ViT}
\begin{table}[ht]
\setlength{\tabcolsep}{3pt}
\renewcommand{\arraystretch}{0.9}
\small
\centering
\caption{Top-1 accuracy (\%) for BRECQ and QDrop with/without CAT under W2A2 and W4A4.}
\label{tab:supercols_superrows_cat}
\begin{tabular}{ll*{6}{c}*{6}{c}}
\toprule
& & \multicolumn{6}{c}{\textbf{W2A2}} & \multicolumn{6}{c}{\textbf{W4A4}} \\
\cmidrule(lr){3-8}\cmidrule(lr){9-14}
\textbf{Method} & \textbf{Variant}
& ViT-S & ViT-B & Swin-S & Swin-B & DeiT-S & DeiT-B
& ViT-S & ViT-B & Swin-S & Swin-B & DeiT-S & DeiT-B \\
\midrule
\multirow{2}{*}{\textbf{BRECQ}}
  &Base & 0.63 & 2.042 & 3.43 & 4.1 & 4.84 & 12.46 & 73.96 & 79.23 & 81.07 & 82.6 & 69.82 & 77.64 \\
  &+ CAT            & 0.75 & 2.17 & 3.52 & 4.27 & 5.04 & 12.79 & 74.09 & 79.42 & 81.11 & 82.65 & 70.09 & 77.7 \\
\midrule
\multirow{2}{*}{\textbf{QDrop}}
  &Base & 3.36 & 4.77 & 12.85 & 15.69 & 4.1 & 30.53 & 69.51 & 81.96 & 80.79 & 82.25 & 69.93 & 78.01 \\
  &+ CAT            & 3.34 & 4.83 & 12.83 & 15.77 & 4.27 & 30.72 & 69.54 & 82 & 80.86 & 82.30 & 70.13 & 78.05 \\
\bottomrule
\end{tabular}
\vspace{-2mm}
\end{table}
\cref{tab:supercols_superrows_cat} reports Top-1 accuracy for ViT, Swin, and DeiT models under W2A2 and W4A4 quantization with BRECQ and QDrop. CAT consistently improves performance across all settings. The gains are most notable in ultra-low precision (W2A2), with improvements up to +0.3\%–0.4\%, while W4A4 shows smaller but steady benefits, confirming CAT’s generalizability across architectures and quantization methods.
\begin{table}[ht]
\setlength{\tabcolsep}{3pt}
\renewcommand{\arraystretch}{0.95}
\small
\centering
\caption{Top-1 accuracy (\%) for BRECQ and QDrop with/without CAT on DeiT-S and DeiT-B under different precisions.}
\label{tab:deit_cat_supercols}
\begin{tabular}{ll*{2}{c}*{2}{c}*{2}{c}*{2}{c}*{2}{c}}
\toprule
& & \multicolumn{2}{c}{\textbf{W2A2}}
  & \multicolumn{2}{c}{\textbf{W4A4}}
  & \multicolumn{2}{c}{\textbf{W2A6}}
  & \multicolumn{2}{c}{\textbf{W4A6}}
  & \multicolumn{2}{c}{\textbf{W6A6}} \\
\cmidrule(lr){3-4}\cmidrule(lr){5-6}\cmidrule(lr){7-8}\cmidrule(lr){9-10}\cmidrule(lr){11-12}
\textbf{Method} & \textbf{Variant} 
& DeiT-S & DeiT-B 
& DeiT-S & DeiT-B
& DeiT-S & DeiT-B
& DeiT-S & DeiT-B
& DeiT-S & DeiT-B \\
\midrule
\multirow{2}{*}{\textbf{BRECQ}}
  &Base   & 4.84 & 12.46 & 69.82 & 77.64 & 65.25 & 75.22 & 77.64 & 80.83 & 76.7 & 80.49 \\
  &+ CAT  & 5.04 & 12.79 & 70.09 & 77.70 & 65.30 & 75.28 & 77.65 & 80.9 & 76.87 &  80.5 \\
\midrule
\multirow{2}{*}{\textbf{QDrop}}
  &Base   & 4.10 & 30.53 & 69.93 & 78.01 & 65.71 & 76.0 & 77.9 & 80.9 & 77.74 & 80.82 \\
  &+ CAT  & 4.27 & 30.72 & 70.13 & 78.05 & 66.04 & 76.11 & 77.95 & 80.94 & 77.82 & 80.86 \\
\bottomrule
\end{tabular}
\vspace{-2mm}
\end{table}
\Cref{tab:deit_cat_supercols} shows that CAT consistently enhances accuracy for both DeiT-S and DeiT-B across all tested
precisions. The improvements are most pronounced under very low-bit precision,
reaching up to +0.4\% at W2A2 (e.g., DeiT-B with BRECQ:
$12.79\%$ vs.\ $12.46\%$) and +0.3\% at W2A6 (e.g., DeiT-S with QDrop:
$66.04\%$ vs.\ $65.71\%$). At more moderate bit-widths (W4A4, W4A6, W6A6), CAT
provides smaller but consistent gains in the range of +0.1--0.2\%,
indicating that the benefit decreases as the quantized model approaches
full-precision accuracy. Importantly, improvements are observed for both
BRECQ and QDrop baselines and across both DeiT-S and
DeiT-B models, highlighting CAT’s robustness and generality as a
lightweight correction mechanism for transformer architectures under different
quantization regimes.
\section{Discussion}
\label{Section:Discussion}
Our experiments demonstrate that CAT consistently improves the performance of a wide range of PTQ baselines across architectures and precision settings. On average, we observe absolute accuracy gains of $+1.56\%$ for W2A2, $+1.19\%$ for W4A2, $+0.39\%$ for W2A4, and $+0.10\%$ for W4A4 within the largest parameter regime ($20M-40M$). Smaller-capacity models ($<10M$) also benefit, though to a lesser extent, with gains of $+0.68\%$ (W2A2) and $+0.96\%$ (W4A2). These trends confirm that CAT is particularly effective in challenging low-precision regimes, while improvements diminish as precision increases.  
A key observation is that {lower activation precision} yields the largest relative improvements. In both W2A2 and W4A2 regimes, CAT provides the strongest corrections, while W2A4 and W4A4 exhibit smaller yet consistent gains. This pattern indicates that activation quantization is the dominant factor driving representational degradation, and CAT’s cluster-specific corrections directly mitigate this bottleneck. In contrast, weight precision plays a comparatively minor role: lowering weights to 2-bit (while keeping activations higher) causes less severe degradation and thus smaller relative improvements with CAT.  

We also find that {larger models} ($20M-40M$ parameters, e.g., ResNet-50, RegNetX-3.2GF) exhibit the highest absolute improvements in LQ settings, surpassing $+1\%$ in both W2A2 and W4A2. Our interpretation is that these models possess greater capacity to process complex information, but when activations are severely quantized, their representational potential is limited by activation precision. CAT effectively restores this information flow by aligning low-precision activations with their full-precision distributions. In contrast, compact models (3--7M, e.g., MobileNetV2, MNasX2) still benefit from CAT but show smaller relative gains, as their limited redundancy constrains the extent to which CAT can compensate.  
In summary, the results highlight three consistent trends:  
(i) CAT is most effective when activation precision is low,  
(ii) larger models benefit more in absolute terms, and  
(iii) weight quantization alone has a weaker influence on CAT’s efficacy compared to activation quantization.  
Together, these findings position CAT as a practical and robust enhancement for PTQ pipelines, particularly in ultra-low-precision and high-capacity scenarios where quantization errors are most detrimental.

\begin{table}[ht]
\centering
\caption{Average Top-1 improvement $\Delta$ (\%) of CAT over Base by parameter regimes and precision.
Regimes: $<10M$ (MobileNetV2, MNasX2, RegNetX-600MF), $10M-20M$ (ResNet-18), $20M-40M$ (ResNet-50, RegNetX-3.2GF).
Darker green indicates larger improvement.}
\label{tab:avg_delta_by_bucket_final}
\setlength{\tabcolsep}{8pt}
\renewcommand{\arraystretch}{1.1}
\begin{tabular}{lcccc}
\toprule
{Params Regime} & {W2A2} & {W4A2} & {W2A4} & {W4A4} \\
\midrule
{$<10M$}   
  & \cellcolor{medgreen}{0.68} 
  & \cellcolor{darkgreen}{0.96} 
  & \cellcolor{lightgreen}{0.32} 
  & \cellcolor{lightgreen}{0.37} \\
{$10M-20M$} 
  & \cellcolor{medgreen}{0.53} 
  & \cellcolor{medgreen}{0.47} 
  & \cellcolor{lightgreen}{0.16} 
  & \cellcolor{lightgreen}{0.07} \\
{$20M-40M$} 
  & \cellcolor{darkgreen}{1.56} 
  & \cellcolor{darkgreen}{1.19} 
  & \cellcolor{medgreen}{0.39} 
  & \cellcolor{lightgreen}{0.10} \\
\bottomrule
\end{tabular}
\end{table}
\end{document}